\documentclass[10pt]{article}
\usepackage{a4wide}
\usepackage[utf8]{inputenc}
\usepackage[T1]{fontenc}
\usepackage{graphicx}
\usepackage{grffile}
\usepackage{longtable}
\usepackage{wrapfig}
\usepackage{rotating}
\usepackage[normalem]{ulem}
\usepackage{amsmath}
\usepackage{textcomp}
\usepackage{amssymb}
\usepackage{capt-of}
\usepackage{hyperref}
\usepackage{multicol}
\usepackage{enumitem}
\usepackage{xcolor}
\usepackage{cancel}

\usepackage{tikz} 
\usepackage{lipsum}

\newcommand{\mybox}[4]{
	\begin{figure}[h!]
		\centering
		\begin{tikzpicture}
			\node[anchor=text,text width=\columnwidth-1.2cm, draw, rounded corners, line width=1pt, fill=#3, inner sep=5mm] (big) {\\#4};
			\node[draw, rounded corners, line width=.5pt, fill=#2, anchor=west, xshift=5mm] (small) at (big.north west) {#1};
		\end{tikzpicture}
	\end{figure}
}

\usepackage{xcolor}
\definecolor{azure}{rgb}{0.1, 0.5, 0.7}
\newcommand{\textbfazure}[1]{\textcolor{azure}{\textbf{#1}}}

\usepackage{fancyhdr}
\title{\LARGE Notes on Various Errors and Jacobian Derivations \\ for SLAM}
\author{Gyubeom Edward Im\thanks{blog: \href{https://alida.tistory.com}{alida.tistory.com}, email: \href{mailto:criterion.im@gmail.com}{criterion.im@gmail.com}}}
\date{May 1, 2024}

\begin{document} 
	\maketitle
	\tableofcontents
	
\section{Introduction}
In this post, we discuss the definitions of various errors used in SLAM and the Jacobians utilized for their optimization. The errors covered in this post are as follows...

\begin{itemize}
	\item  Reprojection error \begin{equation}
		\begin{aligned}
			\mathbf{e} & = \mathbf{p} - \hat{\mathbf{p}} \in \mathbb{R}^{2}
		\end{aligned} 
	\end{equation}
	\item Photometric error \begin{equation}
		\begin{aligned}
			\mathbf{e} & =  \mathbf{I}_{1}(\mathbf{p}_{1})-\mathbf{I}_{2}(\mathbf{p}_{2}) \in \mathbb{R}^{1}
		\end{aligned} 
	\end{equation}
	\item Relative pose error (PGO)  \begin{equation}
		\begin{aligned}
			\mathbf{e}_{ij}  = \text{Log} (\mathbf{z}_{ij}^{-1}\hat{\mathbf{z}}_{ij}) \in \mathbb{R}^{6}
		\end{aligned}
	\end{equation} 
	\item Line reprojection error \begin{equation}
		\begin{aligned}
			\mathbf{e}_{l} =  \begin{bmatrix} \frac{\mathbf{x}_{s}^{\intercal}l_{c}}{\sqrt{l_{1}^{2} + l_{2}^{2}}},  & \frac{\mathbf{x}_{e}^{\intercal}l_{c}}{\sqrt{l_{1}^{2} + l_{2}^{2}}}\end{bmatrix}  \in \mathbb{R}^{2}
		\end{aligned}
	\end{equation}
	\item IMU measurement error  :  \begin{equation} 
		\begin{aligned} 
			\mathbf{e}_{\mathcal{B}} = \begin{bmatrix} 
				\delta \alpha^{b_{k}}_{b_{k+1}} \\ 
				\delta \boldsymbol{\theta}^{b_{k}}_{b_{k+1}} \\ 
				\delta \beta^{b_{k}}_{b_{k+1}} \\ 
				\delta \mathbf{b}_{a}\\ 
				\delta \mathbf{b}_{g}\\ 
			\end{bmatrix} = \begin{bmatrix} 
				\mathbf{R}^{b_{k}}_{w}(\mathbf{p}^{w}_{b_{k+1}} - \mathbf{p}^{w}_{b_{k}} - \mathbf{v}^{w}_{b_{k}}\Delta t_{k} + \frac{1}{2} \mathbf{g}^{w}\Delta t_{k}^{2}) - \hat{\alpha}^{b_{k}}_{b_{k+1}} \\ 
				2 \Big[ \Big( \hat{\gamma}^{b_{k}}_{b_{k+1}} \Big)^{-1} \otimes (\mathbf{q}^{w}_{b_{k}})^{-1} \otimes \mathbf{q}^{w}_{b_{k+1}}  \Big]_{xyz} \\ 
				\mathbf{R}^{b_{k}}_{w} ( \mathbf{v}^{w}_{b_{k+1}} - \mathbf{v}^{w}_{b_{k}} + \mathbf{g}^{w}\Delta t_{k}) - \hat{\beta}^{b_{k}}_{b_{k+1}} \\ 
				\mathbf{b}_{a_{k+1}} - \mathbf{b}_{a_{k}} \\ 
				\mathbf{b}_{g_{k+1}} - \mathbf{b}_{g_{k}} \\ 
			\end{bmatrix} 
		\end{aligned} 
	\end{equation}
\end{itemize}

Depending on whether the camera pose is expressed as a rotation matrix $\mathbf{R} \in SO(3)$ or a transformation matrix $\mathbf{T} \in SE(3)$, different Jacobians are derived. Jacobians for reprojection errors are derived for SO(3), and Jacobians for photometric errors are derived for SE(3). The representation of a point in 3D space as $\mathbf{X}=[X,Y,Z,W]^{\intercal}$ or using inverse depth $\rho$ also affects the Jacobian derivation. The derivation processes for both cases are explained.
~\\ ~\\
The Jacobians discussed in this post are as follows. 
\begin{itemize}
	\item Camera pose (SO(3)-based) \begin{equation}
		\begin{aligned}
			\frac{\partial \mathbf{e}}{\partial \mathbf{R}} \rightarrow \frac{\partial \mathbf{e}}{\partial \Delta \mathbf{w}}  \end{aligned} 
	\end{equation}
	\item Camera pose (SE(3)-based) \begin{equation}
		\begin{aligned}
			\frac{\partial \mathbf{e}}{\partial \mathbf{T}} \rightarrow \frac{\partial \mathbf{e}}{\partial \Delta \xi}  \end{aligned} 
	\end{equation}
	\item Map point \begin{equation}
		\begin{aligned}
			\frac{\partial \mathbf{e}}{\partial \mathbf{X}} \end{aligned} 
	\end{equation}
	\item Relative pose (SE(3)-based) \begin{equation}
		\begin{aligned}
			& \frac{\partial \mathbf{e}_{ij}}{\partial \Delta \xi_{i}},  \frac{\partial \mathbf{e}_{ij}}{\partial \Delta \xi_{j}} \end{aligned} 
	\end{equation} 
	\item 3D plücker line \begin{equation}
		\begin{aligned}
			& \frac{\partial \mathbf{e}_{l}}{\partial l}, \frac{\partial l}{\partial \mathcal{L}_{c}}, \frac{\partial \mathcal{L}_{c}}{\partial \mathcal{L}_{w}}, \frac{\partial \mathcal{L}_{w}}{\partial\delta_{\boldsymbol{\theta}}} \end{aligned} 
	\end{equation}  
	\item Quaternion representation \begin{equation}
		\begin{aligned}
			& \frac{\partial \mathbf{X}'}{\partial \mathbf{q}} 
	\end{aligned} \end{equation}  
	\item Camera intrinsics \begin{equation}
		\begin{aligned}
			& \frac{\partial \mathbf{e}}{\partial f_{x}}, \frac{\partial \mathbf{e}}{\partial f_{y}}, \frac{\partial \mathbf{e}}{\partial c_{x}}, \frac{\partial \mathbf{e}}{\partial c_{y}} \end{aligned} 
	\end{equation}  
	\item Inverse depth  \begin{equation}
		\begin{aligned}
			\frac{\partial \mathbf{e}}{\partial \rho} \end{aligned} 
	\end{equation}  
	\item IMU error-state system kinematics :  \begin{equation}   \begin{aligned}    \mathbf{J}^{b_{k}}_{b_{k+1}}  \end{aligned} \end{equation}
	\item IMU measurement :  \begin{equation}   \begin{aligned} 
			\frac{\partial \mathbf{e}_{\mathcal{B}} }{\partial [\mathbf{p}^{w}_{b_{k}}, \mathbf{q}^{w}_{b_{k}}]},\frac{\partial \mathbf{e}_{\mathcal{B}} }{\partial [\mathbf{v}^{w}_{b_{k}}, \mathbf{b}_{a_{k}}, \mathbf{b}_{g_{k}}]} , \frac{\partial \mathbf{e}_{\mathcal{B}} }{\partial [\mathbf{p}^{w}_{b_{k+1}}, \mathbf{q}^{w}_{b_{k+1}}]}, \frac{\partial \mathbf{e}_{\mathcal{B}} }{\partial [\mathbf{v}^{w}_{b_{k+1}}, \mathbf{b}_{a_{k+1}}, \mathbf{b}_{g_{k+1}}]}
		\end{aligned} 
	\end{equation}
\end{itemize}

\section{Optimization formulation} \label{sec:opt}
\subsection{Error derivation}
In SLAM, the error is defined as the difference between the observed value (measurement) $\mathbf{z}$ and the predicted value (estimate) $\hat{\mathbf{z}}$ based on sensor data.
\begin{equation}
	\boxed{ \begin{aligned}
			\mathbf{e}(\mathbf{x}) = \mathbf{z} - \mathbf{\hat{z}}(\mathbf{x})
	\end{aligned} }
\end{equation}
- $\mathbf{x}$: model state variables \\

As such, the difference between the observed and predicted values is defined as the error, and the optimal state variables $\mathbf{x}$ that minimize this error become the optimization problem in SLAM. In general, since the state variables in SLAM include non-linear terms related to rotation, the non-linear least squares method is mainly used. 

\subsection{Error function derivation}
Typically, when a large amount of sensor data comes in, dozens to hundreds of errors are calculated in vector form. At this time, it is assumed that the error follows a normal distribution, and the work of converting it into an error function is performed. 
\begin{equation}
	\begin{aligned}
		\mathbf{e}(\mathbf{x}) = \mathbf{z} - \mathbf{\hat{z}}  \sim \mathcal{N}(0, \mathbf{\Sigma})
	\end{aligned}
\end{equation}

\mybox{Tip}{gray!40}{gray!10}{
	The multivariate normal distribution of the probability variable $\mathbf{x}$ for modeling the error function is as follows. 
	\begin{equation}
		\begin{aligned}
			p(\mathbf{x})=\frac{1}{\sqrt{(2\pi)^n|\boldsymbol\Sigma|}}
			\exp\left(-\frac{1}{2}({\mathbf{x}}-{\mu})^{\intercal}{\boldsymbol\Omega}({\mathbf{x}}-{\mu})
			\right) \sim \mathcal{N}(\mu, \mathbf{\Sigma})
		\end{aligned}
	\end{equation}
	- $\boldsymbol\Omega = \boldsymbol\Sigma^{-1}$ : information matrix (inverse of covariance matrix) \\
}

The error can be modeled as a multivariate normal distribution with mean $0$ and variance $\mathbf{\Sigma}$. Applying the log-likelihood to this equation, $\ln p(\mathbf{e})$ is as follows.
\begin{equation}
	\begin{aligned}
		\ln p(\mathbf{e}) & \propto -\frac{1}{2}(\mathbf{z} - \mathbf{\hat{z}})^{T}\boldsymbol{\Omega}(\mathbf{z} - \mathbf{\hat{z}}) \\
		& \propto -\frac{1}{2}\mathbf{e}^{\intercal}\boldsymbol{\Omega}\mathbf{e}
	\end{aligned}
\end{equation}

Finding $\mathbf{x}^{*}$ where log-likelihood $\ln p(\mathbf{e})$ is maximized results in the highest probability of the multivariate normal distribution. This is called Maximum Likelihood Estimation (MLE). Since $\ln p(\mathbf{e})$ has a negative (-) sign in front, finding the minimum of the negative log-likelihood $\ln p(\mathbf{e})$ is as follows.
\begin{equation}
	\begin{aligned}
		\mathbf{x}^{*} = \arg\max p(\mathbf{e})= \arg\min\mathbf{e}^{T}\boldsymbol{\Omega}\mathbf{e}
	\end{aligned}
\end{equation}

\textbfazure{If all errors are added instead of a single error, it is expressed as follows, and this is called the error function $\mathbf{E}$. In actual optimization problems, not the single error $\mathbf{e}_{i}$ but the error function $\mathbf{E}$ that minimizes $\mathbf{x}^{*}$ is found.}
\begin{equation}
	\boxed { \begin{aligned}
			\mathbf{E}(\mathbf{x}) & = \sum_{i}\mathbf{e}_{i}^{T}\boldsymbol{\Omega}_{i}\mathbf{e}_{i} \\
			\mathbf{x}^{*} & = \arg\min \mathbf{E}(\mathbf{x})
	\end{aligned} }
\end{equation}

\subsection{Non-linear least squares}
The final optimization equation to be solved is as follows. 
\begin{equation}
	\begin{aligned}
		\mathbf{x}^{*} = \arg\min \mathbf{E}(\mathbf{x}) = \arg\min \sum_{i}\mathbf{e}_{i}^{T}\boldsymbol{\Omega}_{i}\mathbf{e}_{i}
	\end{aligned}
\end{equation}

In the above formula, the optimal parameter $\mathbf{x}^{*}$ that minimizes the error must be found. \textbfazure{However, the above formula typically includes non-linear terms related to rotation in SLAM, so no closed-form solution exists. Therefore, non-linear optimization methods (Gauss-Newton (GN), Levenberg-Marquardt (LM)) must be used to solve the problem.} Among the actual implemented SLAM codes, the information matrix $\boldsymbol{\Omega}_{i}$ is often set to $\mathbf{I}_{3}$ to find the optimal value for $\mathbf{e}_{i}^{\intercal}\mathbf{e}_{i}$.
~\\~\\
For example, let's assume that the problem is solved using the GN method. The order of solving the problem is as follows.
\begin{itemize}
	\item Define the error function
	\item Approximate linearization using Taylor expansion
	\item Set the first derivative to zero.
	\item Calculate the value and substitute it into the error function
	\item Repeat until convergence.
\end{itemize}
~\\
If the error function $\mathbf{e}$ is detailed, it appears as $\mathbf{e}(\mathbf{x})$, meaning that the value of the error function changes according to the robot's pose vector $\mathbf{x}$. The GN method updates the increment $\Delta \mathbf{x}$ iteratively in a direction that reduces the error for $\mathbf{e}(\mathbf{x})$.
\begin{equation}
	\begin{aligned}
		\mathbf{e}(\mathbf{x}+\Delta \mathbf{x})^{\intercal}\Omega\mathbf{e}(\mathbf{x}+\Delta \mathbf{x})
	\end{aligned}
\end{equation}

When $\mathbf{e}(\mathbf{x}+\Delta \mathbf{x})$ is used near $\mathbf{x}$ with a first-order Taylor expansion, the above equation is approximated as follows.
\begin{equation}
	\begin{aligned}
		\mathbf{e}(\mathbf{x} + \Delta \mathbf{x}) \rvert_{\mathbf{x}} & \approx \mathbf{e}(\mathbf{x}) + \mathbf{J}(\mathbf{x} + \Delta \mathbf{x} - \mathbf{x})\\
		& = \mathbf{e}(\mathbf{x}) + \mathbf{J}\Delta \mathbf{x}
	\end{aligned}
\end{equation}

At this time, $\mathbf{J} = \frac{\partial \mathbf{e}(\mathbf{x} + \Delta \mathbf{x})}{\partial \mathbf{x}}$. When this is applied to the entire error function, it is as follows.
\begin{equation}
	\begin{aligned}
		\mathbf{e}(\mathbf{x}+\Delta \mathbf{x})^{\intercal}\Omega\mathbf{e}(\mathbf{x}+\Delta \mathbf{x}) \approx (\mathbf{e}+\mathbf{J}\Delta \mathbf{x})^{\intercal}\Omega
		(\mathbf{e}+\mathbf{J}\Delta \mathbf{x})
	\end{aligned}
\end{equation}

After expanding the above equation and substituting, it is as follows.
\begin{equation}
	\begin{aligned}
		& = \underbrace{\mathbf{e}^{\intercal}\Omega\mathbf{e}}_{\mathbf{c}} + 2 \underbrace{\mathbf{e}^{\intercal}\Omega\mathbf{J}}_{\mathbf{b}} \Delta \mathbf{x} + \Delta \mathbf{x}^{\intercal} \underbrace{\mathbf{J}^{\intercal}\Omega\mathbf{J}}_{\mathbf{H}} \Delta \mathbf{x} \\
		& = \mathbf{c}+ 2\mathbf{b}\Delta \mathbf{x} + \Delta \mathbf{x}^{\intercal} \mathbf{H} \Delta \mathbf{x}
		\end{aligned}
	\end{equation}
	
	The overall error applied is as follows.
	\begin{equation}
		\begin{aligned}
			\mathbf{E}(\mathbf{x}+\Delta \mathbf{x}) = \sum_{i}\mathbf{e}_{i}^{\intercal}\Omega_{i}\mathbf{e}_{i} = \mathbf{c}+ 2\mathbf{b}\Delta \mathbf{x} + \Delta \mathbf{x}^{T} \mathbf{H} \Delta \mathbf{x}
			\end{aligned}
		\end{equation}
		
		$\mathbf{E}(\mathbf{x}+\Delta \mathbf{x})$ is in a quadratic form about $\Delta \mathbf{x}$ and since $\mathbf{H} = \mathbf{J}^{\intercal}\Omega \mathbf{J}$ is a positive definite matrix, the first derivative of $\mathbf{E}(\mathbf{x} + \Delta \mathbf{x})$ set to zero determines the minimum of $\Delta \mathbf{x}$.
		\begin{equation}
			\begin{aligned}
				\frac{\partial \mathbf{E}(\mathbf{x}+\Delta \mathbf{x})}{\partial \Delta \mathbf{x}}  \approx 2\mathbf{b} + 2\mathbf{H} \Delta \mathbf{x} = 0
			\end{aligned}
		\end{equation}
		
		This leads to the following formula being derived.
		\begin{equation}
			\begin{aligned}
				\mathbf{H}\Delta \mathbf{x} = - \mathbf{b}
				\end{aligned}
			\end{equation}
			
			Thus obtained $\Delta \mathbf{x} = -\mathbf{H}^{-1}\mathbf{b}$ is updated to $\mathbf{x}$.
			\begin{equation}
				\begin{aligned}
					\mathbf{x} \leftarrow \mathbf{x} + \Delta \mathbf{x}
					\end{aligned}
				\end{equation}
				
				\textbfazure{The algorithm that iteratively performs the process so far is called the Gauss-Newton method.} The LM method, compared to the GN method, has the same overall process, however, in the formula for calculating the increment, a damping factor $\lambda$ term is added.
				\begin{equation}
					\boxed{ \begin{aligned}
							& \text{(GN) }\mathbf{H}\Delta \mathbf{x} = - \mathbf{b} \\
							& \text{(LM) }(\mathbf{H} + \lambda \mathbf{I})\Delta \mathbf{x} = - \mathbf{b}
					\end{aligned} }
				\end{equation}

\section{Reprojection Error}
Reprojection error is primarily used in feature-based Visual SLAM. It is commonly used when performing feature-based method visual odometry (VO) or bundle adjustment (BA). For more details on BA, refer to the post \href{https://alida.tistory.com/51}{[SLAM] Bundle Adjustment Concept Review}.
~\\ ~\\
\textbf{NOMENCLATURE of reprojection error}
\begin{itemize}
	\item $\tilde{\mathbf{p}} = \pi_{h}(\cdot) : \begin{bmatrix} X' \\ Y' \\ Z' \\1 \end{bmatrix} \rightarrow  \begin{bmatrix} X'/Z' \\ Y'/Z' \\ 1  \end{bmatrix}   = \begin{bmatrix} \tilde{u} \\ \tilde{v} \\ 1 \end{bmatrix}$
	\begin{itemize}
		\item Point $\mathbf{X}'$ in 3D space non-homogeneously transformed to be projected onto the image plane
	\end{itemize}
	
	\item $\hat{\mathbf{p}} = \pi_{k}(\cdot) = \tilde{\mathbf{K}} \tilde{\mathbf{p}} = \begin{bmatrix} f & 0 & c_{x} \\ 0 & f & c_{y} \end{bmatrix} \begin{bmatrix} \tilde{u} \\ \tilde{v} \\ 1 \end{bmatrix}= \begin{bmatrix} f\tilde{u} + c_{x} \\ f\tilde{v} + c_{y} \end{bmatrix} = \begin{bmatrix} u \\ v \end{bmatrix}$
	\begin{itemize}
		\item Point projected onto the image plane after lens distortion correction. If distortion correction has already been performed at the input stage, $\pi_{k}(\cdot) = \tilde{\mathbf{K}}(\cdot)$.
	\end{itemize}
	
	\item $\mathbf{K} = \begin{bmatrix} f & 0 & c_{x} \\ 0 & f & c_{y} \\ 0 & 0 & 1 \end{bmatrix}$: Camera's intrinsic parameters
	\item $\tilde{\mathbf{K}} = \begin{bmatrix} f & 0 & c_{x} \\ 0 & f & c_{y}  \end{bmatrix}$: Omitting the last row of the intrinsic parameters for projection from $\mathbb{P}^{2} \rightarrow \mathbb{R}^{2}$.
	\item $\mathcal{X} = [\mathcal{T}_{1}, \cdots, \mathcal{T}_{m}, \mathbf{X}_{1}, \cdots, \mathbf{X}_{n}]^{\intercal}$: Model's state variables
	\item $m$: Number of camera poses
	\item $n$: Number of 3D points
	\item $\mathcal{T}_{i} = [\mathbf{R}_{i}, \mathbf{t}_{i}]$
	\item $\mathbf{e}_{ij} = \mathbf{e}_{ij}(\mathcal{X})$: Notation simplified by omitting $\mathcal{X}$
	\item $\mathbf{p}_{ij}$: Observed pixel coordinates of a feature point
	\item $\hat{\mathbf{p}}_{ij}$: Estimated pixel coordinates of a feature point
	\item $\mathbf{T}_{i}\mathbf{X}_{j}$: Transformation, 3D point $\mathbf{X}_{j}$ transformed to camera coordinate system $\{i\}$, $\bigg( \mathbf{T}_{i}\mathbf{X}_{j} = \begin{bmatrix} \mathbf{R}_{i} \mathbf{X}_{j} + \mathbf{t}_{i} \\ 1 \end{bmatrix}  \in \mathbb{R}^{4\times1} \bigg)$
	\begin{itemize}
		\item $\mathbf{X}' = \mathbf{TX} = [X', Y', Z',1]^{\intercal} = [\tilde{\mathbf{X}}', 1]^{\intercal} $
	\end{itemize}
	
	\item $\oplus$: Operator that updates rotation matrix $\mathbf{R}$ and 3D vector $\mathbf{t}, \mathbf{X}$ simultaneously.
	\item $\mathbf{J} = \frac{\partial \mathbf{e}}{\partial \mathcal{X}} = \frac{\partial \mathbf{e}}{\partial [\mathcal{T}, \mathbf{X}]}$
	\item $\mathbf{w} = \begin{bmatrix} w_{x}&w_{y}&w_{z} \end{bmatrix}^{\intercal}$: Angular velocity
	\item $[\mathbf{w}]_{\times} = \begin{bmatrix} 0 & -w_{z} & w_{y} \\ w_{z} & 0 & -w_{x} \\ -w_{y} & w_{x} & 0 \end{bmatrix}$: Skew-symmetric matrix of angular velocity $\mathbf{w}$
\end{itemize}
~\\
\begin{figure}[h!]
	\centering
	\includegraphics[width=6cm]{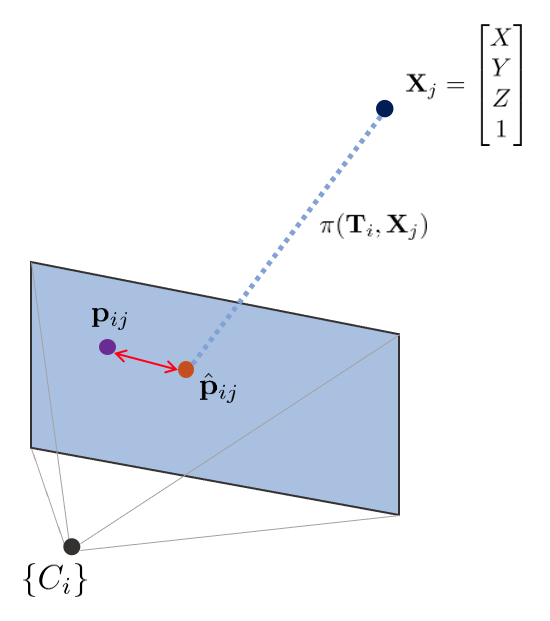}
\end{figure}
When there is a pinhole camera pose $\{C_{i}\}$ and a world point $\mathbf{X}_{j}$, $\mathbf{X}_{j}$ is projected onto the image plane through the following transformation.
\begin{equation} \begin{aligned} \text{projection model:     }\quad\hat{\mathbf{p}}_{ij} = \pi(\mathbf{T}_{i}, \mathbf{X}_{j}) \end{aligned} \end{equation}

\begin{figure}[h!]
	\centering
	\includegraphics[width=16cm]{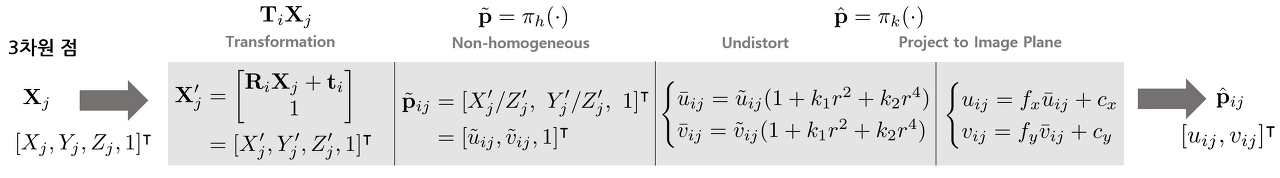}
\end{figure}
The model utilizing the camera's intrinsic/extrinsic parameters is called the projection model. Reprojection error is defined as follows:
\begin{equation}
	\boxed{
		\begin{aligned}
			\mathbf{e}_{ij} & = \mathbf{p}_{ij} - \hat{\mathbf{p}}_{ij} \\
			& = \mathbf{p}_{ij} - \pi(\mathbf{T}_{i}, \mathbf{X}_{j}) \\
			& = \mathbf{p}_{ij} - \pi_{k}(\pi_{h}(\mathbf{T}_{i}\mathbf{X}_{j}))
	\end{aligned} } \label{eq:reproj1}
\end{equation}

The error function for all camera poses and 3D points is defined as follows.
\begin{equation}
	\begin{aligned}
		\mathbf{E}(\mathcal{X}) & =   \sum_{i}\sum_{j} \left\| \mathbf{e}_{ij} \right\|^{2} \\
	\end{aligned}
\end{equation}
\begin{equation}
	\begin{aligned}
		\mathcal{X}^{*} & = \arg\min_{\mathcal{X}^{*}}  \mathbf{E}(\mathcal{X}) \\ 
		& =  \arg\min_{\mathcal{X}^{*}} \sum_{i}\sum_{j} \left\| \mathbf{e}_{ij} \right\|^{2} \\
		& = \arg\min_{\mathcal{X}^{*}} \sum_{i}\sum_{j} \mathbf{e}_{ij}^{\intercal}\mathbf{e}_{ij} \\
		& = \arg\min_{\mathcal{X}^{*}} \sum_{i}\sum_{j} (\mathbf{p}_{ij} - \hat{\mathbf{p}}_{ij})^{\intercal}(\mathbf{p}_{ij} - \hat{\mathbf{p}}_{ij})
	\end{aligned} \label{eq:10}
\end{equation}

The error $\left\|\mathbf{e}(\mathcal{X}^{*})\right\|^{2}$ satisfying $\mathbf{E}(\mathcal{X}^{*})$ can be calculated iteratively through non-linear least squares. By repeatedly updating a small increment $\Delta \mathcal{X}$ to $\mathcal{X}$, the optimal state is found.
\begin{equation}
	\begin{aligned}
		\arg\min_{\mathcal{X}^{*}} \mathbf{E}(\mathcal{X} + \Delta \mathcal{X}) & = \arg\min_{\mathcal{X}^{*}} \sum_{i}\sum_{j} \left\|\mathbf{e}(\mathcal{X} +\Delta \mathcal{X})\right\|^{2} 
	\end{aligned}
\end{equation}

Strictly speaking, since the state increment $\Delta \mathcal{X}$ includes an SO(3) rotation matrix, it is correct to add it using the $\oplus$ operator to the existing state $\mathcal{X}$, but the $+$ operator is used for convenience of expression.
\begin{equation}
	\begin{aligned}
		\mathbf{e}( \mathcal{X} \oplus \Delta \mathcal{X}) 
		\quad \rightarrow \quad\mathbf{e}(\mathcal{X} + \Delta \mathcal{X})
	\end{aligned}
\end{equation}

This equation can be expressed through the first-order Taylor approximation as follows.
\begin{equation}
	\begin{aligned}
		\mathbf{e}(\mathcal{X} + \Delta \mathcal{X})  & \approx \mathbf{e}(\mathcal{X}) + \mathbf{J}\Delta \mathcal{X} \\ & = \mathbf{e}(\mathcal{X}) + \mathbf{J}_{c} \Delta \mathcal{T} + \mathbf{J}_{p} \Delta \mathbf{X} \\
		& = \mathbf{e}(\mathcal{X}) + \frac{\partial \mathbf{e}}{\partial \mathcal{T}} \Delta \mathcal{T} + \frac{\partial \mathbf{e}}{\partial \mathbf{X}}\Delta \mathbf{X} \\
	\end{aligned}
\end{equation}

\begin{equation}
	\begin{aligned}
		\arg\min_{\mathcal{X}^{*}} \mathbf{E}(\mathcal{X} + \Delta \mathcal{X}) & \approx \arg\min_{\mathcal{X}^{*}} \sum_{i}\sum_{j} \left\|\mathbf{e}(\mathcal{X}) + \mathbf{J}\Delta \mathcal{X} \right\|^{2} \\
	\end{aligned}
\end{equation}

The optimal increment $\Delta \mathcal{X}^{*}$ is found by differentiating the above expression. The derivation process is omitted in this section. For detailed information on the derivation process, refer to the \hyperref[sec:opt]{previous section}.
\begin{equation}
	\begin{aligned}
		& \mathbf{J}^{\intercal}\mathbf{J} \Delta \mathcal{X}^{*} = -\mathbf{J}^{\intercal}\mathbf{e} \\ 
		& \mathbf{H}\Delta \mathcal{X}^{*} = - \mathbf{b} \\
	\end{aligned} \label{eq:6}
\end{equation}

This equation is in the form of a linear system $\mathbf{Ax} = \mathbf{b}$, thus $\Delta \mathcal{X}^{*}$ can be found using various linear algebra techniques like schur complement and cholesky decomposition. In this case, $\mathbf{t}, \mathbf{X}$ among the existing states $\mathcal{X}$ exist in a linear vector space, so there is no difference depending on whether they are added from the right or from the left, but \textbfazure{since the rotation matrix $\mathbf{R}$ belongs to the non-linear SO(3) group, it depends on whether it is multiplied from the right or from the left whether to update the pose seen in the local coordinate system (right) or the pose seen in the global coordinate system (left). Reprojection error updates the transformation matrix of the global coordinate system, so it generally uses the left multiplication method.}\begin{equation}
	\begin{aligned}
		\mathcal{X} \leftarrow  \mathcal{X} \oplus \Delta \mathcal{X}^{*} \\
	\end{aligned}
\end{equation}

$\mathcal{X}$ consists of $[\mathcal{T}, \mathbf{X}]$, so it can be described as follows.
\begin{equation}
	\begin{aligned}
		\mathcal{T} \leftarrow  \mathcal{T} &  \oplus \Delta \mathcal{T}^{*}\\
		\mathbf{X} \leftarrow \mathbf{X} &  \oplus \Delta \mathbf{X}^{*}   \\
	\end{aligned} 
\end{equation}

The definition of the left multiplication $\oplus$ operation is as follows.
\begin{equation}
	\begin{aligned}
		\mathbf{R}  \oplus \Delta \mathbf{R}^{*} & = \Delta \mathbf{R}^{*} \mathbf{R} \\ 
		& = \exp([\Delta \mathbf{w}^{*}]_{\times})\mathbf{R} \quad \cdots \text{ globally updated (left mult)} \\ 
		\mathbf{t} \oplus  \Delta \mathbf{t}^{*} & =  \mathbf{t} + \Delta \mathbf{t}^{*} \\
		\mathbf{X} \oplus \Delta \mathbf{X}^{*}  & =  \mathbf{X}  + \Delta \mathbf{X}^{*}
	\end{aligned} \label{eq:1}
\end{equation}

\subsection{Jacobian of the Reprojection Error}
\subsubsection{Jacobian of Camera Pose}
The Jacobian of the pose $\mathbf{J}_{c}$ can be decomposed as follows.
\begin{equation} \begin{aligned} \mathbf{J}_{c} = \frac{\partial \mathbf{e}}{\partial \mathcal{T}} & = \frac{\partial}{\partial \mathcal{T}}(\mathbf{p} - \hat{\mathbf{p}}) \\ & = \frac{\partial }{\partial \mathcal{T}} \bigg(\mathbf{p} - \pi_{k}(\pi_{h}(\mathbf{T}_{i} \mathbf{X}_{j})) \bigg ) \\ & = \frac{\partial}{\partial \mathcal{T}} \bigg(-\pi_{k}(\pi_{h}(\mathbf{T}_{i} \mathbf{X}_{j})) \bigg ) \end{aligned} \end{equation}

Using the chain rule, the above formula is organized as follows. For convenience, $\mathbf{T}_{i}\mathbf{X}_{j} \rightarrow \mathbf{X}' $ is denoted.
\begin{equation}
	\begin{aligned}
		\mathbf{J}_{c}& = \frac{\partial \hat{\mathbf{p}}}{\partial \tilde{\mathbf{p}}} 
		\frac{\partial \tilde{\mathbf{p}}}{\partial \mathbf{X}'}
		\frac{\partial \mathbf{X}'}{\partial [\mathbf{w}, \mathbf{t}]} \\
		& = \mathbb{R}^{2 \times 3} \cdot \mathbb{R}^{3\times 4} \cdot \mathbb{R}^{4 \times 6} = \mathbb{R}^{2 \times 6}
	\end{aligned} \label{eq:2}
\end{equation}

\textbfazure{The reason for calculating the Jacobian $\frac{\partial \mathbf{X}'}{\partial \mathbf{w}}$ for the angular velocity $\mathbf{w}$ instead of the Jacobian $\frac{\partial \mathbf{X}'}{\partial \mathbf{R}}$ for the rotation matrix $\mathbf{R}$ is explained in the next section.} Also, depending on whether the error is defined as $\mathbf{p} - \hat{\mathbf{p}}$ or $\hat{\mathbf{p}} - \mathbf{p}$, the sign of $\mathbf{J}_{c}$ also changes, so this should be carefully applied when implementing the actual code. The sign is considered as $+$ and marked in this material.

If it is assumed that undistortion has already been performed during the image input process, $\frac{\partial \hat{\mathbf{p}}}{\partial \tilde{\mathbf{p}}}$ is as follows.
\begin{equation}
	\boxed{
		\begin{aligned}
			\frac{\partial \hat{\mathbf{p}}}{\partial \tilde{\mathbf{p}}} & = 
			\frac{\partial }{\partial \tilde{\mathbf{p}}} \tilde{\mathbf{K}} \tilde{\mathbf{p}} \\
			& = \tilde{\mathbf{K}} \\
			& = \begin{bmatrix} f & 0 & c_{x} \\ 0 & f & c_{y} \end{bmatrix} \in \mathbb{R}^{2 \times 3}
	\end{aligned} }
\end{equation}

Next, $\frac{\partial \tilde{\mathbf{p}}}{\partial \mathbf{X}'}$ is as follows.
\begin{equation}
	\boxed{
		\begin{aligned}
			\frac{\partial \tilde{\mathbf{p}}}{\partial \mathbf{X}'} & = 
			\frac{\partial [\tilde{u}, \tilde{v}, 1]}{\partial [X', Y', Z',1]} \\
			& = \begin{bmatrix} \frac{1}{Z'} & 0 & \frac{-X'}{Z'^{2}} & 0 \\
				0 & \frac{1}{Z'} & \frac{-Y'}{Z'^{2}} & 0 \\ 0 & 0 & 0 & 0 \end{bmatrix} \in \mathbb{R}^{3\times 4}
	\end{aligned} }
\end{equation}

Next, $\frac{\partial \mathbf{X}'}{\partial \mathbf{t}}$ needs to be calculated. This can be relatively simply obtained as follows.
\begin{equation}
	\boxed{
		\begin{aligned}
			\frac{\partial \mathbf{X}'}{\partial \mathbf{t}} & = \frac{\partial }{\partial [t_{x}, t_{y}, t_{z}]} \begin{bmatrix} \mathbf{R}\mathbf{X} + \mathbf{t} \\ 1 \end{bmatrix} \\
			& = \frac{\partial }{\partial [t_{x}, t_{y}, t_{z}]} \begin{bmatrix} \mathbf{t} \\ 1 \end{bmatrix} \\
			& = \frac{\partial }{\partial [t_{x}, t_{y}, t_{z}]}(\begin{bmatrix} t_{x} \\ t_{y} \\ t_{z} \\ 1 \end{bmatrix}) \\
			& = \begin{bmatrix} 1 & 0 & 0 \\ 0 & 1 & 0 \\ 0 & 0 & 1 \\ 0 & 0 & 0 \end{bmatrix} \in \mathbb{R}^{4\times 3}
	\end{aligned} }
\end{equation}

\subsubsection{Lie Theory-Based SO(3) Optimization} \label{sec:reproj-so3}
Finally, $\frac{\partial \mathbf{X}'}{\partial \mathbf{w}}$ needs to be calculated. In this case, the rotation-related parameter was represented as angular velocity $\mathbf{w}$ instead of the rotation matrix $\mathbf{R}$. The rotation matrix $\mathbf{R}$ has 9 parameters, whereas the actual rotation is limited to 3 degrees of freedom, therefore it is over-parameterized. The disadvantages of an over-parameterized representation are as follows:
\begin{itemize}
	\item Due to the calculation of redundant parameters, the amount of computation required for optimization increases.
	\item Additional degrees of freedom can cause problems with numerical instability.
	\item Each time parameters are updated, it must be checked whether they always satisfy the constraints.
\end{itemize}

Lie theory allows optimization to be performed free from constraints. Therefore, instead of the lie group SO(3) $\mathbf{R}$, the lie algebra so(3) $[\mathbf{w}]_{\times}$ is used to freely update parameters from constraints. Here, $\mathbf{w} \in \mathbb{R}^{3}$ denotes the angular velocity vector.
\begin{equation}
	\begin{aligned}
		\mathbf{J}_{c} = \frac{\partial \mathbf{e}}{\partial [\mathbf{R}, \mathbf{t}]} \rightarrow \frac{\partial \mathbf{e}}{\partial [\mathbf{w}, \mathbf{t}]} \\
	\end{aligned}
\end{equation}

However, since $\mathbf{w}$ is not directly visible from $\mathbf{X}'$, $\mathbf{X}'$ must be represented in lie algebra. At this time, since the Jacobian for the $\mathbf{w}$ term related to rotation needs to be calculated, let's assume that the 3D point $\mathbf{X}_{t}$ is the point $\mathbf{X}$ translated by $\mathbf{t}$ and then $\mathbf{X}'$ is the point $\mathbf{X}_{t}$ rotated by $\mathbf{R}$.
\begin{equation} \begin{aligned} \mathbf{X}_{t} & = \mathbf{X} + \mathbf{t} \\ \mathbf{X}'  & = \mathbf{R}\mathbf{X}_{t} \\ & = \exp([\mathbf{w}]{\times})\mathbf{X}_{t}   \end{aligned} \end{equation}

\mybox{Tip}{gray!40}{gray!10}{
	$\exp([\mathbf{w}]_{\times}) \in SO(3)$ denotes the operation of converting angular velocity $\mathbf{w}$ into a 3D rotation matrix $\mathbf{R}$ using exponential mapping. For detailed information on exponential mapping, see \href{https://alida.tistory.com/60\#2.3.2-the-capitalized-exponential-map}{this link}.
	\begin{equation} \begin{aligned} \exp([\mathbf{w}]_{\times}) = \mathbf{R}   \end{aligned} \end{equation}
}

In this case, depending on how the small lie algebra increment $\Delta \mathbf{w}$ is updated to the existing $\exp([\mathbf{w}]_{\times})$, there are two ways to update. First, there is $[1]$ the basic lie algebra update method. Next, there is $[2]$ the update method using the perturbation model.
\begin{equation}
	\begin{aligned}
		\exp([\mathbf{w}]_{\times}) & \leftarrow \exp([\mathbf{w} + \Delta \mathbf{w}]_{\times}) \quad \cdots \text{[1]} \\ 
		\exp([\mathbf{w}]_{\times}) &  \leftarrow \exp([\Delta \mathbf{w}]_{\times})\exp([\mathbf{w}]_{\times}) \quad \cdots \text{[2]} 
	\end{aligned}
\end{equation}

\mybox{Tip}{gray!40}{gray!10}{
	There is the following relationship between the above two methods. For detailed information, see \href{https://docs.google.com/document/d/1icPjUyT3nPvjZ1OVMtWp9afUtuJ4gXLJL-ex7A9FpNs/edit?fbclid=IwAR2VfhZ3js52zkFpZpJ5HZv_qQLPz7WCTBWkwn6IF1MmHa3Ksyhi5TQSAfY\#}{this link} chapter 4.3.3.
	\begin{equation}
		\begin{aligned}
			& \exp([\Delta \mathbf{w}]_{\times}) \exp([\mathbf{w}]_{\times}) = \exp([\mathbf{w}+ \mathbf{J}_{l}^{-1}\Delta \mathbf{w}]_{\times} ) \\
			& \exp([\mathbf{w}+ \Delta \mathbf{w}]_{\times} ) = \exp([\mathbf{J}_{l} \Delta \mathbf{w}]_{\times}) \exp([\mathbf{w}]_{\times})
		\end{aligned}
	\end{equation}
}

\textbfazure{$[1]$ Lie Algebra-Based Update:} First, using method $[1]$ to directly calculate the Jacobian $\frac{\partial \mathbf{R}\mathbf{X}_{t} }{\partial \mathbf{w}}$ results in the following complex formula.
\begin{equation} \begin{aligned}
		\frac{\partial \mathbf{R}\mathbf{X}_{t} }{\partial \mathbf{w}} & = \lim_{\Delta \mathbf{w} \rightarrow 0} \frac{\exp([\mathbf{w} + \Delta \mathbf{w}]_{\times}) \mathbf{X}_{t}  - \exp([\mathbf{w}]_{\times}) \mathbf{X}_{t} }{\Delta \mathbf{w}} \\
		& \approx \lim_{\Delta \mathbf{w} \rightarrow 0} \frac{ \exp([\mathbf{J}_{l}\Delta \mathbf{w}]_{\times}) (\exp([\mathbf{w}]_{\times}) \mathbf{X}_{t}  - \exp([\mathbf{w}]_{\times}) \mathbf{X}_{t} }{\Delta \mathbf{w}} \\
		& \approx \lim_{\Delta \mathbf{w} \rightarrow 0} \frac{(\mathbf{I} + [\mathbf{J}_{l} \Delta \mathbf{w}]_{\times}) (\exp([\mathbf{w}]_{\times})\mathbf{X}_{t}  - \exp([\mathbf{w}]_{\times})\mathbf{X}_{t}}{\Delta \mathbf{w}} \\
		& = \lim_{\Delta \mathbf{w} \rightarrow 0} \frac{[\mathbf{J}_{l}\Delta \mathbf{w}]_{\times} \mathbf{R}\mathbf{X}_{t} }{\Delta \mathbf{w}}   \quad \quad (\because \exp([\mathbf{w}]_{\times})\mathbf{X}_{t} = \mathbf{R}\mathbf{X}_{t}) \\
		& = \lim_{\Delta \mathbf{w} \rightarrow 0} \frac{-[\mathbf{R}\mathbf{X}_{t}]_{\times} \mathbf{J}_{l}\Delta \mathbf{w}}{\Delta \mathbf{w}} \\
		& = -[\mathbf{R}\mathbf{X}_{t}]_{\times} \mathbf{J}_{l} \\ & = -[\mathbf{X}']_{\times} \mathbf{J}_{l}
\end{aligned} \end{equation}

\mybox{Tip}{gray!40}{gray!10}{
	In the above formula, the second row uses the BCH approximation to derive the left Jacobian (left jacobian) $\mathbf{J}_{l}$, and the third row applies the first-order Taylor approximation for small rotation $\exp([\mathbf{J}_{l} \Delta \mathbf{w}]_{\times})$. For more information on $\mathbf{J}_{l}$, see \href{https://docs.google.com/document/d/1icPjUyT3nPvjZ1OVMtWp9afUtuJ4gXLJL-ex7A9FpNs/edit?fbclid=IwAR2VfhZ3js52zkFpZpJ5HZv_qQLPz7WCTBWkwn6IF1MmHa3Ksyhi5TQSAfY\#}{Visual SLAM Introduction Chapter 4}.
	
	To understand the third row's approximation, given an arbitrary rotation vector $\mathbf{w} = [w_{x}, w_{y}, w_{z}]^{\intercal}$, the rotation matrix can be expanded in exponential mapping form as follows.
	\begin{equation}
		\begin{aligned}
			\mathbf{R} = \exp([\mathbf{w}]_{\times}) = \mathbf{I} +  [\mathbf{w}]_{\times} + \frac{1}{2}[\mathbf{w}]_{\times}^{2} + \frac{1}{3!}[\mathbf{w}]_{\times}^{3} + \frac{1}{4!}[\mathbf{w}]^{4}_{\times} + \cdots
		\end{aligned}
	\end{equation}
	
	For a small rotation matrix $\Delta \mathbf{R}$, higher-order terms beyond the second can be ignored, and it can be approximated as follows.
	\begin{equation}
		\begin{aligned}
			\Delta \mathbf{R} \approx  \mathbf{I} +  [\Delta \mathbf{w}]_{\times} 
		\end{aligned}
	\end{equation}
}

\textbfazure{$[2]$ Perturbation Model-Based Update: } To calculate a simpler Jacobian without using $\mathbf{J}_{l}$, the perturbation model of lie algebra so(3) is generally used. Calculating the Jacobian $\frac{\partial \mathbf{R}\mathbf{X}_{t}}{\partial \Delta \mathbf{w}}$ using the perturbation model results in the following.
\begin{equation}  \begin{aligned}
		\frac{\partial \mathbf{R}\mathbf{X}_{t}}{\partial \Delta \mathbf{w}} & = \lim_{\Delta \mathbf{w} \rightarrow 0} \frac{\exp([\Delta \mathbf{w}]_{\times}) \exp([\mathbf{w}]_{\times}) \mathbf{X}_{t}  - \exp([\mathbf{w}]_{\times})  \mathbf{X}_{t} }{\Delta \mathbf{w}} \\
		& \approx \lim_{\Delta \mathbf{w} \rightarrow 0} \frac{(\mathbf{I} + [\Delta \mathbf{w}]_{\times}) \exp([\mathbf{w}]_{\times}) \mathbf{X}_{t} - \exp([\mathbf{w}]_{\times}) \mathbf{X}_{t}}{\Delta \mathbf{w}}  \\
		& = \lim_{\Delta \mathbf{w} \rightarrow 0} \frac{[\Delta \mathbf{w}]_{\times} \mathbf{R}\mathbf{X}_{t}}{\Delta \mathbf{w}} \quad \quad (\because \exp([\mathbf{w}]_{\times})\mathbf{X}_{t}  = \mathbf{R}\mathbf{X}_{t}) \\
		& = \lim_{\Delta \mathbf{w} \rightarrow 0} \frac{-[\mathbf{R}\mathbf{X}_{t}]_{\times} \Delta \mathbf{w}}{\Delta \mathbf{w}} \\
		& = -[\mathbf{R}\mathbf{X}_{t}]_{\times} \\ & = -[\mathbf{X}']_{\times} \\
\end{aligned}  \end{equation}

The second row in the above formula uses the approximation $\exp([\Delta \mathbf{w}]_{\times}) \approx  \mathbf{I} +  [\Delta \mathbf{w}]_{\times} $ for a small rotation matrix. \textbfazure{Therefore, using method $[2]$, there is an advantage that the Jacobian can be simply calculated using the skew-symmetric matrix of the 3D point $\mathbf{X}'$. In the case of reprojection error optimization, since the error of feature points in sequentially incoming images is optimized, the camera pose changes are not large, and thus $\Delta \mathbf{w}$ is also not large, so the above Jacobian is commonly used.} Using method $[2]$, the existing rotation matrix $\mathbf{R}$ is updated with a small increment $\Delta \mathbf{w}$ as in (\ref{eq:1}).
\begin{equation}
	\begin{aligned}
		\mathbf{R} \leftarrow \Delta \mathbf{R}^{*} \mathbf{R} \quad \text{where, } \Delta \mathbf{R}^{*} = \exp([\Delta \mathbf{w}^{*}]_{\times})
	\end{aligned}
\end{equation}

Therefore, the existing Jacobian changes from $\frac{\partial \mathbf{X}'}{\partial [\mathbf{w}, \mathbf{t}]}$ to $\frac{\partial \mathbf{X}'}{\partial [\Delta \mathbf{w}, \mathbf{t}]}$ and this is as follows.
\begin{equation}
	\boxed{
		\begin{aligned}
			\frac{\partial }{\partial [\Delta \mathbf{w}, \mathbf{t}]} \begin{bmatrix} \mathbf{RX} + \mathbf{t} \\ 1 \end{bmatrix} & = \begin{bmatrix} 0 & Z' & - Y' & 1 & 0 & 0
				\\ -Z' & 0 & X' & 0 & 1 & 0 \\ Y' & -X' & 0 & 0 & 0 & 1 \\ 0&0&0&0&0&0 \end{bmatrix} \in \mathbb{R}^{4 \times 6}
	\end{aligned} } 
\end{equation}

The final Jacobian of the pose $\mathbf{J}_{c}$ is as follows.
\begin{equation}
	\boxed{
		\begin{aligned}
			\mathbf{J}_{c} & = \frac{\partial \hat{\mathbf{p}}}{\partial \tilde{\mathbf{p}}} 
			\frac{\partial \tilde{\mathbf{p}}}{\partial \mathbf{X}'}
			\frac{\partial \mathbf{X}'}{\partial [\Delta \mathbf{w}, \mathbf{t}]} \\
			& = \begin{bmatrix} f & 0 & c_{x} \\ 0 & f & c_{y} \end{bmatrix} \begin{bmatrix} \frac{1}{Z'} & 0 & \frac{-X'}{Z'^{2}} & 0 \\
				0 & \frac{1}{Z'} & \frac{-Y'}{Z'^{2}} & 0  \\ 0 & 0 & 0 &0  \end{bmatrix}
			\begin{bmatrix} 0 & Z' & -Y' & 1 & 0 & 0 \\
				-Z' & 0 & X' & 0 & 1 & 0 \\
				Y' & -X' & 0 & 0 & 0 & 1 \\ 0&0&0&0&0&0 \end{bmatrix} \\
			& = \begin{bmatrix} -\frac{fX'Y'}{Z'^{2}} & \frac{f(1 + X'^{2})}{Z'^{2}} & -\frac{fY'}{Z'} & \frac{f}{Z'} & 0 & -\frac{fX'}{Z'^{2}} \\
				-\frac{f(1+y^{2})}{Z'^{2}} & \frac{fX'Y'}{Z'^{2}} & \frac{fX'}{Z'} & 0 & \frac{f}{Z'} & -\frac{fY'}{Z'^{2}} \end{bmatrix} \in \mathbb{R}^{2 \times 6}
	\end{aligned} } 
\end{equation}

\subsection{Jacobian of Map Point}
The Jacobian $\mathbf{J}_{p}$ of the 3D point $\mathbf{X}$ can be calculated as follows.
\begin{equation}
	\begin{aligned}
		\mathbf{J}_{p} = \frac{\partial \mathbf{e}}{\partial \mathbf{X}} & = \frac{\partial}{\partial \mathbf{X}}(\mathbf{p} - \hat{\mathbf{p}}) \\
		& = \frac{\partial}{\partial \mathbf{X}} \bigg(\mathbf{p} - \pi_{k}(\pi_{h}(\mathbf{T}_{i} \mathbf{X}_{j})) \bigg ) \\
		& = \frac{\partial}{\partial \mathbf{X}} \bigg(-\pi_{k}(\pi_{h}(\mathbf{T}_{i} \mathbf{X}_{j})) \bigg ) \\
	\end{aligned}
\end{equation}

Using the chain rule, the above formula is organized as follows.
\begin{equation}
	\begin{aligned}
		\mathbf{J}_{p}& = \frac{\partial \hat{\mathbf{p}}}{\partial  \tilde{\mathbf{p}}} 
		\frac{\partial \tilde{\mathbf{p}}}{\partial \mathbf{X}'}
		\frac{\partial \mathbf{X}'}{\partial \mathbf{X}} \\
		& = \mathbb{R}^{2 \times 3} \cdot \mathbb{R}^{3\times 4} \cdot \mathbb{R}^{4 \times 4} = \mathbb{R}^{2 \times 4}
	\end{aligned} 
\end{equation}

Among these, $\frac{\partial \hat{\mathbf{p}}}{\partial \tilde{\mathbf{p}}} \frac{\partial \hat{\mathbf{p}}}{\partial \mathbf{X}'}$ is the same as the Jacobian calculated earlier. Therefore, only $\frac{\partial \mathbf{X}'}{\partial \mathbf{X}}$ needs to be calculated.
\begin{equation}
	\boxed{
		\begin{aligned}
			\frac{\partial \mathbf{X}'}{\partial \mathbf{X}} & = \frac{\partial }{\partial \mathbf{X}} \begin{bmatrix} \mathbf{R}\mathbf{X} + \mathbf{t} \\ 1 \end{bmatrix}  \\
			& = \begin{bmatrix} \mathbf{R} \\ 0 \end{bmatrix}
	\end{aligned} }
\end{equation}

Therefore, $\mathbf{J}_{p}$ is as follows.
\begin{equation}
	\boxed{
		\begin{aligned}
			\mathbf{J}_{p} & = \begin{bmatrix} f & 0 & c_{x} \\ 0 & f & c_{y} \end{bmatrix} \begin{bmatrix} \frac{1}{Z'} & 0 & \frac{-X'}{Z'^{2}} & 0 \\
				0 & \frac{1}{Z'} & \frac{-Y'}{Z'^{2}} & 0\\ 0 & 0 & 0 & 0\end{bmatrix} \begin{bmatrix} \mathbf{R} \\ 0 \end{bmatrix} \\
			& = \begin{bmatrix} \frac{f}{Z'} & 0 & -\frac{fX'}{Z'^{2}} & 0 \\ 0 & \frac{f}{Z'} & -\frac{fY'}{Z'^{2}} & 0 \end{bmatrix} \begin{bmatrix} \mathbf{R} \\ 0 \end{bmatrix} \in \mathbb{R}^{2\times 4}
	\end{aligned} }
\end{equation}

Typically, the last column of $\mathbf{J}_{p}$ is always 0, so it is often omitted and represented in non-homogeneous form.
\begin{equation}
	\boxed{
		\begin{aligned}
			\mathbf{J}_{p} & = \begin{bmatrix} \frac{f}{Z'} & 0 & -\frac{fX'}{Z'^{2}} \\ 0 & \frac{f}{Z'} & -\frac{fY'}{Z'^{2}} \end{bmatrix} \mathbf{R} \in \mathbb{R}^{2\times 3}
	\end{aligned} }
\end{equation}

\subsection{Code Implementations}
\begin{itemize}
	\item g2o code: \href{https://github.com/RainerKuemmerle/g2o/blob/master/g2o/types/sba/edge_project_xyz.cpp#L80}{edge\_project\_xyz.cpp\#L80}
	\item g2o code: \href{https://github.com/RainerKuemmerle/g2o/blob/master/g2o/types/sba/edge_project_xyz.cpp#L82}{edge\_project\_xyz.cpp\#L82}
\end{itemize}

\section{Photometric Error}
Photometric error is primarily used in direct Visual SLAM. It is commonly utilized in direct method-based visual odometry (VO) or bundle adjustment (BA). For more detailed information on the direct method, refer to the post at \href{https://alida.tistory.com/52}{[SLAM] Optical Flow and Direct Method Concept and Code Review}.
~\\ ~\\
\textbf{NOMENCLATURE of Photometric Error}
\begin{itemize}
	\item $\tilde{\mathbf{p}}_{2} = \pi_{h}(\cdot) : \begin{bmatrix} X' \\ Y' \\ Z' \\1 \end{bmatrix} \rightarrow   \begin{bmatrix} X'/Z' \\ Y'/Z' \\ 1  \end{bmatrix}   = \begin{bmatrix} \tilde{u}_{2} \\ \tilde{v}_{2} \\ 1 \end{bmatrix}$
	\begin{itemize}
		\item The point $\mathbf{X}'$ in 3D space transformed to a non-homogeneous point on the image plane.
	\end{itemize}
	
	\item $\mathbf{p}_{2} = \pi_{k}(\cdot) = \tilde{\mathbf{K}} \tilde{\mathbf{p}}_{2} = \begin{bmatrix} f & 0 & c_{x} \\ 0 & f & c_{y} \end{bmatrix} \begin{bmatrix} \tilde{u}_{2} \\ \tilde{v}_{2} \\ 1 \end{bmatrix}= \begin{bmatrix} f\tilde{u} + c_{x} \\ f\tilde{v} + c_{y} \end{bmatrix} = \begin{bmatrix} u_{2} \\ v_{2} \end{bmatrix}$
	\begin{itemize}
		\item The point projected onto the image plane after correcting for lens distortion. If distortion correction has already been performed at the input stage, $\pi_{k}(\cdot) = \tilde{\mathbf{K}}(\cdot)$.
	\end{itemize}
	\item $\mathbf{K} = \begin{bmatrix} f & 0 & c_{x} \\ 0 & f & c_{y} \\ 0 & 0 & 1 \end{bmatrix}$: Camera's intrinsic parameters.
	\item $\tilde{\mathbf{K}} = \begin{bmatrix} f & 0 & c_{x} \\ 0 & f & c_{y}  \end{bmatrix}$: Omitting the last row of the intrinsic parameters for projection from $\mathbb{P}^{2} \rightarrow \mathbb{R}^{2}$.
	\item $\mathcal{P}$: Set of all feature points in the image.
	\item $\mathbf{e}(\mathbf{T}) \rightarrow \mathbf{e}$: Generally abbreviated for simplicity.
	\item $\mathbf{p}^{i}_{1}, \mathbf{p}^{i}_{2}$: Pixel coordinates of the ith feature point in the first and second images.
	\item $\oplus$ : Operator for combining two SE(3) groups (composition).
	\item $\mathbf{J} = \frac{\partial \mathbf{e}}{\partial \mathbf{T}} = \frac{\partial \mathbf{e}}{\partial [\mathbf{R}, \mathbf{t}]}$
	\item $\mathbf{X}' = [X,Y,Z,1]^{\intercal} = [\tilde{\mathbf{X'}}, 1]^{\intercal} = \mathbf{TX}$
	\item $\mathbf{T}\mathbf{X}$: Transformation, transforming the 3D point $\mathbf{X}$ into camera coordinates, $\bigg( \mathbf{T}\mathbf{X} = \begin{bmatrix} \mathbf{R} \mathbf{X} + \mathbf{t} \\ 1 \end{bmatrix}  \in \mathbb{R}^{4\times1} \bigg)$
	\item $\mathbf{X}' = [X',Y',Z',1]^{\intercal} = [\tilde{\mathbf{X}}', 1]^{\intercal}$
	\item $\xi = [\mathbf{w}, \mathbf{v}]^{\intercal} = [w_{x}, w_{y}, w_{z}, v_{x}, v_{y}, v_{z}]^{\intercal}$: Vector consisting of 3D angular velocity and velocity, called a twist.
	\item $[\xi]_{\times} = \begin{bmatrix} [\mathbf{w}]_{\times} & \mathbf{v} \\ \mathbf{0}^{\intercal} & 0 \end{bmatrix} \in \text{se}(3)$ : Lie algebra of the twist applied with the hat operator (4x4 matrix)
	\item $\mathcal{J}_{l}$: Jacobian for left multiplication. It is not used in actual calculations and hence not detailed here.
\end{itemize}
~\\
\begin{figure}[h!]
	\centering
	\includegraphics[width=8cm]{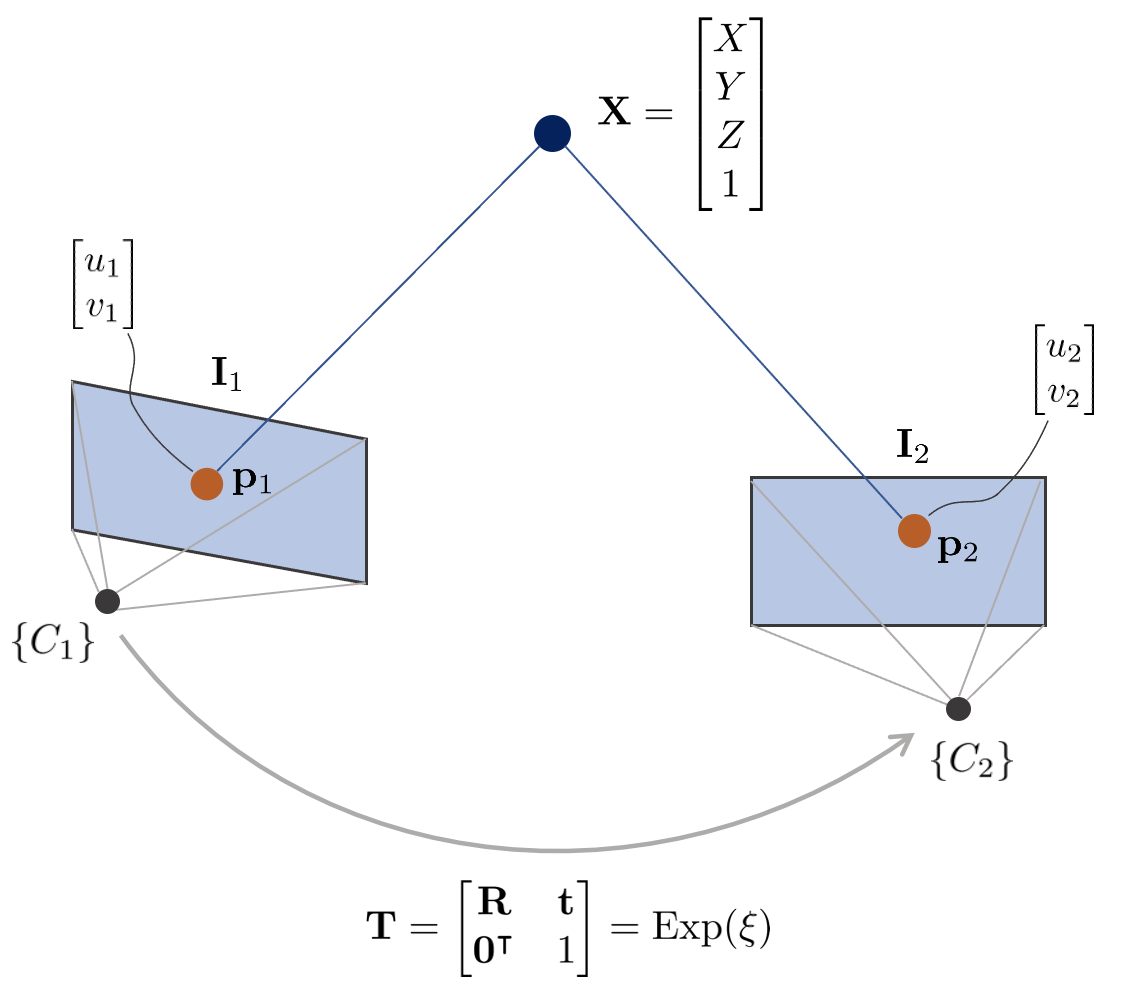}
\end{figure}
In the above figure, the world coordinates of the 3D point $\mathbf{X}$ are $[X,Y,Z,1]^{\intercal} \in \mathbb{P}^{3}$, and the corresponding pixel coordinates on the two camera image planes are $\mathbf{p}_{1}, \mathbf{p}_{2} \in \mathbb{P}^{2}$. Assuming the internal parameters $\mathbf{K}$ of the two cameras $\{C_{1}\}, \{C_{2}\}$ are the same. When camera $\{C_{1}\}$ is considered the origin ($\mathbf{R} = \mathbf{I}, \mathbf{t} = \mathbf{0}$), the pixel coordinates $\mathbf{p}_{1}, \mathbf{p}_{2}$ are projected through the 3D point $\mathbf{X}$ as follows:
\begin{equation}
	\begin{aligned} \mathbf{p} = \pi(\mathbf{T}, \mathbf{X})
	\end{aligned}
\end{equation} 

\begin{figure}[h!]
	\centering
	\includegraphics[width=16cm]{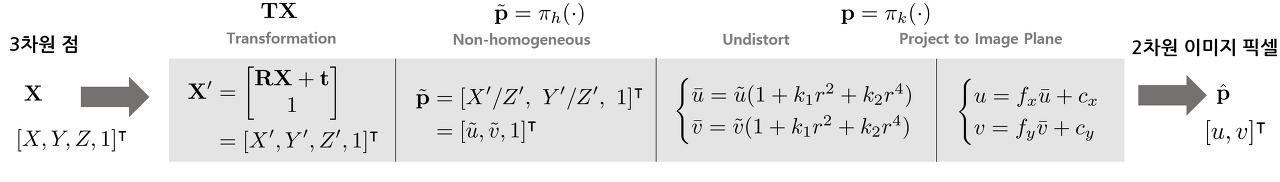}
\end{figure}
\begin{equation}
	\begin{aligned}
		\mathbf{p}_{1}  = \begin{pmatrix}
			u_{1}\\v_{1}\end{pmatrix}
		& = \pi(\mathbf{I}, \mathbf{X}) = \pi_{k}(\pi_{h}(\mathbf{X}))   \\
		\mathbf{p}_{2}  = \begin{pmatrix} u_{2}\\ v_{2}\end{pmatrix} & = \pi(\mathbf{T}, \mathbf{X}) = \pi_{k}(\pi_{h}(\mathbf{TX}))    \\ \end{aligned}
\end{equation} 

\textbfazure{One characteristic of the direct method, unlike feature-based methods, is the absence of a way to determine which $\mathbf{p}_{2}$ matches $\mathbf{p}_{1}$. Therefore, the position of $\mathbf{p}_{2}$ is found based on the current pose estimate.} Thus, the camera's pose is optimized to make $\mathbf{p}_{2}$ and $\mathbf{p}_{1}$ similar, and this problem is solved by minimizing the photometric error. The photometric error is as follows:
\begin{equation}
	\boxed{
		\begin{aligned}
			\mathbf{e}(\mathbf{T}) & =  \mathbf{I}_{1}(\mathbf{p}_{1})-\mathbf{I}_{2}(\mathbf{p}_{2}) \\
			& = \mathbf{I}_{1} \bigg( \pi_{k}(\pi_{h}(\mathbf{X})) \bigg)-\mathbf{I}_{2}\bigg(  \pi_{k}(\pi_{h}(\mathbf{TX})) \bigg) \\
	\end{aligned} } \label{eq:photo3}
\end{equation}

Photometric error is based on the assumption of grayscale invariance and holds scalar values. The following error function $\mathbf{E}(\mathbf{T})$ can be defined to solve non-linear least squares:
\begin{equation}
	\begin{aligned}
		\mathbf{E}(\mathbf{T}) & =   \sum_{i \in \mathcal{P}} \left\| \mathbf{e}_{i} \right\|^{2} \\
	\end{aligned}
\end{equation}
\begin{equation}
	\begin{aligned}
		\mathbf{T}^{*}  & =  \arg\min_{\mathbf{T}^{*}} \mathbf{E}(\mathbf{T}) \\
		& =  \arg\min_{\mathbf{T}^{*}} \sum_{i \in \mathcal{P}} \left\| \mathbf{e}_{i} \right\|^{2} \\
		& = \arg\min_{\mathbf{T}^{*}} \sum_{i \in \mathcal{P}} \mathbf{e}_{i}^{\intercal}\mathbf{e}_{i} \\
		& = \arg\min_{\mathbf{T}^{*}} \sum_{i in \mathcal{P}} \Big( \mathbf{I}_{1}(\mathbf{p}^{i}_{1})-\mathbf{I}_{2}(\mathbf{p}^{i}_{2}) \Big)^{\intercal} \Big( \mathbf{I}_{1}(\mathbf{p}^{i}_{1})-\mathbf{I}_{2}(\mathbf{p}^{i}_{2}) \Big)
	\end{aligned}
\end{equation}

$\mathbf{E}(\mathbf{T}^{*})$ that satisfies $\left\|\mathbf{e}(\mathbf{T}^{*})\right\|^{2}$ can be calculated iteratively through non-linear least squares. Small increments $\Delta \mathbf{T}$ are iteratively updated to $\mathbf{T}$ to find the optimal state.
\begin{equation}
	\begin{aligned}
		\arg\min_{\mathbf{T}^{*}} \mathbf{E}(\mathbf{T} + \Delta \mathbf{T}) & = \arg\min_{\mathbf{T}^{*}} \sum_{i \in \mathcal{P}} \left\|\mathbf{e}_{i}(\mathbf{T} + \Delta \mathbf{T})\right\|^{2} 
	\end{aligned}
\end{equation}

Technically, since the state increment $\Delta \mathbf{T}$ is a SE(3) transformation matrix, it should be added to the existing state $\mathbf{T}$ using the $\oplus$ operator, but the $+$ operator is used here for convenience of expression.
\begin{equation}
	\begin{aligned}
		\mathbf{T}  \oplus \Delta \mathbf{T}  \quad  \rightarrow \quad \mathbf{T} + \Delta \mathbf{T}
	\end{aligned}
\end{equation}

It is expressed through the first-order Taylor approximation as follows.
\begin{equation}
	\begin{aligned}
		\mathbf{e}(\mathbf{T} + \Delta \mathbf{T}) & \approx \mathbf{e}_{i}(\mathbf{T}) + \mathbf{J}\Delta \mathbf{T}  \\
		& = \mathbf{e}_{i}(\mathbf{T}) + \frac{\partial \mathbf{e}}{\partial \mathbf{T}}\Delta \mathbf{T}
	\end{aligned}
\end{equation}

\begin{equation}
	\begin{aligned}
		\arg\min_{\mathbf{T}^{*}}  \mathbf{E}(\mathbf{T} + \Delta \mathbf{T}) & = \arg\min_{\mathbf{T}^{*}} \sum_{i \in \mathcal{P}} \left\|\mathbf{e}_{i}(\mathbf{T}) + \mathbf{J}\Delta \mathbf{T}\right\|^{2} 
	\end{aligned}
\end{equation}

When differentiating to find the optimal increment $\Delta \mathbf{T}^{*}$, the following results. The detailed derivation process is omitted in this section. If you want to know more about the derivation process, refer to the previous section \hyperref[sec:opt]{here}.
\begin{equation}
	\begin{aligned}
		& \mathbf{J}^{\intercal}\mathbf{J} \Delta \mathbf{T}^{*} = -\mathbf{J}^{\intercal}\mathbf{e} \\ 
		& \mathbf{H}\Delta \mathbf{T}^{*} = - \mathbf{b} \\
	\end{aligned} \label{eq:photo1}
\end{equation}

Since the above formula forms a linear system $\mathbf{Ax} = \mathbf{b}$, various linear algebra techniques such as schur complement, cholesky decomposition can be used to find $\Delta \mathbf{T}^{*}$. The optimal increment found in this way is then added to the current state. \textbfazure{Depending on whether it is multiplied to the right or left of the existing state $\mathbf{T}$, it changes whether to update the pose in the local coordinate system (right) or the pose in the global coordinate system (left). Since photometric error updates the transformation matrix of the global coordinate system, the left multiplication method is generally used.}
\begin{equation}
	\begin{aligned}
		\mathbf{T} \leftarrow  \mathbf{T} \oplus \Delta \mathbf{T}^{*}
	\end{aligned} 
\end{equation}

The definition of the left multiplication $\oplus$ operation is as follows.
\begin{equation}
	\begin{aligned}
		\mathbf{T} \oplus \Delta \mathbf{T}^{*}& = \Delta \mathbf{T}^{*}\mathbf{T} \\ & = \exp([\Delta \xi^{*}]_{\times}) \mathbf{T}  \quad \cdots \text{ globally updated (left mult)} \end{aligned} \label{eq:photo2}
\end{equation}

\subsection{Jacobian of the Photometric Error}
To perform (\ref{eq:photo1}), the Jacobian $\mathbf{J}$ of the photometric error must be determined. It can be represented as follows.
\begin{equation}
	\begin{aligned}
		\mathbf{J} & = \frac{\partial \mathbf{e}}{\partial \mathbf{T}}   \\
		& =  \frac{\partial \mathbf{e}}{\partial [\mathbf{R}, \mathbf{t}]} 
	\end{aligned}
\end{equation}

Expanding this in detail results in the following.
\begin{equation}
	\begin{aligned}
		\mathbf{J} = \frac{\partial \mathbf{e}}{\partial \mathbf{T}} & = \frac{\partial }{\partial \mathbf{T}} \bigg( \mathbf{I}_{1}(\mathbf{p}_{1}) - \mathbf{I}_{2}(\mathbf{p}_{2}) \bigg) \\ & = \frac{\partial }{\partial \mathbf{T}} \bigg(
		\mathbf{I}_{1} \bigg(    \pi_{k}(\pi_{h}(\mathbf{X}))   \bigg)-\mathbf{I}_{2}\bigg( \pi_{k}(\pi_{h}(\mathbf{TX})) \bigg) \bigg) \\
		& =  \frac{\partial }{\partial \mathbf{T}} \bigg( -\mathbf{I}_{2}\bigg( \pi_{k}(\pi_{h}(\mathbf{TX})) \bigg) \bigg) \\ & =  \frac{\partial }{\partial \mathbf{T}} \bigg( -\mathbf{I}_{2}\bigg( \pi_{k}(\pi_{h}(\mathbf{X}')) \bigg) \bigg)
	\end{aligned}
\end{equation}

Applying the chain rule re-expresses the above equation as follows.
\begin{equation}
	\begin{aligned}
		\frac{\partial \mathbf{e}}{\partial \xi} & =  \frac{\partial \mathbf{I}}{\partial \mathbf{p}_{2}} \frac{\partial \mathbf{p}_{2}}{\partial \tilde{\mathbf{p}}_{2}} \frac{\partial \tilde{\mathbf{p}}_{2}}{\partial \mathbf{X}'} \frac{\partial \mathbf{X}'}{\partial \xi}  \\
		& = \mathbb{R}^{1\times2} \cdot \mathbb{R}^{2\times3} \cdot \mathbb{R}^{3\times4} \cdot \mathbb{R}^{4\times6} = \mathbb{R}^{1\times6} 
	\end{aligned}
\end{equation}

\textbfazure{The reason for computing the Jacobian $\frac{\partial \mathbf{X}'}{\partial \xi}$ instead of $\frac{\partial \mathbf{X}'}{\partial \mathbf{T}}$ will be explained in the next section.} First, $\frac{\partial \mathbf{I}}{\partial \mathbf{p}_{2}}$ refers to the gradient of the image.
\begin{equation}
	\boxed{
		\begin{aligned}
			\frac{\partial \mathbf{I}}{\partial \mathbf{p}_{2}}  & =  \begin{bmatrix} \frac{\partial \mathbf{I}}{\partial u} & \frac{\partial \mathbf{I}}{\partial v} \end{bmatrix} \\
			& = \begin{bmatrix} \nabla \mathbf{I}_{u} & \nabla \mathbf{I}_{v} \end{bmatrix}
	\end{aligned} } \label{eq:photo4}
\end{equation}

If it is assumed that undistortion was already performed during image input, $\frac{\partial \mathbf{p}_{2}}{\partial \tilde{\mathbf{p}}_{2}}$ is as follows.
\begin{equation}
	\boxed{
		\begin{aligned}
			\frac{\partial \mathbf{p}_{2}}{\partial \tilde{\mathbf{p}}_{2}} & = 
			\frac{\partial }{\partial \tilde{\mathbf{p}}_{2}} \tilde{\mathbf{K}} \tilde{\mathbf{p}}_{2} \\
			& = \tilde{\mathbf{K}} \\
			& = \begin{bmatrix} f & 0 & c_{x} \\ 0 & f & c_{y} \end{bmatrix} \in \mathbb{R}^{2 \times 3}
	\end{aligned} }
\end{equation}

Next, $\frac{\partial \tilde{\mathbf{p}}_{2}}{\partial \mathbf{X}'}$ is as follows.
\begin{equation}
	\boxed{
		\begin{aligned}
			\frac{\partial \tilde{\mathbf{p}}_{2}}{\partial \mathbf{X}'} & = 
			\frac{\partial [\tilde{u}_{2}, \tilde{v}_{2}, 1]}{\partial [X', Y', Z',1]} \\
			& = \begin{bmatrix} \frac{1}{Z'} & 0 & \frac{-X'}{Z'^{2}} & 0\\
				0 & \frac{1}{Z'} & \frac{-Y'}{Z'^{2}} &0 \\ 0 & 0 & 0 & 0 \end{bmatrix} \in \mathbb{R}^{3\times 4}
	\end{aligned} } \label{eq:photo5}
\end{equation}

\subsubsection{Lie Theory-based SE(3) Optimization}
Finally, $\frac{\partial \mathbf{X}'}{\partial \mathbf{T}} = \frac{\partial \mathbf{X}'}{\partial [\mathbf{R}, \mathbf{t}]}$ must be computed. At this time, the term related to position $\mathbf{t}$ is a 3D vector, and the size of this vector is the minimum degree of freedom, 3 degrees of freedom, for representing 3D position, so there is no separate constraint when performing optimization updates. \textbfazure{On the other hand, the rotation matrix $\mathbf{R}$ has 9 parameters, which is more than the minimum degrees of freedom, 3 degrees of freedom, for representing 3D rotation, so various constraints exist. This is called being over-parameterized. The disadvantages of over-parameterized representation are as follows.}
\begin{itemize}
	\item It is necessary to calculate redundant parameters, which increases the computation during optimization.
	\item Additional degrees of freedom can cause numerical instability.
	\item It is necessary to check whether the constraints are satisfied each time the parameters are updated.
\end{itemize}

Therefore, the optimization method based on lie theory, which is free from constraints, is generally used. \textbfazure{The lie group SE(3) based optimization method refers to the method of updating SE(3) by finding the optimal twist $\Delta \xi^{*}$ after changing the term related to rotation from $\mathbf{R} \rightarrow \mathbf{w}$ and the term related to position from $\mathbf{t} \rightarrow \mathbf{v}$, and then updating SE(3) through exponential mapping of lie algebra se(3) $[\Delta \xi]_{\times}$.}
\begin{equation}
	\begin{aligned}
		\Delta \mathbf{T}^{*} & \rightarrow \Delta \xi^{*} 
	\end{aligned}
\end{equation}

The Jacobian $\xi$ is as follows.
\begin{equation}
	\begin{aligned}
		\mathbf{J}  & =  \frac{\partial \mathbf{e}}{\partial [\mathbf{R}, \mathbf{t}]} && \rightarrow  \frac{\partial \mathbf{e}}{\partial [\mathbf{w}, \mathbf{v}]}  \\
		& && \rightarrow  \frac{\partial \mathbf{e}}{\partial \xi} 
	\end{aligned}
\end{equation}

The existing equation is changed as follows through this.
\begin{equation}
	\begin{aligned}
		\mathbf{e}(\mathbf{T}) \quad \quad & \rightarrow \quad \quad \mathbf{e}(\xi) \\
		\mathbf{E}(\mathbf{T}) \quad \quad & \rightarrow \quad \quad \mathbf{E}(\xi) \\
		\mathbf{e}(\mathbf{T}) + \mathbf{J}'\Delta \mathbf{T} \quad \quad & \rightarrow \quad \quad \mathbf{e}(\xi) + \mathbf{J}\Delta \xi \\
		\mathbf{H}\Delta\mathbf{T}^{*} = -\mathbf{b} \quad \quad & \rightarrow \quad \quad \mathbf{H}\Delta \xi^{*}  = -\mathbf{b} \\
		\mathbf{T} \leftarrow \Delta \mathbf{T}^{*}\mathbf{T} \quad \quad & \rightarrow \quad \quad \mathbf{T} \leftarrow \exp([\Delta \xi^{*}]_{\times}) \mathbf{T} 
	\end{aligned}
\end{equation}
- $\mathbf{J}'  = \frac{\partial \mathbf{e}}{\partial \mathbf{T}}$ \\
- $\mathbf{J}  = \frac{\partial \mathbf{e}}{\partial \xi}$ \\

\mybox{Tip}{gray!40}{gray!10}{
	$\exp([\xi]_{\times}) \in \text{SE}(3)$ refers to the operation of transforming the twist $\xi$ through exponential mapping into a 3D pose. For more details on exponential mapping, refer to the \href{https://alida.tistory.com/60\#2.3.2-the-capitalized-exponential-map}{related link}.
	\begin{equation}
		\begin{aligned}
			\exp([\Delta \xi]_{\times}) = \Delta \mathbf{T} 
		\end{aligned}
	\end{equation}
}

Until now, the Jacobians were easy to calculate, whereas $\frac{\partial \mathbf{X}'}{\partial \xi}$ requires changing $\mathbf{X}'$ into a term related to lie algebra as it is not immediately apparent from $\mathbf{X}'$ parameters $\xi$.
\begin{equation}
	\begin{aligned}
		\mathbf{X}' \rightarrow \mathbf{TX} \rightarrow \exp([\xi]_{\times})\mathbf{X}
	\end{aligned}
\end{equation}

At this time, depending on the update method of the small lie algebra increment $\Delta \xi$ to the existing $\exp([\xi]_{\times})$, it is divided into two methods. First, there is [1] the basic update method using lie algebra. Next, there is [2] the update method using the perturbation model.
\begin{equation}
	\begin{aligned}
		\exp([\xi]_{\times}) & \leftarrow \exp([\xi + \Delta \xi]_{\times}) \quad \cdots \text{[1]} \\ 
		\exp([\xi]_{\times}) &  \leftarrow \exp([\Delta \xi]_{\times})\exp([\xi]_{\times}) \quad \cdots \text{[2]} 
	\end{aligned}
\end{equation}

Among the two methods, method $[1]$ is a method of adding a fine increment $\Delta \xi$ to the existing $\xi$ and performing exponential mapping to obtain the Jacobian, while method $[2]$ is a method of updating the existing state by multiplying the perturbation model $\exp([\Delta \xi]_{\times})$ to the left of the existing $\xi$.

\mybox{Tip}{gray!40}{gray!10}{
	The following transformation exists between the two methods, known as the BCH approximation. For more details, refer to \href{https://docs.google.com/document/d/1icPjUyT3nPvjZ1OVMtWp9afUtuJ4gXLJL-ex7A9FpNs/edit?fbclid=IwAR2VfhZ3js52zkFpZpJ5HZv_qQLPz7WCTBWkwn6IF1MmHa3Ksyhi5TQSAfY\#}{Introduction to Visual SLAM Chapter 4}.
	\begin{equation}
		\begin{aligned}
			\exp([\Delta \xi]_{\times})\exp([\xi]_{\times}) & = \exp([\xi + \mathcal{J}^{-1}_{l}\Delta \xi]_{\times}) \\
			\exp([\xi + \Delta \xi]_{\times}) & = \exp([\mathcal{J}_{l}\Delta \xi]_{\times})\exp([\xi]_{\times}) \\
		\end{aligned} \label{eq:bch}
	\end{equation}
}

\textbfazure{Since a very complex equation is derived when using method $[1]$, this method is not commonly used and method $[2]$ of the perturbation model is mainly used. Therefore, $\frac{\partial \mathbf{X}'}{\partial \xi}$ is transformed as follows.}
\begin{equation}
	\begin{aligned}
		\frac{\partial \mathbf{X}'}{\partial \xi} & \rightarrow \frac{\partial \mathbf{X}'}{\partial \Delta \xi}
	\end{aligned}
\end{equation}

The Jacobian for $\frac{\partial \mathbf{X}'}{\partial \Delta \xi}$ can be calculated as follows.
\begin{equation} \boxed{ \begin{aligned}
			\frac{\partial \mathbf{X}'}{\partial \Delta \xi} & = \lim_{\Delta \xi \rightarrow 0} \frac{\exp([\Delta \xi]_{\times}) \mathbf{X}' - \mathbf{X}'}{\Delta \xi} \\
			& \approx \lim_{\Delta \xi \rightarrow 0} \frac{(\mathbf{I} + [\Delta \xi]_{\times})\mathbf{X}' - \mathbf{X}'}{\Delta \xi} \\
			& = \lim_{\Delta \xi \rightarrow 0} \frac{[\Delta \xi]_{\times}\mathbf{X}'}{\Delta \xi} \\
			& = \lim_{\Delta \xi \rightarrow 0} \frac{\begin{bmatrix} [\Delta \mathbf{w}]_{\times} & \Delta \mathbf{v} \\ \mathbf{0}^{\intercal} & 0 \end{bmatrix} \begin{bmatrix} \tilde{\mathbf{X}}' \\ 1 \end{bmatrix}}{\Delta \xi} \\
			& = \lim_{\Delta \xi \rightarrow 0} \frac{\begin{bmatrix} [\Delta \mathbf{w}]_{\times} \tilde{\mathbf{X}}' + \Delta \mathbf{v} \\ \mathbf{0}^{\intercal} \end{bmatrix}}{[\Delta \mathbf{w}, \Delta \mathbf{v}]^{\intercal}}  = \begin{bmatrix} -[\tilde{\mathbf{X}}']_{\times} & \mathbf{I}  \\ \mathbf{0}^{\intercal} & \mathbf{0}^{\intercal} \end{bmatrix}  \in \mathbb{R}^{4 \times 6}
\end{aligned} } \end{equation}

\textbfazure{Therefore, using method $[2]$ of the perturbation model has the advantage of simplifying the Jacobian calculation using the skew-symmetric matrix of the 3D point $\mathbf{X}'$. Since photometric error optimization generally involves optimizing the error in brightness changes in sequentially incoming images, the camera pose changes are not large, and thus $\Delta \xi$ is also not large, so the above Jacobian is commonly used.} Method $[2]$ of the perturbation model is used, so the small increment $\Delta \xi^{*}$ is updated as (\ref{eq:photo2}).
\begin{equation}
	\begin{aligned}
		\mathbf{T} \leftarrow \Delta \mathbf{T}^{*}\mathbf{T}= \exp([\Delta \xi^{*}]_{\times}) \mathbf{T}
	\end{aligned} 
\end{equation}

\mybox{Tip}{gray!40}{gray!10}{
	The second row of the above equation is a form where the first-order Taylor approximation is applied to a small twist increment $\exp([\Delta \xi]_{\times})$. To understand the approximation in the second row, when an arbitrary twist $\xi = [\mathbf{w}, \mathbf{v}]^{\intercal}$ is given, the transformation matrix $\mathbf{T}$ can be expanded into an exponential mapping form as follows.
	\begin{equation}
		\begin{aligned}
			\mathbf{T} = \exp([\xi]_{\times}) & =  \mathbf{I} +  \begin{bmatrix} [\mathbf{w}]_{\times} & \mathbf{v} \\ \mathbf{0}^{\intercal} & 0 \end{bmatrix}  + \frac{1}{2!} \begin{bmatrix} [\mathbf{w}]_{\times}^{2} & [\mathbf{w}]_{\times}\mathbf{v} \\ \mathbf{0}^{\intercal} & 0 \end{bmatrix} + \frac{1}{3!} \begin{bmatrix} [\mathbf{w}]_{\times}^{3} & [\mathbf{w}]_{\times}^{2}\mathbf{v} \\ \mathbf{0}^{\intercal} & 0 \end{bmatrix}  + \cdots \\ 
			& = \mathbf{I} + [ \xi]_{\times} + \frac{1}{2!} [ \xi]_{\times}^{2} +  \frac{1}{3!} [ \xi]_{\times}^{3} + \cdots
		\end{aligned}
	\end{equation}
	
	For a small magnitude of twist increment $\Delta \xi$, higher-order terms can be ignored to approximately express it as follows.
	\begin{equation}
		\begin{aligned}
			\exp([\Delta \xi]_{\times}) \approx  \mathbf{I} +  [\Delta \xi]_{\times} 
		\end{aligned}
	\end{equation}
}

The final Jacobian $\mathbf{J}$ for the pose is as follows.
\begin{equation}
	\boxed{
		\begin{aligned}
			\mathbf{J} = \frac{\partial \mathbf{e}}{\partial \Delta \xi} & =  \frac{\partial \mathbf{I}}{\partial \mathbf{p}_{2}} \frac{\partial \mathbf{p}_{2}}{\partial \tilde{\mathbf{p}}_{2}} \frac{\partial \tilde{\mathbf{p}}_{2}}{\partial \mathbf{X}'} \frac{\partial \mathbf{X}'}{\partial \Delta \xi}  \\
			& = \begin{bmatrix} \nabla \mathbf{I}_{u} & \nabla \mathbf{I}_{v} \end{bmatrix} 
			\begin{bmatrix} f & 0 & c_{x} \\ 0 & f & c_{y} \end{bmatrix}
			\begin{bmatrix} \frac{1}{Z'} & 0 & \frac{-X'}{Z'^{2}} & 0 \\ 0 & \frac{1}{Z'} & \frac{-Y'}{Z'^{2}}&0\\ 0 & 0 & 0 & 0\end{bmatrix}
			\begin{bmatrix} -[\tilde{\mathbf{X}}']_{\times} & \mathbf{I} \\ \mathbf{0}^{\intercal} & \mathbf{0}^{\intercal}  \end{bmatrix} \\
			& = \begin{bmatrix} \nabla \mathbf{I}_{u} & \nabla \mathbf{I}_{v} \end{bmatrix}  \begin{bmatrix} -\frac{fX'Y'}{Z'^{2}} & \frac{f(1 + X'^{2})}{Z'^{2}} & -\frac{fY'}{Z'} & \frac{f}{Z'} & 0 & -\frac{fX'}{Z'^{2}} \\
				-\frac{f(1+Y'^{2})}{Z'^{2}} & \frac{fX'Y'}{Z'^{2}} & \frac{fX'}{Z'} & 0 & \frac{f}{Z'} & -\frac{fY'}{Z'^{2}} \end{bmatrix} \in \mathbb{R}^{1 \times 6}
	\end{aligned} }
\end{equation}

Since the last row of $\frac{\partial \mathbf{X}'}{\partial \Delta \xi}$ is always 0, it is often omitted and calculated.

\subsection{Code Implementations}
\begin{itemize}
	\item Introduction to Visual SLAM Chapter 8 code: \href{https://github.com/gaoxiang12/slambook/blob/master/ch8/directMethod/direct_sparse.cpp#L111}{direct\_sparse.cpp\#L111}
	\item DSO code: \href{https://github.com/JakobEngel/dso/blob/master/src/FullSystem/CoarseInitializer.cpp#L430}{CoarseInitializer.cpp\#L430}
	\item DSO code2: \href{https://github.com/JakobEngel/dso/blob/master/src/FullSystem/CoarseTracker.cpp#L320}{CoarseTracker.cpp\#L320}
\end{itemize}

\section{Relative pose error}
Relative pose error is commonly used in pose graph optimization (PGO). For more information about PGO, refer to the post \href{https://alida.tistory.com/16}{[SLAM] Conceptual explanation and example code analysis of Pose Graph Optimization}.
~\\ ~\\
\textbf{NOMENCLATURE of relative pose error}
\begin{itemize}
	\item $\text{(Node) } \mathbf{x}_{i}  = \begin{bmatrix} \mathbf{R}_{i} & \mathbf{t}_{i} \\ \mathbf{0}^{\intercal} & 1 \end{bmatrix} \in \mathbb{R}^{4\times 4}$
	\item $\text{(Edge) } \mathbf{z}_{ij}  = \begin{bmatrix} \mathbf{R}_{ij} & \mathbf{t}_{ij} \\ \mathbf{0}^{\intercal} & 1 \end{bmatrix} \in \mathbb{R}^{4\times 4}$
	\item $\hat{\mathbf{z}}_{ij}   = \ \mathbf{x}_{i}^{-1}\mathbf{x}_{j}$ : Predicted value
	\item $\mathbf{z}_{ij}$ : Observed value (virtual measurement)
	\item $\mathbf{x} = [\mathbf{x}_{1}, \cdots, \mathbf{x}_{n}]$: All pose nodes in the pose graph
	\item $\mathbf{e}_{ij}(\mathbf{x}_{i},\mathbf{x}_{j}) \leftrightarrow \mathbf{e}_{ij}$: Notation is simplified for convenience.
	\item $\mathbf{J} = \frac{\partial \mathbf{e}}{\partial \mathbf{x}}$
	\item $\oplus$ : Operator that combines two SE(3) groups (composition)
	\item $\text{Log}(\cdot)$: Operator that transforms SE(3) into a twist $\xi \in \mathbb{R}^{6}$. For detailed information about Logarithm mapping, refer to \href{https://alida.tistory.com/9#org608a5f4-1}{this post}.
\end{itemize}

\begin{figure}[h!]
	\centering
	\includegraphics[width=10cm]{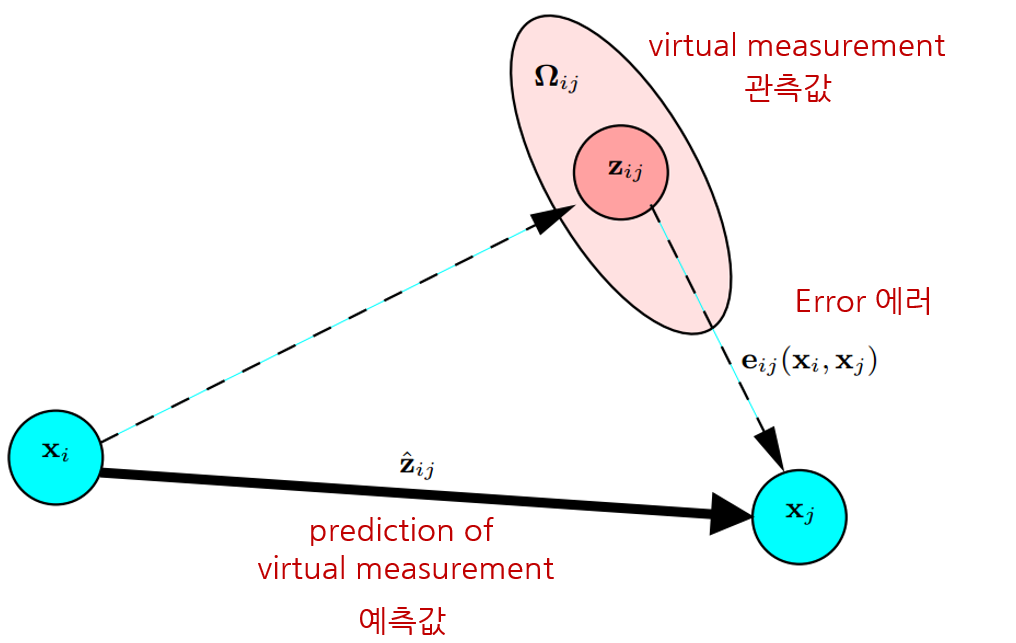}
\end{figure}
When two nodes $\mathbf{x}_{i}, \mathbf{x}_{j}$ are given on the pose graph, the difference between the newly calculated relative pose (observed value) $\mathbf{z}_{ij}$ and the known relative pose (predicted value) $\hat{\mathbf{z}}_{ij}$ is defined as the relative pose error. (Refer to the figure from Freiburg Univ. Robot Mapping Course).
\begin{equation}
	\begin{aligned}
		\mathbf{e}_{ij}(\mathbf{x}_{i},\mathbf{x}_{j}) = \mathbf{z}_{ij}^{-1}\hat{\mathbf{z}}_{ij} = \mathbf{z}_{ij}^{-1} \mathbf{x}_{i}^{-1}\mathbf{x}_{j}
	\end{aligned}
\end{equation}

\textbfazure{The process of optimizing the relative pose error is called pose graph optimization (PGO), and it is also known as the back-end algorithm of graph-based SLAM.} The nodes $\mathbf{x}_{i}, \mathbf{x}_{i+1}, \cdots$, sequentially calculated by the front-end visual odometry (VO) or lidar odometry (LO), do not undergo PGO because the observed and predicted values are the same. However, when loop closing occurs and a non-sequential edge connects two nodes $\mathbf{x}_{i}, \mathbf{x}_{j}$, a difference between the observed and predicted values arises, leading to the execution of PGO.

\textbfazure{In other words, PGO is typically performed when special situations like loop closing occur.} When the robot revisits the same location while moving, a loop detection algorithm operates to determine the loop. At this time, if a loop is detected, the existing node $\mathbf{x}_{i}$ and the node $\mathbf{x}_{j}$ created by revisiting are connected by a loop edge, and an observed value is produced by various matching algorithms (GICP, NDT, etc...). \textbfazure{Such observed values, not actually observed but created by matching algorithms, are called virtual measurements.}

The relative pose error for all nodes on the pose graph can be defined as follows.
\begin{equation}
	\begin{aligned}
		\mathbf{E}(\mathbf{x}) & =  \sum_{i}\sum_{j} \left\| \mathbf{e}_{ij} \right\|^{2} \\
	\end{aligned}
\end{equation}
\begin{equation}
	\begin{aligned}
		\mathbf{x}^{*} & = \arg\min_{\mathbf{x}^{*}} \mathbf{E}(\mathbf{x}) \\ 
		& =  \arg\min_{\mathbf{x}^{*}} \sum_{i}\sum_{j} \left\| \mathbf{e}_{ij} \right\|^{2} \\
		& = \arg\min_{\mathbf{x}^{*}} \sum_{i}\sum_{j} \mathbf{e}_{ij}^{\intercal}\mathbf{e}_{ij} 
	\end{aligned}
\end{equation}

$\mathbf{E}(\mathbf{x}^{*})$ can be calculated iteratively through non-linear least squares by updating a small increment $\Delta \mathbf{x}$ to $\mathbf{x}$ repeatedly to find the optimal state.
\begin{equation}
	\begin{aligned}
		\arg\min_{\mathbf{x}^{*}} \mathbf{E}(\mathbf{x} + \Delta \mathbf{x}) & = \arg\min_{\mathbf{x}^{*}} \sum_{i}\sum_{j} \left\|\mathbf{e}_{ij}(\mathbf{x}_{i} + \Delta \mathbf{x}_{i},\mathbf{x}_{j} + \Delta \mathbf{x}_{j})\right\|^{2} 
	\end{aligned}
\end{equation}

Technically speaking, since the state increment $\Delta \mathbf{x}$ is an SE(3) transformation matrix, it should be added to the existing state $\mathbf{x}$ through the $\oplus$ operator, but for convenience of expression, the $+$ operator is used.
\begin{equation}
	\begin{aligned}
		\mathbf{e}_{ij}(\mathbf{x}_{i} \oplus \Delta \mathbf{x}_{i},\mathbf{x}_{j} \oplus \Delta \mathbf{x}_{j}) 
		\quad \rightarrow \quad\mathbf{e}_{ij}(\mathbf{x}_{i} + \Delta \mathbf{x}_{i},\mathbf{x}_{j} + \Delta \mathbf{x}_{j})
	\end{aligned}
\end{equation}

This equation can be expressed through a first-order Taylor approximation as follows.
\begin{equation}
	\begin{aligned}
		\mathbf{e}_{ij}(\mathbf{x}_{i} + \Delta \mathbf{x}_{i},\mathbf{x}_{j} + \Delta \mathbf{x}_{j})  & \approx \mathbf{e}_{ij}(\mathbf{x}_{i} ,\mathbf{x}_{j} ) + \mathbf{J}_{ij} \begin{bmatrix} \Delta \mathbf{x}_{i} \\ \Delta \mathbf{x}_{j} \end{bmatrix} \\ & = \mathbf{e}_{ij}(\mathbf{x}_{i} ,\mathbf{x}_{j} ) + \mathbf{J}_{i}\Delta \mathbf{x}_{i} + \mathbf{J}_{j}\Delta \mathbf{x}_{j} \\
		& = \mathbf{e}_{ij}(\mathbf{x}_{i} ,\mathbf{x}_{j} ) + \frac{\partial \mathbf{e}_{ij}}{\partial \mathbf{x}_{i}} \Delta \mathbf{x}_{i} + + \frac{\partial \mathbf{e}_{ij}}{\partial \mathbf{x}_{j}} \Delta \mathbf{x}_{j} 
	\end{aligned} \label{eq:rel14}
\end{equation}

\begin{equation}
	\begin{aligned}
		\arg\min_{\mathbf{x}^{*}} \mathbf{E}(\mathbf{x} + \Delta \mathbf{x}) & \approx \arg\min_{\mathbf{x}^{*}} \sum_{i}\sum_{j} \left\|\mathbf{e}_{ij}(\mathbf{x}_{i},\mathbf{x}_{j}) + \mathbf{J}_{ij} \begin{bmatrix} \Delta \mathbf{x}_{i} \\ \Delta \mathbf{x}_{j} \end{bmatrix} \right\|^{2} \\
	\end{aligned}
\end{equation}

Differentiating this to find the optimal increment $\Delta \mathbf{x}^{*}$ for all nodes results in the following. The derivation process is omitted in this section. If you want to know the detailed derivation process, refer to the \hyperref[sec:opt]{previous section}.
\begin{equation}
	\begin{aligned}
		& \mathbf{J}^{\intercal}\mathbf{J} \Delta \mathbf{x}^{*} = -\mathbf{J}^{\intercal}\mathbf{e} \\ 
		& \mathbf{H}\Delta \mathbf{x}^{*} = - \mathbf{b} \\
	\end{aligned} \label{eq:rel6}
\end{equation}

This equation forms a linear system $\mathbf{Ax} = \mathbf{b}$, and the optimal increment $\Delta \mathbf{x}^{*}$ can be found using various linear algebra techniques such as the schur complement and Cholesky decomposition. The obtained optimal increment is then added to the current state. Depending on whether it is multiplied on the right or the left of the existing state $\mathbf{x}$, it updates the pose viewed from the local coordinate system (right) or the global coordinate system (left). Since the relative pose error is related to the relative pose of the two nodes, right multiplication, which updates in the local coordinate system, is applied.
\begin{equation}
	\begin{aligned}
		\mathbf{x} \leftarrow  \mathbf{x} \oplus \Delta \mathbf{x}^{*}
	\end{aligned} 
\end{equation}

The definition of the right multiplication $\oplus$ operation is as follows.
\begin{equation}
	\begin{aligned}
		\mathbf{x} \oplus \Delta \mathbf{x}^{*}& = \mathbf{x}  \Delta \mathbf{x}^{*} \\& = \mathbf{x}  \exp([\Delta \xi^{*}]_{\times})  \quad \cdots \text{ locally updated (right mult)} \end{aligned} \label{eq:rel2}
\end{equation}

\subsection{Jacobian of relative pose error}
To perform (\ref{eq:rel6}), it is necessary to compute the Jacobian $\mathbf{J}$ of the relative pose error. For the given non-sequential nodes $\mathbf{x}_{i}, \mathbf{x}_{j}$, their Jacobian $\mathbf{J}_{ij}$ can be expressed as follows.
\begin{equation}
	\begin{aligned}
		\mathbf{J}_{ij} & = \frac{\partial \mathbf{e}_{ij}}{\partial \mathbf{x}_{ij}} \\ & =  \frac{\partial \mathbf{e}_{ij}}{\partial [\mathbf{x}_{i}, \mathbf{x}_{j}]}  \\ & = [\mathbf{J}_{i}, \mathbf{J}_{j}]
	\end{aligned}
\end{equation}

If we elaborate on this, it looks like the following.
\begin{equation}
	\begin{aligned}
		\mathbf{J}_{ij} = \frac{\partial \mathbf{e}_{ij}}{\partial [\mathbf{x}_{i}, \mathbf{x}_{j}]} & = \frac{\partial }{\partial [\mathbf{x}_{i}, \mathbf{x}_{j}]} \bigg( \mathbf{z}_{ij}^{-1} \hat{\mathbf{z}}_{ij} \bigg) \\ & = \frac{\partial }{\partial [\mathbf{R}_{i}, \mathbf{t}_{i}, \mathbf{R}_{j}, \mathbf{t}_{j}]} \bigg( \mathbf{z}_{ij}^{-1} \hat{\mathbf{z}}_{ij} \bigg) \\
	\end{aligned}
\end{equation}

\subsubsection{Lie theory-based SE(3) optimization}
When calculating the above Jacobian, since the term $\mathbf{t}$ related to the position is a 3-dimensional vector and the size of this vector is the minimum degrees of freedom to represent 3-dimensional position, which is 3 degrees of freedom, there is no separate constraint when performing optimization updates. \textbfazure{However, the rotation matrix $\mathbf{R}$ has 9 parameters, which is more than the minimum degrees of freedom to represent 3-dimensional rotation, which is 3 degrees of freedom, thus various constraints exist. This is referred to as being over-parameterized. The disadvantages of an over-parameterized representation are as follows.}
\begin{itemize}
	\item Because redundant parameters must be calculated, the computational load increases during optimization.
	\item Additional degrees of freedom can cause numerical instability issues.
	\item Parameters must always be checked to satisfy constraints whenever they are updated.
\end{itemize}

Therefore, a minimal parameter representation free from constraints, a Lie theory-based optimization method, is generally used. \textbfazure{The Lie group SE(3) based optimization method refers to the method of updating SE(3) by calculating the optimal twist $\Delta \xi^{*}$ by changing the term related to rotation from $\mathbf{R} \rightarrow \mathbf{w}$ and the term related to position from $\mathbf{t} \rightarrow \mathbf{v}$, and then using exponential mapping of the lie algebra se(3) $[\Delta \xi]_{\times}$.}
\begin{equation}
	\begin{aligned}
		\begin{bmatrix} \Delta \mathbf{x}_{i}^{*}, \Delta \mathbf{x}_{j}^{*} \end{bmatrix}  \rightarrow [\Delta \xi_{i}^{*}, \Delta \xi_{j}^{*}] 
	\end{aligned}
\end{equation}

The Jacobian for $\xi$ is as follows.
\begin{equation}
	\begin{aligned}
		\mathbf{J}_{ij} & = \frac{\partial \mathbf{e}_{ij}}{\partial [\mathbf{x}_{i}, \mathbf{x}_{j}]}  && \rightarrow  \frac{\partial \mathbf{e}_{ij}}{\partial [\xi_{i}, \xi_{j}]} 
	\end{aligned}
\end{equation}

This changes the existing formula as follows.
\begin{equation}
	\begin{aligned}
		\mathbf{e}_{ij}(\mathbf{x}_{i}, \mathbf{x}_{j}) \quad \quad & \rightarrow \quad \quad \mathbf{e}_{ij}(\xi_{i}, \xi_{j}) \\
		\mathbf{E}(\mathbf{x}) \quad \quad & \rightarrow \quad \quad \mathbf{E}(\xi) \\
		\mathbf{e}_{ij}(\mathbf{x}_{i}, \mathbf{x}_{j}) + \mathbf{J}_{i}' \Delta \mathbf{x}_{i} +  \mathbf{J}_{j}' \Delta \mathbf{x}_{j}  \quad \quad & \rightarrow \quad \quad \mathbf{e}_{ij}(\xi_{i}, \xi_{j}) + \mathbf{J}_{i}\Delta \xi_{i} + \mathbf{J}_{j}\Delta \xi_{j}   \\
		\mathbf{H}\Delta\mathbf{x}^{*} = -\mathbf{b} \quad \quad & \rightarrow \quad \quad \mathbf{H}\Delta \xi^{*}  = -\mathbf{b} \\
		\mathbf{x} \leftarrow \Delta \mathbf{x}^{*}\mathbf{x} \quad \quad & \rightarrow \quad \quad \mathbf{x} \leftarrow \exp([\Delta \xi^{*}]_{\times}) \mathbf{x} 
	\end{aligned}
\end{equation}
- $\mathbf{J}_{ij}'  = \frac{\partial \mathbf{e}}{\partial [\mathbf{x}_{i}, \mathbf{x}_{j}]}$ \\
- $\mathbf{J}_{ij}  = \frac{\partial \mathbf{e}}{\partial [\xi_{i}, \xi_{j}]}$ \\

\mybox{Tip}{gray!40}{gray!10}{
	$\exp([\xi]_{\times}) \in \text{SE}(3)$ refers to the operation of transforming the twist $\xi$ through exponential mapping into a 3-dimensional pose. For detailed information about exponential mapping, refer to \href{https://alida.tistory.com/60\#2.3.2-the-capitalized-exponential-map}{this link}.
	\begin{equation}
		\begin{aligned}
			\exp([\Delta \xi]_{\times}) = \Delta \mathbf{x} 
		\end{aligned}
	\end{equation}
}

$\frac{\partial }{\partial \xi}( \mathbf{z}_{ij}^{-1} \hat{\mathbf{z}}_{ij})$ does not directly appear in the parameters $\xi$ from $\mathbf{z}_{ij}^{-1} \hat{\mathbf{z}}_{ij}$, so it needs to be changed into a term related to lie algebra.
\begin{equation}
	\begin{aligned}
		\mathbf{z}_{ij}^{-1}\hat{\mathbf{z}}_{ij} & \rightarrow \text{Log}(\mathbf{z}_{ij}^{-1}\hat{\mathbf{z}}_{ij}) 
	\end{aligned}
\end{equation}

At this time, $\text{Log}(\cdot)$ means logarithm mapping that changes SE(3) into twist $\xi \in \mathbb{R}^{6}$. For detailed information about Logarithm mapping, refer to \href{https://alida.tistory.com/9#org608a5f4-1}{this post}. Therefore, the SE(3) version of the relative pose error $\mathbf{e}_{ij}$ is changed as follows.
\begin{equation}
	\boxed{ \begin{aligned}
			\mathbf{e}_{ij}(\mathbf{x}_{i},\mathbf{x}_{j}) = \mathbf{z}_{ij}^{-1}\hat{\mathbf{z}}_{ij} \quad \rightarrow \quad \mathbf{e}_{ij}(\xi_{i}, \xi_{j})& = \text{Log}(\mathbf{z}_{ij}^{-1}\hat{\mathbf{z}}_{ij}) 
	\end{aligned} }  \label{eq:rel22}
\end{equation}

This is elaborated as follows.
\begin{equation}
	\begin{aligned}
		\mathbf{e}_{ij}(\xi_{i}, \xi_{j}) & = \text{Log}(\mathbf{z}_{ij}^{-1}\hat{\mathbf{z}}_{ij}) \\ & = \text{Log}(\mathbf{z}_{ij}^{-1}\mathbf{x}_{i}^{-1}\mathbf{x}_{j}) \\ & = \text{Log}(\exp([-\xi_{ij}]_{\times})\exp([-\xi_{i}]_{\times})\exp([\xi_{j}]_{\times}))
	\end{aligned}
\end{equation}

From this equation, we can see that the parameters $\xi_{i}, \xi_{j}$ in $\mathbf{z}_{ij}$ are connected through exponential mapping. If we apply the left perturbation model to the second line of the formula and express the increment, it looks like this.
\begin{equation}
	\begin{aligned}
		\mathbf{e}_{ij}(\xi_{i} + \Delta \xi_{i}, \xi_{j} + \Delta \xi_{j}) & = \text{Log}(\hat{\mathbf{z}}_{ij}^{-1}\mathbf{x}_{i}^{-1}\exp(-[\Delta \xi_{i}]_{\times})\exp([\Delta \xi_{j}]_{\times})\mathbf{x}_{j})
	\end{aligned} \label{eq:rel10}
\end{equation}
	
\mybox{Tip}{gray!40}{gray!10}{
	To arrange the term in the form $\mathbf{e} + \mathbf{J}\Delta\xi$ by moving the incremental term to the left or right, the following property of the adjoint matrix of SE(3) must be used. For more information about the adjoint matrix, refer to \href{https://alida.tistory.com/9\#org8b9a1c3}{this post}.
	\begin{equation}
		\begin{aligned}
			\exp([\text{Ad}_{\mathbf{T}} \cdot \xi]_{\times}) = \mathbf{T} \cdot \exp([\xi]_{\times}) \cdot \mathbf{T}^{-1}
		\end{aligned}
	\end{equation}
	
	Transforming the above formula for $\mathbf{T} \rightarrow \mathbf{T}^{-1}$, we get the following.
	\begin{equation}
		\begin{aligned}
			\exp([\text{Ad}_{\mathbf{T}^{-1}} \cdot \xi]_{\times}) = \mathbf{T}^{-1} \cdot \exp([\xi]_{\times}) \cdot \mathbf{T}
		\end{aligned}
	\end{equation}
	
	And simplifying gives the following formula.
	\begin{equation}
		\begin{aligned}
			\exp([\xi]_{\times}) \cdot \mathbf{T} = \mathbf{T} \exp([\text{Ad}_{\mathbf{T}^{-1}} \cdot \xi]_{\times})
		\end{aligned} \label{eq:rel11}
	\end{equation}
}

Using (\ref{eq:rel11}), it is possible to move the $\exp(\cdot)\exp(\cdot)$ term in the middle of (\ref{eq:rel10}) to the right or left. This post describes the process of moving it to the right. This is expanded for each $\Delta \xi_{i}, \Delta \xi_{j}$ as follows.
\begin{equation}
	\begin{aligned}
		\mathbf{e}_{ij}(\xi_{i} + \Delta \xi_{i}, \xi_{j}) & = \text{Log}(\hat{\mathbf{z}}_{ij}^{-1}\mathbf{x}_{i}^{-1}\exp(-[\Delta \xi_{i}]_{\times}) \mathbf{x}_{j}) \\ &= \text{Log}(\mathbf{z}_{ij}^{-1}\mathbf{x}_{i}^{-1}\mathbf{x}_{j} \exp([-\text{Ad}_{\mathbf{x}_{j}^{-1}}\Delta \xi_{i}]_{\times})) \quad \cdots \text{[1]} \\
		\mathbf{e}_{ij}(\xi_{i}, \xi_{j} + \Delta \xi_{j}) & = \text{Log}(\hat{\mathbf{z}}_{ij}^{-1}\mathbf{x}_{i}^{-1}\exp([\Delta \xi_{j}]_{\times})\mathbf{x}_{j}) \\ &= \text{Log}(\mathbf{z}_{ij}^{-1}\mathbf{x}_{i}^{-1}\mathbf{x}_{j} \exp([\text{Ad}_{\mathbf{x}_{j}^{-1}}\Delta \xi_{j}]_{\times})) \quad \cdots \text{[2]}
	\end{aligned} \label{eq:rel21}
\end{equation}

To express this simply using substitution, $[1], [2]$ are as follows.
\begin{equation}
	\begin{aligned}
		\text{Log}( \exp([\mathbf{a}]_{\times}) \exp([\mathbf{b}]_{\times})) \quad \cdots \text{[1]} \\
		\text{Log}( \exp([\mathbf{a}]_{\times}) \exp([\mathbf{c}]_{\times})) \quad \cdots \text{[2]}
	\end{aligned} \label{eq:rel13}
\end{equation}
- $\exp([\mathbf{a}]_{\times}) = \mathbf{z}_{ij}^{-1}\mathbf{x}_{i}^{-1}\mathbf{x}_{j}$: Transformation matrix expressed as an exponential term. According to (\ref{eq:rel22}), $\mathbf{a} = \mathbf{e}_{ij}(\xi_{i},  \xi_{j})$. \\
- $\mathbf{b} = -\text{Ad}_{\mathbf{x}_{j}^{-1}}\Delta \xi_{i}$\\
- $\mathbf{c} = \text{Ad}_{\mathbf{x}_{j}^{-1}}\Delta \xi_{j}$\\

This formula can be organized using the right BCH approximation.
\mybox{Tip}{gray!40}{gray!10}{
	The right BCH approximation is as follows.
	\begin{equation}
		\begin{aligned}
			\exp([\xi]_{\times})\exp([\Delta \xi]_{\times}) & = \exp([\xi + \mathcal{J}^{-1}_{r}\Delta \xi]_{\times}) \\
			\exp([\xi + Delta \xi]_{\times}) & = \exp([\xi]_{\times}) \exp([\mathcal{J}_{r}\Delta \xi]_{\times})\\
		\end{aligned} 
	\end{equation}
	
	For detailed information, refer to \href{https://docs.google.com/document/d/1icPjUyT3nPvjZ1OVMtWp9afUtuJ4gXLJL-ex7A9FpNs/edit?fbclid=IwAR2VfhZ3js52zkFpZpJ5HZv_qQLPz7WCTBWkwn6IF1MmHa3Ksyhi5TQSAfY\#}{Introduction to Visual SLAM Chapter 4}.
}

Using the BCH approximation, (\ref{eq:rel13}) is organized as follows.
\begin{equation}
	\begin{aligned} 
		\text{Log}(\exp([\mathbf{a}]_{\times}) \exp([\mathbf{b}]_{\times}) ) & = \text{Log}(\exp([\mathbf{a} + \mathcal{J}^{-1}_{r}\mathbf{b}]_{\times})) \\ & = \mathbf{a} + \mathcal{J}^{-1}_{r}\mathbf{b} \quad \cdots \text{[1]} \\
		\text{Log}(\exp([\mathbf{a}]_{\times}) \exp([\mathbf{c}]_{\times}) ) & = \text{Log}(\exp([\mathbf{a} + \mathcal{J}^{-1}_{r}\mathbf{c}]_{\times})) \\ & = \mathbf{a} +  \mathcal{J}^{-1}_{r}\mathbf{c} \quad \cdots \text{[2]}
	\end{aligned}
\end{equation}

Finally, undoing the substitution and combining the $\Delta \xi_{i}, \Delta \xi_{j}$ formulas gives the SE(3) version of the formula in (\ref{eq:rel14}).
\begin{equation}
	\boxed{ \begin{aligned}
			\mathbf{e}_{ij}(\xi_{i} + \Delta \xi_{i}, \xi_{j} + \Delta \xi_{j}) & = \mathbf{a} + \mathcal{J}^{-1}_{r}\mathbf{b} + \mathcal{J}^{-1}_{r}\mathbf{c} \\& = \mathbf{e}_{ij}(\xi_{i} , \xi_{j} ) -  \mathcal{J}^{-1}_{r}\text{Ad}_{\mathbf{x}_{j}^{-1}}\Delta \xi_{i} + \mathcal{J}^{-1}_{r}\text{Ad}_{\mathbf{x}_{j}^{-1}}\Delta \xi_{j} \\ &  = \mathbf{e}_{ij}(\xi_{i} , \xi_{j} ) + \frac{\partial \mathbf{e}_{ij} }{\partial \Delta \xi_{i}} \Delta \xi_{i} + \frac{\partial \mathbf{e}_{ij} }{\partial \Delta \xi_{j}} \Delta \xi_{j}
	\end{aligned} }
\end{equation}

Therefore, the final relative pose error Jacobian for SE(3) is as follows.
\begin{equation}
	\boxed{ \begin{aligned}
			& \frac{\partial \mathbf{e}_{ij} }{\partial \Delta \xi_{i}} = -\mathcal{J}^{-1}_{r}\text{Ad}_{\mathbf{x}_{j}^{-1}} \in \mathbb{R}^{6\times6} \\
			& \frac{\partial \mathbf{e}_{ij} }{\partial \Delta \xi_{j}} = \mathcal{J}^{-1}_{r}\text{Ad}_{\mathbf{x}_{j}^{-1}} \in \mathbb{R}^{6\times6}
	\end{aligned} }
\end{equation}

At this time, $\mathcal{J}^{-1}_{r}$ is generally approximated as follows or used by setting it as $\mathbf{I}_{6}$.
\begin{equation}
	\begin{aligned}
		\mathcal{J}^{-1}_{r} \approx \mathbf{I}_{6} + \frac{1}{2} \begin{bmatrix} [\mathbf{w}]_{\times} & [\mathbf{v}]_{\times} \\ \mathbf{0} & [\mathbf{w}]_{\times} \end{bmatrix} \in \mathbb{R}^{6\times6}
	\end{aligned} 
\end{equation}

If $\mathcal{J}^{-1}_{r} = \mathbf{I}_{6}$ is assumed and optimization is performed, there is a reduction in computational load, but the optimization performance is slightly superior when using the approximated Jacobian as above. For detailed information, refer to \href{https://docs.google.com/document/d/1s3kG2QG8qEIie1ZpDqXBP9TVYwKJJtHv2HHADYwoldw/edit?usp=drivesdk}{Introduction to Visual SLAM Chapter 11}.

\subsection{Code implementations}
\begin{itemize}
	\item g2o code: \href{https://github.com/RainerKuemmerle/g2o/blob/master/g2o/types/sba/edge_se3_expmap.cpp#L55}{edge\_se3\_expmap.cpp\#L55}
	\begin{itemize}  
		\item In the above g2o code, the error is defined as $\mathbf{e}_{ij}=\mathbf{x}_{j}^{-1}\mathbf{z}_{ij}\mathbf{x}_{i}$, so the Jacobian is slightly different from the explanation above.
		\item $\frac{\partial \mathbf{e}_{ij} }{\partial \Delta \xi_{i}} = \mathcal{J}^{-1}_{l}\text{Ad}_{\mathbf{x}_{j}^{-1}\mathbf{z}_{ij}}$
		\item $\frac{\partial \mathbf{e}_{ij} }{\partial \Delta \xi_{j}} = -\mathcal{J}^{-1}_{r}\text{Ad}_{\mathbf{x}_{i}^{-1}\mathbf{z}_{ij}^{-1}}$
		\item This follows the same form as combining after arranging the $\Delta \xi_{i}$ to the left and $\Delta \xi_{j}$ to the right in (\ref{eq:rel21}).
		\item It also appears that $\mathcal{J}^{-1}_{l} \approx \mathbf{I}_{6}, \mathcal{J}^{-1}_{r} \approx \mathbf{I}_{6}$ is approximated. Thus, the actual implemented code is as follows.
		\begin{itemize}
			\item $\frac{\partial \mathbf{e}_{ij} }{\partial \Delta \xi_{i}} \approx \text{Ad}_{\mathbf{x}_{j}^{-1}\mathbf{z}_{ij}}$
			\item $\frac{\partial \mathbf{e}_{ij} }{\partial \Delta \xi_{j}} \approx -\text{Ad}_{\mathbf{x}_{i}^{-1}\mathbf{z}_{ij}^{-1}}$
		\end{itemize}
	\end{itemize}
\end{itemize}

\section{Line Reprojection Error}
Line reprojection error is used to optimize a 3D line expressed in Plücker coordinates. For more details on Plücker coordinates, refer to the post \href{https://alida.tistory.com/12}{Plücker Coordinate Concept Summary}.
~\\ ~\\
\textbf{NOMENCLATURE of line reprojection error}
\begin{itemize}
	\item $\mathcal{T}_{cw} \in \mathbb{R}^{6\times6}$: Transformation matrix for the Plücker line
	\item $\mathcal{K}_{L}$: Internal parameter matrix for the line (line intrinsic matrix)
	\item $\mathbf{U} \in SO(3)$: Rotation matrix for the 3D line
	\item $\mathbf{W} \in SO(2)$: Matrix containing distance information of the 3D line from the origin
	\item $\boldsymbol{\theta} \in \mathbb{R}^{3}$: Parameters corresponding to the SO(3) rotation matrix
	\item $\theta \in \mathbb{R}$: Parameter corresponding to the SO(2) rotation matrix
	\item $\mathbf{u}_{i}$: $i$th column vector
	\item $\mathcal{X} = [\delta_{\boldsymbol{\theta}}, \delta_\xi ]$: State variable
	\item $\delta_{\boldsymbol{\theta}} = [ \boldsymbol{\theta}^{\intercal}, \theta ] \in \mathbb{R}^{4}$: State variable in orthonormal representation
	\item $\delta_{\xi} = [\delta \xi] \in se(3)$: Update method through Lie theory, refer to \href{https://alida.tistory.com/52#lie-theory-based-optimization}{this link}
	\item $\oplus$ : Operator to update the state variables $\delta_{\boldsymbol{\theta}}, \delta_\xi$ at once.
	\item $\mathbf{J} = \frac{\partial \mathbf{e}_{l}}{\partial \mathcal{X}} = \frac{\partial \mathbf{e}_{l}}{\partial [\delta_{\boldsymbol{\theta}}, \delta_\xi]}$
\end{itemize}

\begin{figure}[h!]
	\centering
	\includegraphics[width=10cm]{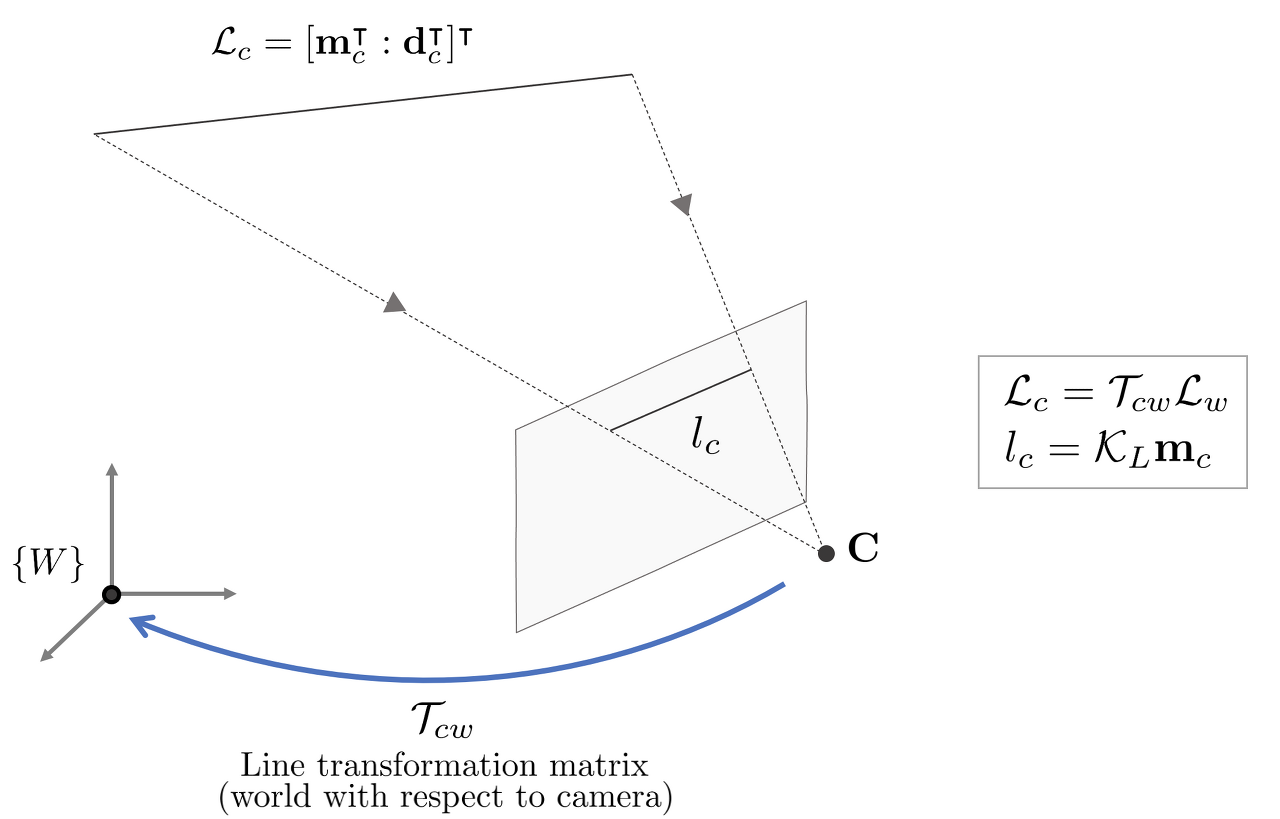}
\end{figure}
A line in 3D space can be expressed as a 6-dimensional column vector using Plücker Coordinates. 
\begin{equation}
	\begin{aligned}
		\mathcal{L} & = [\mathbf{m}^{\intercal} : \mathbf{d}^{\intercal}]^{\intercal} = [ m_{x} : m_{y} : m_{z} : d_{x} : d_{y} : d_{z}]^{\intercal}
	\end{aligned}
\end{equation}

The order in papers using Plücker Coordinates is mostly $[\mathbf{m} : \mathbf{d}]$, hence this section uses this order to represent the line. This line representation has scale ambiguity (up to scale), so it has 5 degrees of freedom; $\mathbf{m}, \mathbf{d}$ do not need to be unit vectors, and the line can be uniquely represented by the ratio of the two vector values.

\subsection{Line Transformation and Projection}
If we refer to a line in the world coordinate system as $\mathcal{L}_{w}$, then its transformation to the camera coordinate system can be expressed as follows:
\begin{equation}
	\begin{aligned}
		\mathcal{L}_{c} & = \begin{bmatrix} \mathbf{m}_{c} \\ \mathbf{d}_{c} \end{bmatrix} = \mathcal{T}_{cw}\mathcal{L}_{w} = \begin{bmatrix} \mathbf{R}_{cw} & \mathbf{t}^{\wedge}\mathbf{R}_{cw} \\ 0 & \mathbf{R}_{cw} \end{bmatrix} \begin{bmatrix} \mathbf{m}_{w} \\ \mathbf{d}_{w} \end{bmatrix}
	\end{aligned}
\end{equation}

The projection of this line onto the image plane is as follows:
\begin{equation}
	\begin{aligned}
		l_{c}  & = \begin{bmatrix} l_{1} \\ l_{2} \\ l_{3} \end{bmatrix} = \mathcal{K}_{L}\mathbf{m}_{c} = \begin{bmatrix} f_{y} && \\ &f_{x} & \\ -f_{y}c_{x} & -f_{x}c_{y} & f_{x}f_{y} \end{bmatrix} \begin{bmatrix} m_{x} \\ m_{y} \\ m_{z} \end{bmatrix}
	\end{aligned}
\end{equation}

$\mathcal{K}_{L}$ means $\mathcal{P} = [\det(\mathbf{N}) \mathbf{N}^{-\intercal} | \mathbf{n}^{\wedge} \mathbf{N} ]$ where $\mathbf{P} = K [ \mathbf{I} | \mathbf{0}]$. Thus, $\mathcal{P} = [\det(\mathbf{K})\mathbf{K}^{-\intercal} | \mathbf{0}]$, so the $\mathbf{d}$ term of $\mathcal{L}$ is eliminated. Therefore, when $\mathbf{K} = \begin{bmatrix} f_x & & c_x \\ & f_y & c_y \\ &&1 \end{bmatrix}$, the following equation is derived:
\begin{equation}
	\begin{aligned}
		\mathcal{K}_{L} = \det(\mathbf{K})\mathbf{K}^{-\intercal} = \begin{bmatrix} f_{y} && \\ &f_{x} & \\ -f_{y}c_{x} & -f_{x}c_{y} & f_{x}f_{y} \end{bmatrix} \in \mathbb{R}^{3\times3}
	\end{aligned}
\end{equation}

\subsection{Line Reprojection Error}
\begin{figure}[h!]
	\centering
	\includegraphics[width=12cm]{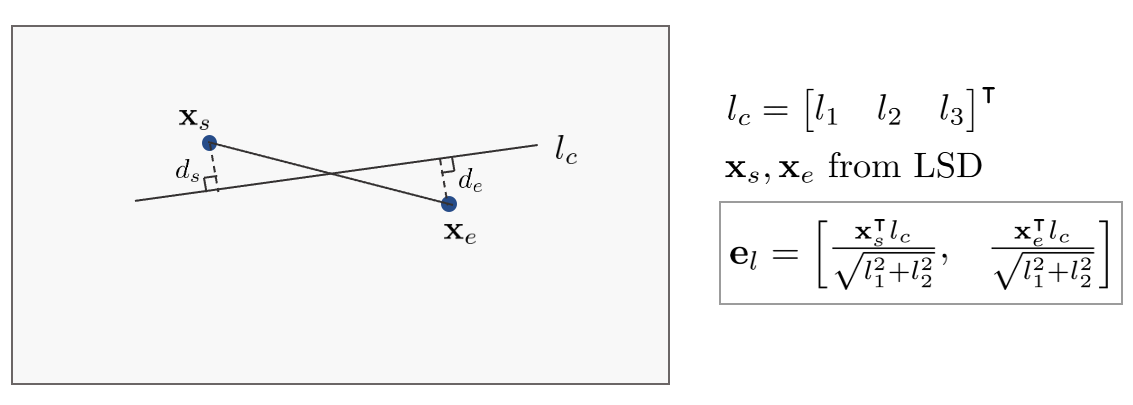}
\end{figure}
The reprojection error $\mathbf{e}_{l}$ of the line can be expressed as follows:
\begin{equation}
	\begin{aligned}
		\mathbf{e}_{l} = \begin{bmatrix} d_{s}, \ d_{e} \end{bmatrix}= \begin{bmatrix} \frac{\mathbf{x}_{s}^{\intercal}l_{c}}{\sqrt{l_{1}^{2} + l_{2}^{2}}},  & \frac{\mathbf{x}_{e}^{\intercal}l_{c}}{\sqrt{l_{1}^{2} + l_{2}^{2}}}\end{bmatrix} \in \mathbb{R}^{2}
	\end{aligned}
\end{equation}

This can be expressed using the \href{https://en.wikipedia.org/wiki/Distance_from_a_point_to_a_line#Line_defined_by_an_equation}{distance from a point to a line formula}. Here, $\{\mathbf{x}_{s}, \mathbf{x}_{e}\}$ represent the starting and ending points of the line extracted using a line feature extractor (e.g., LSD). \textbfazure{In other words, $l_{c}$ is the predicted value obtained through modeling, and the line connecting $\mathbf{x}_{s}, \mathbf{x}_{e}$ becomes the observed value measured through sensor data.}

\subsection{Orthonormal Representation}
Using the previously calculated $\mathbf{e}_{l}$ for BA optimization poses problems when using the Plücker Coordinate representation directly because Plücker Coordinates must always satisfy the Klein quadric constraint $\mathbf{m}^{\intercal}\mathbf{d} = 0$, which implies 5 degrees of freedom, making it over-parameterized compared to the minimum 4 parameters needed to represent a line. The disadvantages of an over-parameterized representation are as follows:
\begin{itemize}
	\item Redundant parameters must be calculated, increasing computational load during optimization.
	\item Additional degrees of freedom can lead to numerical instability.
	\item The parameters must always be checked to satisfy the constraint after each update.
\end{itemize}

Therefore, when optimizing a line, it is common to change to a 4-degree of freedom using the orthonormal representation method. \textbfazure{That is, while lines are represented using Plücker Coordinates, optimization is performed using the orthonormal representation, and then the optimized values are converted back to Plücker Coordinates.}

Orthonormal representation is as follows. A line in 3D space can always be represented as follows:
\begin{equation}
	\begin{aligned}
		(\mathbf{U}, \mathbf{W}) \in SO(3) \times SO(2)
	\end{aligned}
\end{equation}

Any given Plücker line $\mathcal{L} = [\mathbf{m}^{\intercal} : \mathbf{d}^{\intercal}]^{\intercal}$ always has a corresponding $(\mathbf{U}, \mathbf{W})$, and this representation method is called the orthonormal representation. When a line $\mathcal{L}_{w} = [ \mathbf{m}_{w}^{\intercal} : \mathbf{d}_{w}^{\intercal}]^{\intercal}$ in the world is given, $\mathcal{L}_{w}$ can be obtained through QR decomposition as follows:
\begin{equation}
	\begin{aligned}
		\begin{bmatrix} \mathbf{m}_{w} \ | \ \mathbf{d}_{w}  \end{bmatrix}= \mathbf{U} \begin{bmatrix} w_{1} & 0\\ 0& w_{2} \\0 & 0 \end{bmatrix}, \quad \text{with set: } \mathbf{W} = \begin{bmatrix} w_{1} & -w_{2} \\ w_{2} & w_{1} \end{bmatrix}
	\end{aligned}
\end{equation}

In this case, the upper triangle matrix $\mathbf{R}$'s $(1,2)$ element is always 0 due to the Plücker constraint (Klein quadric). $\mathbf{U}, \mathbf{W}$ represent 3D and 2D rotation matrices, respectively, so $\mathbf{U} = \mathbf{R}(\boldsymbol{\theta}), \mathbf{W} = \mathbf{R}(\theta)$ can be represented as follows:
\begin{equation}
	\begin{aligned}
		\mathbf{R}(\boldsymbol{\theta}) & = \mathbf{U} = \begin{bmatrix} \mathbf{u}_{1} & \mathbf{u}_{2} & \mathbf{u}_{3} \end{bmatrix} = \begin{bmatrix} \frac{\mathbf{m}_{w}}{\left\| \mathbf{m}_{w} \right\|} &  \frac{\mathbf{d}_{w}}{\left\| \mathbf{d}_{w} \right\|} &  \frac{\mathbf{m}_{w} \times \mathbf{d}_{w}}{\left\| \mathbf{m}_{w} \times \mathbf{d}_{w} \right\|} \end{bmatrix} \\ 
		\mathbf{R}(\theta) & = \mathbf{W} =  \begin{bmatrix} w_{1} & -w_{2} \\ w_{2} & w_{1} \end{bmatrix} =  \begin{bmatrix} \cos\theta & -\sin\theta \\ \sin\theta & \cos\theta \end{bmatrix} \\ 
		& = \frac{1}{\sqrt{\left\| \mathbf{m}_{w} \right\|^{2} + \left\| \mathbf{d}_{w} \right\|^{2}}} \begin{bmatrix} \left\| \mathbf{m}_{w} \right\|  & \left\| \mathbf{d}_{w} \right\|  \\ -\left\| \mathbf{d}_{w} \right\| & \left\| \mathbf{m}_{w} \right\| \end{bmatrix}
	\end{aligned}
\end{equation}

When actually performing optimization, $\mathbf{U} \leftarrow \mathbf{U}\mathbf{R}(\boldsymbol{\theta}), \mathbf{W} \leftarrow \mathbf{W}\mathbf{R}(\theta)$ are updated as follows. \textbfazure{Therefore, the orthonormal representation can represent a 3D line through $\delta_{\boldsymbol{\theta}} = [ \boldsymbol{\theta}^{\intercal}, \theta ] \in \mathbb{R}^{4}$.} The updated $[ \boldsymbol{\theta}^{\intercal}, \theta ]$ is converted back to $\mathcal{L}_{w}$ as follows:
\begin{equation}
	\begin{aligned}
		\mathcal{L}_{w} = \begin{bmatrix} w_{1} \mathbf{u}_{1}^{\intercal} & w_{2} \mathbf{u}_{2}^{\intercal} \end{bmatrix}
	\end{aligned}
\end{equation}

\subsection{Error Function Formulation}
To optimize the line reprojection error $\mathbf{e}_{l}$, nonlinear least squares methods such as Gauss-Newton (GN), Levenberg-Marquardt (LM), etc., are used to iteratively update the optimal variables. The error function using reprojection error is expressed as follows:
\begin{equation}
	\begin{aligned}
		\mathbf{E}_{l}(\mathcal{X}) & =   \sum_{i}\sum_{j} \left\| \mathbf{e}_{l,ij} \right\|^{2} \\
	\end{aligned} 
\end{equation}
\begin{equation}
	\begin{aligned}
		\mathcal{X}^{*} &= \arg\min_{\mathcal{X}^{*}}  \mathbf{E}_{l}(\mathcal{X}) \\
		& =  \arg\min_{\mathcal{X}^{*}} \sum_{i}\sum_{j} \left\| \mathbf{e}_{l,ij} \right\|^{2} \\
		& = \arg\min_{\mathcal{X}^{*}} \sum_{i}\sum_{j} \mathbf{e}_{l,ij}^{\intercal}\mathbf{e}_{l,ij} \\
	\end{aligned} \label{eq:line10}
\end{equation}

The $\mathbf{E}_{l}(\mathcal{X}^{*})$ that satisfies $\left\|\mathbf{e}_{l}(\mathcal{X}^{*})\right\|^{2}$ can be computed iteratively through non-linear least squares. A small increment $\Delta \mathcal{X}$ is iteratively updated to $\mathcal{X}$ to find the optimal state.
\begin{equation}
	\begin{aligned}
		\arg\min_{\mathcal{X}^{*}} \mathbf{E}_{l}(\mathcal{X} + \Delta \mathcal{X}) & = \arg\min_{\mathcal{X}^{*}} \sum_{i}\sum_{j} \left\|\mathbf{e}_{l}(\mathcal{X} +\Delta \mathcal{X})\right\|^{2} 
	\end{aligned}
\end{equation}

Strictly speaking, the state increment $\Delta \mathcal{X}$ includes an SE(3) transformation matrix, so it is correct to add it to the existing state $\mathcal{X}$ through the $\oplus$ operator, but the $+$ operator is used for simplicity of expression.
\begin{equation}
	\begin{aligned}
		\mathbf{e}_{l}( \mathcal{X} \oplus \Delta \mathcal{X}) 
		\quad \rightarrow \quad\mathbf{e}_{l}(\mathcal{X} + \Delta \mathcal{X})
	\end{aligned}
\end{equation}

The above equation can be expressed through a Taylor first-order approximation as follows:
\begin{equation}
	\begin{aligned}
		\mathbf{e}_{l}(\mathcal{X} + \Delta \mathcal{X})  & \approx \mathbf{e}_{l}(\mathcal{X}) + \mathbf{J}\Delta \mathcal{X} \\ & = \mathbf{e}_{l}(\mathcal{X}) + \mathbf{J}_{\boldsymbol{\theta}} \Delta \delta_{\boldsymbol{\theta}} +\mathbf{J}_{\xi} \Delta \delta_\xi \\
		& = \mathbf{e}_{l}(\mathcal{X}) + \frac{\partial \mathbf{e}_{l}}{\partial \delta_{\boldsymbol{\theta}}} \Delta \delta_{\boldsymbol{\theta}} + \frac{\partial \mathbf{e}_{l}}{\partial \delta_\xi}\Delta \delta_\xi \\
	\end{aligned}
\end{equation}

\begin{equation}
	\begin{aligned}
		\arg\min_{\mathcal{X}^{*}} \mathbf{E}_{l}(\mathcal{X} + \Delta \mathcal{X}) & \approx \arg\min_{\mathcal{X}^{*}} \sum_{i}\sum_{j} \left\|\mathbf{e}_{l}(\mathcal{X}) + \mathbf{J}\Delta \mathcal{X} \right\|^{2} \\
	\end{aligned}
\end{equation}

The optimal increment $\Delta \mathcal{X}^{*}$ is obtained by differentiating the above. The detailed derivation process is omitted in this section. For detailed information on the derivation process, refer to \hyperref[sec:opt]{the previous section}.
\begin{equation}
	\begin{aligned}
		& \mathbf{J}^{\intercal}\mathbf{J} \Delta \mathcal{X}^{*} = -\mathbf{J}^{\intercal}\mathbf{e} \\ 
		& \mathbf{H}\Delta \mathcal{X}^{*} = - \mathbf{b} \\
	\end{aligned}
\end{equation}

\subsubsection{The Analytical Jacobian of 3D Line}
As explained in the previous section, to perform nonlinear optimization, $\mathbf{J}$ must be calculated. $\mathbf{J}$ is composed as follows:
\begin{equation}
	\begin{aligned}
		& \mathbf{J} = [ \mathbf{J}_{\boldsymbol{\theta}}, \mathbf{J}_{\xi} ] 
	\end{aligned}
\end{equation}

$[ \mathbf{J}_{\boldsymbol{\theta}}, \mathbf{J}_{\xi} ]$ can be expanded as follows:
\begin{equation}
	\begin{aligned}
		& \mathbf{J}_{\boldsymbol{\theta}} = \frac{\partial \mathbf{e}_{l}}{\partial \delta_{\boldsymbol{\theta}}} = \frac{\partial \mathbf{e}_{l}}{\partial l} \frac{\partial l}{\partial \mathcal{L}_{c}} \frac{\partial \mathcal{L}_{c}}{\partial \mathcal{L}_{w}} \frac{\partial \mathcal{L}_{w}}{\partial \delta_{\boldsymbol{\theta}}} \\
		& \mathbf{J}_{\xi} = \frac{\partial \mathbf{e}_{l}}{\partial \delta_{\xi}} = \frac{\partial \mathbf{e}_{l}}{\partial l} \frac{\partial l}{\partial \mathcal{L}_{c}} \frac{\partial \mathcal{L}_{c}}{\partial \delta_{\xi}}
	\end{aligned}
\end{equation}

$\frac{\partial \mathbf{e}_{l}}{\partial l}$ can be obtained as follows. Note that $l$ is a vector and $l_{i}$ is a scalar.
\begin{equation}
	\boxed{ \begin{aligned}
			\frac{\partial \mathbf{e}_{l}}{\partial l}  = \frac{1}{\sqrt{l_{1}^{2} + l_{2}^{2}}} \begin{bmatrix} x_{s} - \frac{l_{1} \mathbf{x}_{s}l}{\sqrt{l_{1}^{2} + l_{2}^{2}}} & y_{s} - \frac{l_{2} \mathbf{x}_{s}l}{\sqrt{l_{1}^{2} + l_{2}^{2}}} & 1 \\
				x_{e} - \frac{l_{1} \mathbf{x}_{e}l}{\sqrt{l_{1}^{2} + l_{2}^{2}}} & y_{e} - \frac{l_{2} \mathbf{x}_{e}l}{\sqrt{l_{1}^{2} + l_{2}^{2}}} & 1 \end{bmatrix} \in \mathbb{R}^{2\times3}
	\end{aligned} }
\end{equation}

$\frac{\partial l}{\partial \mathcal{L}_{c}}$ can be obtained as follows:
\begin{equation}
	\boxed{ \begin{aligned}
			\frac{\partial l}{\partial \mathcal{L}_{c}} = \frac{\partial \mathcal{K}_{L}\mathbf{m}_{c}}{\partial \mathcal{L}_{c}} = \begin{bmatrix} \mathcal{K}_{L} & \mathbf{0}_{3\times3} \end{bmatrix} = \begin{bmatrix} f_{y} && &   0 &0&0  \\ &f_{x} &  & 0 &0&0  \\ && 1 &0&0 &0 \end{bmatrix} \in \mathbb{R}^{3\times6}
	\end{aligned} }
\end{equation}

$\frac{\partial \mathcal{L}_{c}}{\partial \mathcal{L}_{w}}$ can be obtained as follows:
\begin{equation}
	\boxed{ \begin{aligned}
			\frac{\partial \mathcal{L}_{c}}{\partial \mathcal{L}_{w}}  = \mathbf{J}_{SE(3)}(\mathcal{L}_{c}) = \begin{bmatrix} \mathbf{I}_{3\times3} & -[\mathbf{t}]_{\times} \end{bmatrix} = \begin{bmatrix} 1 & 0 & -y_{c} \\ 0 & 1 & x_{c} \\ 0 & 0 & 0 \end{bmatrix} \in \mathbb{R}^{3\times3}
	\end{aligned} }
\end{equation}

$\frac{\partial \mathcal{L}_{w}}{\partial \delta_{\boldsymbol{\theta}}}$ can be obtained as follows:
\begin{equation}
	\boxed{ \begin{aligned}
			\frac{\partial \mathcal{L}_{w}}{\partial \delta_{\boldsymbol{\theta}}} = \frac{\partial \mathcal{L}_{w}}{\partial \mathbf{R}}\frac{\partial \mathbf{R}}{\partial \delta_{\boldsymbol{\theta}}} = \begin{bmatrix} 1&0&0 \\ 0 & 1 & 0 \end{bmatrix} \begin{bmatrix} \mathbf{0} & -\mathbf{W} \\ \mathbf{W} & \mathbf{0} \end{bmatrix} \in \mathbb{R}^{2\times4}
	\end{aligned} }
\end{equation}

$\frac{\partial \mathcal{L}_{c}}{\partial \delta_{\xi}}$ can be obtained as follows:
\begin{equation}
	\boxed{ \begin{aligned}
			\frac{\partial \mathcal{L}_{c}}{\partial \delta_{\xi}}  = \mathbf{J}_{SO(3)}(\mathcal{L}_{c}) = \frac{\partial \mathcal{L}_{c}}{\partial \mathbf{R}}\frac{\partial \mathbf{R}}{\partial \delta_{\xi}} = \begin{bmatrix} -\mathbf{R} & \mathbf{0} \\ \mathbf{0} & \mathbf{R} \end{bmatrix} \begin{bmatrix} -\sin\theta & -\cos\theta & 0 \\ \cos\theta & -\sin\theta & 0 \\ 0 & 0 & 0 \\ 0 & 0 & 0 \\ 0 & 0 & -\sin\theta \\ 0 & 0 & \cos\theta \end{bmatrix} \in \mathbb{R}^{6\times1}
	\end{aligned} }
\end{equation}
	
	\subsection{Code implementations}
	\begin{itemize}
		\item Structure PLP SLAM code: \href{https://github.com/PeterFWS/Structure-PLP-SLAM/blob/main/src/PLPSLAM/optimize/g2o/se3/pose_opt_edge_line3d_orthonormal.h#L62}{g2o/se3/pose\_opt\_edge\_line3d\_orthonormal.h\#L62}
		\item Structure PLP SLAM code2: \href{https://github.com/PeterFWS/Structure-PLP-SLAM/blob/main/src/PLPSLAM/optimize/g2o/se3/pose_opt_edge_line3d_orthonormal.h#L81}{g2o/se3/pose\_opt\_edge\_line3d\_orthonormal.h\#L81}
	\end{itemize}
	
\section{IMU measurement error}
\begin{figure}[h!]
	\centering
	\includegraphics[width=16cm]{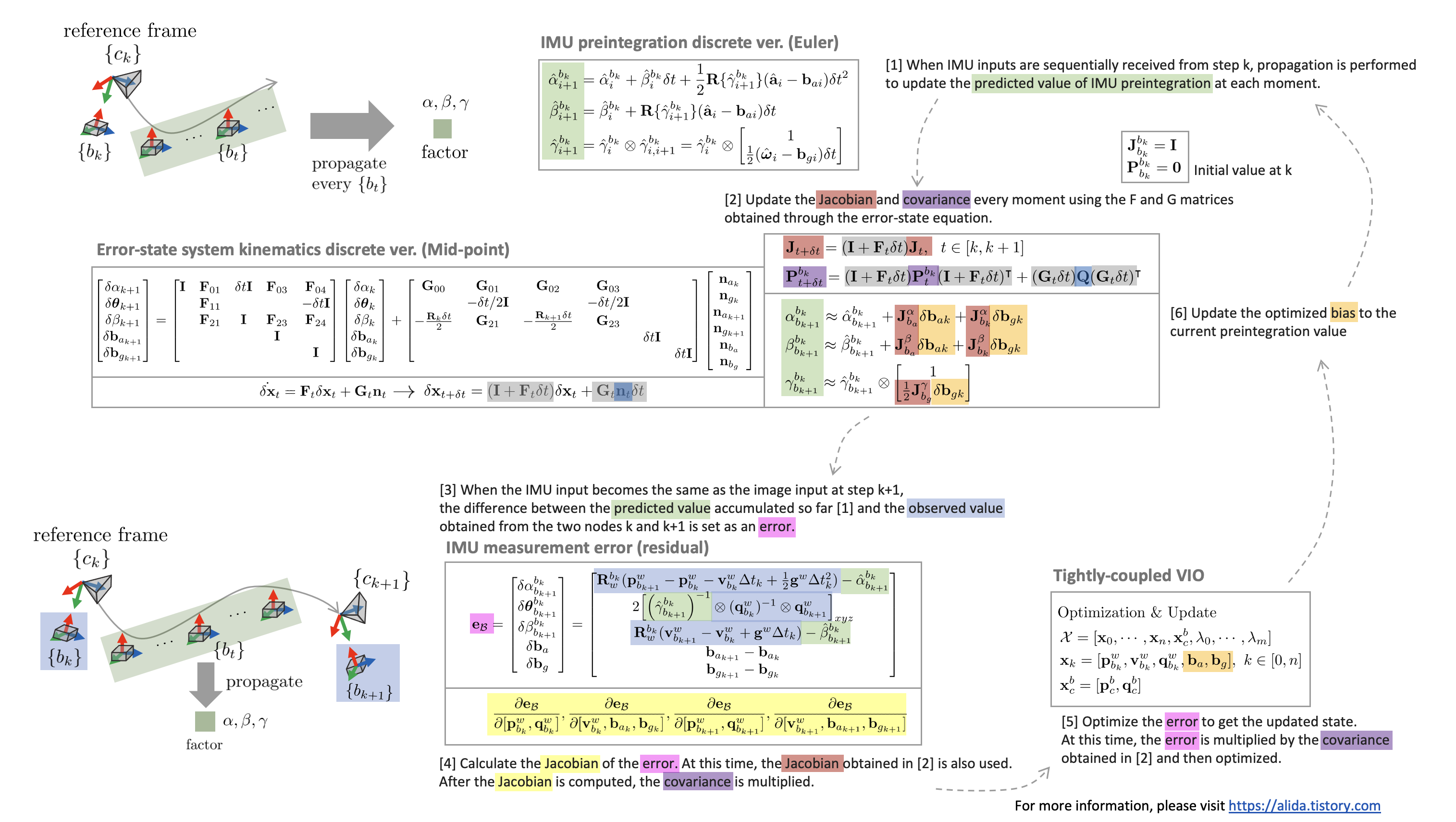}
	\label{fig:imu}
\end{figure}
To calculate the error in IMU measurement, it is first necessary to understand IMU preintegration techniques and error-state modeling. The figure above illustrates the overall IMU measurement error-based optimization process. Steps [1]-[6] should be followed in order. For more details, refer to \href{https://alida.tistory.com/64}{[SLAM] Formula Derivation and Analysis of VINS-mono content summary}.
~\\ ~\\ 
\textbf{NOMENCLATURE of IMU measurement error}
\begin{itemize}
	\item $\alpha_{b_{k+1}}^{b_{k}} \in \mathbb{R}^{3 \times 1}$: observed accumulated position during $t \in [b_{k}, b_{k+1}]$
	\item $\hat{\alpha}_{b_{k+1}}^{b_{k}}\in \mathbb{R}^{3 \times 1}$: predicted accumulated position during $t \in [b_{k}, b_{k+1}]$
	\item $\beta_{b_{k+1}}^{b_{k}}\in \mathbb{R}^{3 \times 1}$: observed accumulated velocity during $t \in [b_{k}, b_{k+1}]$
	\item $\hat{\beta}_{b_{k+1}}^{b_{k}}\in \mathbb{R}^{3 \times 1}$: predicted accumulated velocity during $t \in [b_{k}, b_{k+1}]$
	\item $\gamma_{b_{k+1}}^{b_{k}}\in \mathbb{R}^{3 \times 1}$: observed accumulated orientation during $t \in [b_{k}, b_{k+1}]$
	\item $\hat{\gamma}_{b_{k+1}}^{b_{k}}\in \mathbb{R}^{3 \times 1}$: predicted accumulated orientation during $t \in [b_{k}, b_{k+1}]$
	\item $\mathcal{X} = [\mathbf{x}_{0}, \mathbf{x}_{1}, \cdots, \mathbf{x}_{n}, \mathbf{x}^{b}_{c}, \lambda_{0}, \lambda_{1}, \cdots, \lambda_{m}]$: all state variables \\ 
	\item $\mathbf{x}_{k} = [\mathbf{p}^{w}_{b_{k}}, \mathbf{v}^{w}_{b_{k}}, \mathbf{q}^{w}_{b_{k}}, \mathbf{b}_{a}, \mathbf{b}_{g}]$: IMU model state variables at specific $k$ \\ 
	\item $\mathbf{x}^{b}_{c} = [\mathbf{p}^{b}_{c}, \mathbf{q}^{b}_{c}]$: extrinsic parameters of the camera and IMU
	\item $\mathcal{X}_{k}$: state variables for the specific two points $[b_{k}, b_{k+1}]$. This is thus $\mathcal{X}_{k} = (\mathbf{x}_{k}, \mathbf{x}_{k+1})$.
	\item $\lambda$: inverse depth of feature points
	\item $\otimes$: quaternion multiplication operator. (e.g., $\mathbf{q} = \mathbf{q}_{1} \otimes \mathbf{q}_{2}$)
	\item $\mathcal{B}$: set of all IMU $b_{k}$ values
	\item $\ominus$: operator for subtracting vectors and quaternions at once
	\item $\mathbf{P}_{\mathcal{B}}$: covariance of all IMU $b_{k}$ values
	\item $\boldsymbol{\Omega}_{\mathcal{B}}$: inverse matrix of covariance $\mathbf{P}_{\mathcal{B}}$. Represents the information matrix.
	\item $\mathbf{e}_{\mathcal{B},k} = \mathbf{e}_{\mathcal{B}}(\mathcal{X}_{k})$
\end{itemize}

\begin{figure}[h!]
	\centering
	\includegraphics[width=12cm]{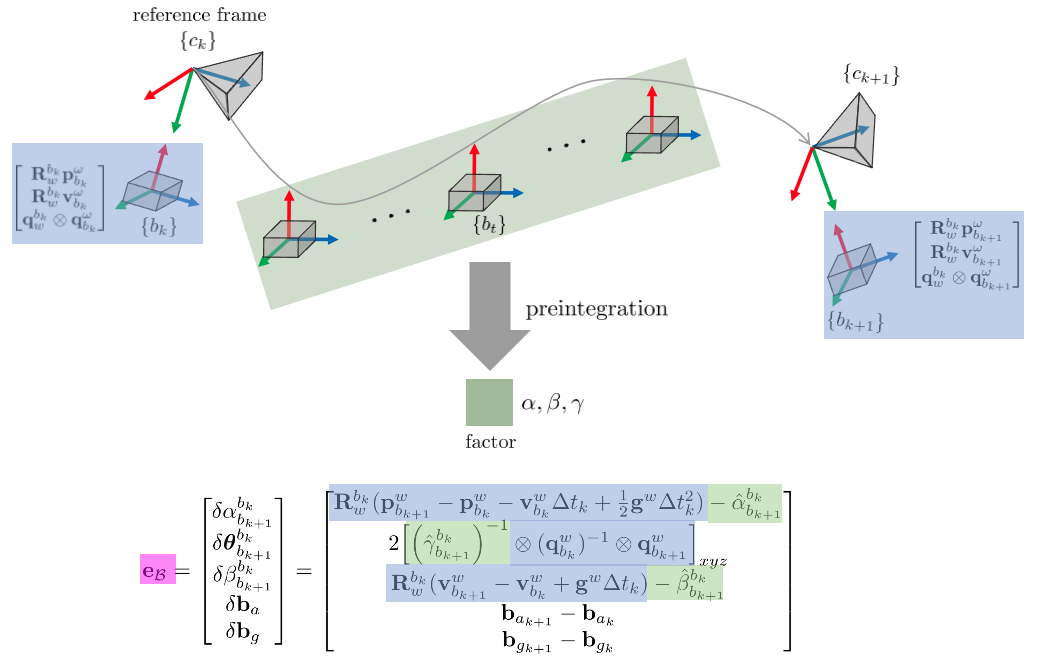}
	\label{fig:imu}
\end{figure}
IMU measurement error is defined as the difference between observed and predicted values, similar to the errors described in the previous section. \textbfazure{In detail, the IMU measurement error $\mathbf{e}_{\mathcal{B}}$ refers to the difference between the observed values $(\mathbf{z}_{b_{k+1}}^{b_{k}})$ and predicted values $(\hat{\mathbf{z}}_{b_{k+1}}^{b_{k}})$ of the accumulated IMU data and bias $[\alpha, \beta, \gamma, \mathbf{b}_{a}, \mathbf{b}_{g}]$ over the time $t \in [b_{k}, b_{k+1}]$.}
\begin{equation}
	\boxed{ \begin{aligned}
			\mathbf{e}_{\mathcal{B}}(\mathcal{X}_{k}) & = \mathbf{z}_{b_{k+1}}^{b_{k}} \ominus \hat{\mathbf{z}}_{b_{k+1}}^{b_{k}} = \begin{bmatrix}
				\alpha_{b_{k+1}}^{b_{k}} - \hat{\alpha}_{b_{k+1}}^{b_{k}} \\ \beta_{b_{k+1}}^{b_{k}} -\hat{\beta}_{b_{k+1}}^{b_{k}}\\ \gamma_{b_{k+1}}^{b_{k}} \otimes \hat{\gamma}_{b_{k+1}}^{b_{k}} \\ \mathbf{b}_{a_{k}} - \hat{\mathbf{b}}_{a} \\ \mathbf{b}_{g} - \hat{\mathbf{b}}_{g}
			\end{bmatrix} 
	\end{aligned} } \label{eq:imu3}
\end{equation}
	
	Let's look in detail at the observed and predicted values. \textbfazure{First, the observed values can be obtained using the positions $\mathbf{p}$, velocities $\mathbf{v}$, and orientations $\mathbf{q}$ at two points $b_{k}, b_{k+1}$.} The formula for IMU kinematics over the interval $[b_{k}, b_{k+1}]$ is as follows.
	\begin{equation}  
		\begin{aligned} 
			& \mathbf{R}^{b_{k}}_{w} \mathbf{p}^{w}_{b_{k+1}} = \mathbf{R}^{b_{k}}_{w} (\mathbf{p}^{w}_{b_{k}} + \mathbf{v}^{w}_{b_{k}}\Delta t - \frac{1}{2}\mathbf{g}^{w} \Delta t_{k}^{2}) + \alpha^{b_{k}}_{b_{k+1}} \\ 
			& \mathbf{R}^{b_{k}}_{w}\mathbf{v}^{w}_{b_{k+1}} = \mathbf{R}^{b_{k}}_{w} (\mathbf{v}^{w}_{b_{k}} - \mathbf{g}^{w} \Delta t_{k}) + \beta^{b_k}_{b_{k+1}} \\ 
			& \mathbf{q}^{b_{k}}_{w} \otimes \mathbf{q}^{w}_{b_{k+1}} = \gamma^{b_{k}}_{b_{k+1}} 
		\end{aligned}  
	\end{equation}
	
	Therefore, the observed values can be calculated as follows.
	\begin{equation} 
		\boxed{ \begin{aligned} 
				\mathbf{z}_{b_{k+1}}^{b_{k}}
				= \begin{bmatrix} 
					\alpha^{b_{k}}_{b_{k+1}} \\
					\beta^{b_{k}}_{b_{k+1}} \\
					\gamma^{b_{k}}_{b_{k+1}} \\
					\mathbf{b}_{a_{k+1}} - \mathbf{b}_{a_{k}} \\ 
					\mathbf{b}_{g_{k+1}} - \mathbf{b}_{g_{k}} \\ 
				\end{bmatrix} 
				= \begin{bmatrix} 
					\mathbf{R}^{b_{k}}_{w}(\mathbf{p}^{w}_{b_{k+1}} - \mathbf{p}^{w}_{b_{k}} - \mathbf{v}^{w}_{b_{k}}\Delta t_{k} + \frac{1}{2} \mathbf{g}^{w}\Delta t_{k}^{2}) \\ 
					\mathbf{R}^{b_{k}}_{w} ( \mathbf{v}^{w}_{b_{k+1}} - \mathbf{v}^{w}_{b_{k}} + \mathbf{g}^{w}\Delta t_{k})  \\ 
					(\mathbf{q}^{w}_{b_{k}})^{-1} \otimes \mathbf{q}^{w}_{b_{k+1}}  \\ 
					\mathbf{b}_{a_{k+1}} - \mathbf{b}_{a_{k}} \\ 
					\mathbf{b}_{g_{k+1}} - \mathbf{b}_{g_{k}} \\ 
				\end{bmatrix} 
		\end{aligned} }
	\end{equation}
	
	\textbfazure{Next, the predicted values can be obtained through the accumulated preintegration values during the time $t \in [b_{k}, b_{k+1}]$.} To calculate the predicted values using the preintegration formula, see the following.
	\begin{equation} 
		\begin{aligned} 
			& \hat{\alpha}^{b_{k}}_{b_{k+1}} = \iint_{t \in [k,k+1]} \mathbf{R}^{b_{k}}_{t} (\hat{\mathbf{a}}_{t} - \mathbf{b}_{at} - \mathbf{n}_{a}) dt^{2} \\ 
			& \hat{\beta}^{b_{k}}_{b_{k+1}} = \int_{t \in [k,k+1]} \mathbf{R}^{b_{k}}_{t} (\hat{\mathbf{a}}_{t} - \mathbf{b}_{at} -\mathbf{n}_{a}) dt \\ 
			& \hat{\gamma}^{b_{k}}_{b_{k+1}} = \int_{t \in [k,k+1]} \frac{1}{2} \Omega_{R}(\hat{\boldsymbol{\omega}}_{t} - \mathbf{b}_{gt} - \mathbf{n}_{g})\gamma^{b_{k}}_{t} dt 
		\end{aligned} 
	\end{equation}
	
	The above formula is applicable for continuous signals, but real IMU signals come as discrete signals, so the differential equation should be expressed as a difference equation. In this process, various numerical integration algorithms are used, such as zero-order hold (Euler), first-order hold (mid-point), and higher order (RK4). \textbfazure{Among these, the mid-point method used in VINS-mono is expressed as follows.}
	\begin{equation}
		\begin{aligned}
			\hat{\alpha}^{b_{k}}_{t+1} & = \hat{\alpha}^{b_{k}}_{t} + \frac{1}{2}(\hat{\beta}^{b_{k}}_{t} + \hat{\beta}^{b_{k}}_{t+1})\delta t \\
			& = \hat{\alpha}^{b_{k}}_{t} + \hat{\beta}^{b_{k}}_{t} \delta t + \frac{1}{4}[\mathbf{R}\{\hat{\gamma}^{b_{k}}_{t}\}(\hat{\mathbf{a}}_{t} - \mathbf{b}_{at}) + \mathbf{R}\{\hat{\gamma}^{b_{k}}_{t+1}\}(\hat{\mathbf{a}}_{t+1} - \mathbf{b}_{at})]\delta t^{2} \\  
			\hat{\beta}^{b_{k}}_{t+1} & = \hat{\beta}^{b_{k}}_{t} + \frac{1}{2}[\mathbf{R}\{\hat{\gamma}^{b_{k}}_{t}\}(\hat{\mathbf{a}}_{t} - \mathbf{b}_{at}) + \mathbf{R}\{\hat{\gamma}^{b_{k}}_{t+1}\}(\hat{\mathbf{a}}_{t+1} - \mathbf{b}_{at})]\delta t \\
			\hat{\gamma}^{b_{k}}_{t+1} & = \hat{\gamma}^{b_{k}}_{t} \otimes \hat{\gamma}^{b_{k}}_{t, t+1} = \hat{\gamma}^{b_{k}}_{t} \otimes \begin{bmatrix} 1 \\ 1/4(\hat{\boldsymbol{\omega}}_{t} + \hat{\boldsymbol{\omega}}_{t+1} - 2 \mathbf{b}_{gt}) \delta t \end{bmatrix} \\
		\end{aligned} \label{eq:imu1}
	\end{equation}
	
	Thus, the predicted values can be obtained as the accumulated values of (\ref{eq:imu1}) over the time $t \in [b_{k}, b_{k+1}]$. Since bias values cannot be predicted, they are set to zero.
	\begin{equation} 
		\boxed{ \begin{aligned} 
				\hat{\mathbf{z}}_{b_{k+1}}^{b_{k}}
				= \begin{bmatrix} \hat{\alpha}^{b_{k}}_{b_{k+1}} \\ \hat{\beta}^{b_{k}}_{b_{k+1}} \\ \hat{\gamma}^{b_{k}}_{b_{k+1}} \\ \mathbf{0} \\ \mathbf{0} \end{bmatrix} 
		\end{aligned} }
	\end{equation}
	
	Based on the values obtained so far, IMU measurement error can be represented as follows.
	\begin{equation} 
		\boxed{ \begin{aligned} 
				\mathbf{e}_{\mathcal{B}}(\mathcal{X}_{k})  =   \mathbf{z}_{b_{k+1}}^{b_{k}} \ominus \hat{\mathbf{z}}_{b_{k+1}}^{b_{k}} = \begin{bmatrix} 
					\mathbf{R}^{b_{k}}_{w}(\mathbf{p}^{w}_{b_{k+1}} - \mathbf{p}^{w}_{b_{k}} - \mathbf{v}^{w}_{b_{k}}\Delta t_{k} + \frac{1}{2} \mathbf{g}^{w}\Delta t_{k}^{2}) - \hat{\alpha}^{b_{k}}_{b_{k+1}} \\ 
					\mathbf{R}^{b_{k}}_{w} ( \mathbf{v}^{w}_{b_{k+1}} - \mathbf{v}^{w}_{b_{k}} + \mathbf{g}^{w}\Delta t_{k}) - \hat{\beta}^{b_{k}}_{b_{k+1}} \\ 
					\Big( \hat{\gamma}^{b_{k}}_{b_{k+1}} \Big)^{-1} \otimes (\mathbf{q}^{w}_{b_{k}})^{-1} \otimes \mathbf{q}^{w}_{b_{k+1}} \\ 
					\mathbf{b}_{a_{k+1}} - \mathbf{b}_{a_{k}} \\ 
					\mathbf{b}_{g_{k+1}} - \mathbf{b}_{g_{k}} \\ 
				\end{bmatrix} 
		\end{aligned}  } 
	\end{equation}
	
	\subsection{Error function formulation}
	The error function for all preintegrations and biases is defined as follows.
	\begin{equation}
		\begin{aligned}
			\mathbf{E}_{\mathcal{B}}(\mathcal{X}) & =  \sum_{k \in \mathcal{B}} \left\| \mathbf{e}_{\mathcal{B},k} \right\|_{\mathbf{P}_{\mathcal{B}}}^{2} \\
		\end{aligned} 
	\end{equation}
	\begin{equation}
		\begin{aligned}
			\mathcal{X}^{*} &= \arg\min_{\mathcal{X}^{*}} \mathbf{E}_{\mathcal{B}}(\mathcal{X}) \\
			& =  \arg\min_{\mathcal{X}^{*}} \sum_{k \in \mathcal{B}} \left\| \mathbf{e}_{\mathcal{B},k} \right\|_{\mathbf{P}_{\mathcal{B}}}^{2} \\
			& = \arg\min_{\mathcal{X}^{*}} \sum_{k \in \mathcal{B}} \mathbf{e}_{_{\mathcal{B}},k}^{\intercal}\boldsymbol{\Omega}_{\mathcal{B}}\mathbf{e}_{_{\mathcal{B}},k} \\
			& = \arg\min_{\mathcal{X}^{*}} \sum_{k \in \mathcal{B}} (\mathbf{z}_{b_{k+1}}^{b_{k}} \ominus \hat{\mathbf{z}}_{b_{k+1}}^{b_{k}})^{\intercal} \boldsymbol{\Omega}_{\mathcal{B}} (\mathbf{z}_{b_{k+1}}^{b_{k}} \ominus \hat{\mathbf{z}}_{b_{k+1}}^{b_{k}})
		\end{aligned} 
	\end{equation}
	The formula $\mathbf{e}_{\mathcal{B},k} = \mathbf{e}_{\mathcal{B}}(\mathcal{X}_{k})$ is implied here.
	
	\mybox{Tip}{gray!40}{gray!10}{
		In actual VINS-mono implementation, not only the IMU measurement error but also the visual residual $\mathbf{r}_{\mathcal{C}}$, marginalization prior residual $\mathbf{r}_{p}$ are simultaneously optimized to perform tightly-coupled VIO. In VINS-mono, the IMU measurement error is expressed as the residual $\mathbf{r}_{\mathcal{B}}(\hat{\mathbf{z}}_{b_{k+1}}^{b_{k}}, \mathcal{X})$.
		\begin{equation} 
			\begin{aligned} 
				\min_{\mathcal{X}} \bigg\{ \left\| \mathbf{r}_{p} - \mathbf{J}_{p}\mathcal{X} \right\|_{\mathbf{P}_{M}} + \sum_{k \in \mathcal{B}} \left\| \mathbf{r}_{\mathcal{B}} \Big( \hat{\mathbf{z}}^{b_{k}}_{b_{k+1}}, \mathcal{X} \Big) \right\|_{\mathbf{P}_{B}} + \sum_{(l,j) \in \mathcal{C}} \left\| \mathbf{r}_{\mathcal{C}} \Big( \hat{\mathbf{z}}^{c_{j}}_{l}, \mathcal{X} \Big) \right\|_{\mathbf{P}^{c_{j}}_{l}} \bigg\} 
			\end{aligned} 
		\end{equation}
		This section explains only the IMU measurement error $\mathbf{r}_{\mathcal{B}}(\hat{\mathbf{z}}_{b_{k+1}}^{b_{k}}, \mathcal{X})$.
	}
	
	$\mathbf{E}_{\mathcal{B}}(\mathcal{X}^{*})$ that satisfies $\left\|\mathbf{e}_{\mathcal{B}}(\mathcal{X}_{k}^{*})\right\|_{\mathbf{P}_{\mathcal{B}}}^{2}$ can be iteratively computed through non-linear least squares. Small increments $\Delta \mathcal{X}$ are iteratively updated to $\mathcal{X}$ to find the optimal state.
	\begin{equation}
		\begin{aligned}
			\arg\min_{\mathcal{X}^{*}} \mathbf{E}_{\mathcal{B}}(\mathcal{X} + \Delta \mathcal{X}) & = \arg\min_{\mathcal{X}^{*}} \sum_{k \in \mathcal{B}} \left\|\mathbf{e}_{\mathcal{B}}(\mathcal{X}_{k} +\Delta \mathcal{X}_{k})\right\|^{2} 
		\end{aligned}
	\end{equation}
	
	Strictly speaking, since the state increment $\Delta \mathcal{X}$ includes quaternions, it should be added to the existing state $\mathcal{X}$ using the $\oplus$ operator, but the $+$ operator is used for simplicity of expression.
	\begin{equation}
		\begin{aligned}
			\mathbf{e}_{\mathcal{B}}( \mathcal{X}_{k} \oplus \Delta \mathcal{X}_{k}) 
			\quad \rightarrow \quad\mathbf{e}_{\mathcal{B}}(\mathcal{X}_{k} + \Delta \mathcal{X}_{k})
		\end{aligned}
	\end{equation}
	
	The above equation can be expressed through a first-order Taylor approximation as follows.
	\begin{equation}
		\begin{aligned}
			\mathbf{e}_{\mathcal{B}}(\mathcal{X}_{k} + \Delta \mathcal{X}_{k})  & \approx \mathbf{e}_{\mathcal{B}}(\mathcal{X}) + \mathbf{J}\Delta \mathcal{X}_{k} \\ 
			& = \mathbf{e}_{\mathcal{B}}(\mathcal{X}_{k}) + \begin{bmatrix}
				\frac{\partial \mathbf{e}_{\mathcal{B}}}{\partial [\mathbf{p}^{w}_{b_{k}}, \mathbf{q}^{w}_{b_{k}}]} & \frac{\partial \mathbf{e}_{\mathcal{B}}}{\partial [\mathbf{v}^{w}_{b_{k}}, \mathbf{b}_{ak}, \mathbf{b}_{gk}]} & \frac{\partial \mathbf{e}_{\mathcal{B}}}{\partial [\mathbf{p}^{w}_{b_{k+1}}, \mathbf{q}^{w}_{b_{k+1}}]} & \frac{\partial \mathbf{e}_{\mathcal{B}}}{\partial [\mathbf{v}^{w}_{b_{k+1}}, \mathbf{b}_{ak+1}, \mathbf{b}_{ak+1}]}
			\end{bmatrix} \begin{bmatrix}
				\Delta \mathbf{p}^{w}_{k} \\ \Delta \mathbf{q}^{w}_{k} \\ \Delta \mathbf{v}^{w}_{k} \\ \Delta \mathbf{b}_{ak} \\ \Delta \mathbf{b}_{gk} \\ \Delta \mathbf{p}^{w}_{k+1} \\ \Delta \mathbf{q}^{w}_{k+1} \\ \Delta \mathbf{v}^{w}_{k+1} \\ \Delta \mathbf{b}_{ak+1} \\ \Delta \mathbf{b}_{gk+1} 
			\end{bmatrix} \\ 
			& = \mathbf{e}_{\mathcal{B}}(\mathcal{X}_{k}) 
			+ \frac{\partial \mathbf{e}_{\mathcal{B}}}{\partial [\mathbf{p}^{w}_{b_{k}}, \mathbf{q}^{w}_{b_{k}}]} (\Delta \mathbf{p}^{w}_{k}, \Delta \mathbf{q}^{w}_{k})
			+ \frac{\partial \mathbf{e}_{\mathcal{B}}}{\partial [\mathbf{v}^{w}_{b_{k}}, \mathbf{b}_{ak}, \mathbf{b}_{gk}]} (\Delta \mathbf{v}^{w}_{k} , \Delta \mathbf{b}_{ak} ,\Delta \mathbf{b}_{gk})
			\\ 
			& + \frac{\partial \mathbf{e}_{\mathcal{B}}}{\partial [\mathbf{p}^{w}_{b_{k+1}}, \mathbf{q}^{w}_{b_{k+1}}]} (\Delta \mathbf{p}^{w}_{k+1} , \Delta \mathbf{q}^{w}_{k+1})
			+ \frac{\partial \mathbf{e}_{\mathcal{B}}}{\partial [\mathbf{v}^{w}_{b_{k+1}}, \mathbf{b}_{ak+1}, \mathbf{b}_{ak+1}]} (\Delta \mathbf{v}^{w}_{k+1} , \Delta \mathbf{b}_{ak+1} , \Delta \mathbf{b}_{gk+1}) 
		\end{aligned}
	\end{equation}
	
	Both $[\mathbf{p}^{w}_{b_{k}},\mathbf{v}^{w}_{b_{k}},\mathbf{q}^{w}_{b_{k}}, \mathbf{b}_{a_{k}}, \mathbf{b}_{g_{k}}]$ at the point $b_{k}$ and $[\mathbf{p}^{w}_{b_{k+1}},\mathbf{v}^{w}_{b_{k+1}},\mathbf{q}^{w}_{b_{k+1}}, \mathbf{b}_{a_{k+1}}, \mathbf{b}_{g_{k+1}}]$ at the point $b_{k+1}$ are involved in the error value, so the Jacobian for all 10 variables must be calculated. In VINS-mono, state variables are grouped into 4 groups as follows.
	\begin{equation}
		\begin{aligned}
			&    [\mathbf{p}^{w}_{b_{k}}, \mathbf{q}^{w}_{b_{k}}] \quad &&\cdots \text{ for } \mathbf{J}[0] \\
			&    [\mathbf{v}^{w}_{b_{k}}, \mathbf{b}_{ak}, \mathbf{b}_{gk}] \quad &&\cdots \text{ for } \mathbf{J}[1] \\
			&    [\mathbf{p}^{w}_{b_{k+1}}, \mathbf{q}^{w}_{b_{k+1}}] \quad &&\cdots \text{ for } \mathbf{J}[2] \\
			&    [\mathbf{v}^{w}_{b_{k+1}}, \mathbf{b}_{ak+1}, \mathbf{b}_{ak+1}] \quad &&\cdots \text{ for } \mathbf{J}[3] \\
		\end{aligned}
	\end{equation}
	
	\textbfazure{Tightly-coupled VIO optimizes the state variables $\mathcal{X}$ which include the inverse depth $\lambda$ and external parameters (extrinsic parameters) $\mathbf{x}^{b}_{c}$, time difference $td$, but it is important to note that in the IMU measurement error, only pose, velocity, and bias values for two points $[b_{k}, b_{k+1}]$ are updated.}
	
	The error function can be approximated as follows.
	\begin{equation}
		\begin{aligned}
			\arg\min_{\mathcal{X}^{*}}  \mathbf{E}_{\mathcal{B}}(\mathcal{X} + \Delta \mathcal{X}) & \approx \arg\min_{\mathcal{X}^{*}} \sum_{k \in \mathcal{B}} \left\|\mathbf{e}_{\mathcal{B}}(\mathcal{X}_{k}) + \mathbf{J}\Delta \mathcal{X}_{k} \right\|_{\mathbf{P}_{\mathcal{B}}}^{2} \\
		\end{aligned}
	\end{equation}
	
	Differentiating this to find the optimal increment $\Delta \mathcal{X}^{*}$ results in the following. The detailed derivation process is omitted in this section. For a detailed derivation, refer to \hyperref[sec:opt]{the previous section}.
	\begin{equation}
		\begin{aligned}
			& \mathbf{J}^{\intercal}\mathbf{J} \Delta \mathcal{X}^{*} = -\mathbf{J}^{\intercal}\mathbf{e} \\ 
			& \mathbf{H}\Delta \mathcal{X}^{*} = - \mathbf{b} \\
		\end{aligned} \label{eq:imu2}
	\end{equation}
	
	This equation is in the form of a linear system $\mathbf{Ax} = \mathbf{b}$, so $\Delta \mathcal{X}^{*}$ can be found using various linear algebra techniques such as schur complement, cholesky decomposition. The optimal increment found in this way is added to the current state. \textbfazure{In this case, whether the existing state $\mathbf{x}$ is multiplied on the right or left determines whether the pose viewed from the local coordinate system is updated (right) or the pose viewed from the global coordinate system is updated (left). Since IMU measurement error is related to two nodes $b_{k}, b_{k+1}$, right multiplication applicable to local coordinate system updates is applied.}
	\begin{equation}
		\begin{aligned}
			\mathcal{X} \leftarrow  \mathcal{X} \oplus \Delta \mathcal{X}^{*} \\
		\end{aligned}
	\end{equation}
	
	$\mathcal{X}$ being updated by IMU measurement error $\mathcal{X}_{k}$ consists of $[\mathbf{p}^{w}_{b_{k}},\mathbf{v}^{w}_{b_{k}},\mathbf{q}^{w}_{b_{k}}, \mathbf{b}_{a_{k}}, \mathbf{b}_{g_{k}}, \mathbf{p}^{w}_{b_{k+1}},\mathbf{v}^{w}_{b_{k+1}},\mathbf{q}^{w}_{b_{k+1}}, \mathbf{b}_{a_{k+1}}, \mathbf{b}_{g_{k+1}}]$ so it can be expressed as follows.
	\begin{equation}
		\begin{aligned}
			& \mathbf{p}^{w}_{b_{k}} \leftarrow  \mathbf{p}^{w}_{b_{k}}   \oplus \Delta \mathbf{p}^{w*}_{b_{k}} \\
			& \mathbf{q}^{w}_{b_{k}} \leftarrow  \mathbf{q}^{w}_{b_{k}}   \oplus \Delta \mathbf{q}^{w*}_{b_{k}} \\
			& \mathbf{v}^{w}_{b_{k}} \leftarrow  \mathbf{v}^{w}_{b_{k}}   \oplus \Delta \mathbf{v}^{w*}_{b_{k}} \\
			& \mathbf{b}_{ak} \leftarrow  \mathbf{b}_{ak}   \oplus \Delta \mathbf{b}_{ak}^{*} \\
			& \mathbf{b}_{gk} \leftarrow  \mathbf{b}_{gk}   \oplus \Delta \mathbf{b}_{gk}^{*} \\ 
			& \mathbf{p}^{w}_{b_{k+1}} \leftarrow  \mathbf{p}^{w}_{b_{k+1}}   \oplus \Delta \mathbf{p}^{w*}_{b_{k+1}} \\
			& \mathbf{q}^{w}_{b_{k+1}} \leftarrow  \mathbf{q}^{w}_{b_{k+1}}  \oplus \Delta \mathbf{q}^{w*}_{b_{k+1}} \\
			& \mathbf{v}^{w}_{b_{k+1}} \leftarrow  \mathbf{v}^{w}_{b_{k+1}}   \oplus \Delta \mathbf{v}^{w*}_{b_{k+1}} \\
			& \mathbf{b}_{ak+1} \leftarrow  \mathbf{b}_{ak+1}   \oplus \Delta \mathbf{b}_{ak+1}^{*} \\
			& \mathbf{b}_{gk+1} \leftarrow  \mathbf{b}_{gk+1}  \oplus \Delta \mathbf{b}_{gk+1}^{*} \\ 
		\end{aligned} 
	\end{equation}
	
	Right multiplication $\oplus$ operation definition is as follows.
	\begin{equation}
		\begin{aligned}
			& \mathbf{p}^{w}_{b_{k}} \leftarrow  \mathbf{p}^{w}_{b_{k}}   + \Delta \mathbf{p}^{w*}_{b_{k}} \\
			& \mathbf{q}^{w}_{b_{k}} \leftarrow  \mathbf{q}^{w}_{b_{k}}   \otimes \Delta \mathbf{q}^{w*}_{b_{k}} \quad \cdots \text{ locally updated (right mult)} \\
			& \mathbf{v}^{w}_{b_{k}} \leftarrow  \mathbf{v}^{w}_{b_{k}}   + \Delta \mathbf{v}^{w*}_{b_{k}} \\
			& \mathbf{b}_{ak} \leftarrow  \mathbf{b}_{ak}   + \Delta \mathbf{b}_{ak}^{*} \\
			& \mathbf{b}_{gk} \leftarrow  \mathbf{b}_{gk}   + \Delta \mathbf{b}_{gk}^{*} \\ 
			& \mathbf{p}^{w}_{b_{k+1}} \leftarrow  \mathbf{p}^{w}_{b_{k+1}}   + \Delta \mathbf{p}^{w*}_{b_{k+1}} \\
			& \mathbf{q}^{w}_{b_{k+1}} \leftarrow  \mathbf{q}^{w}_{b_{k+1}}   \otimes \Delta \mathbf{q}^{w*}_{b_{k+1}}  \quad \cdots \text{ locally updated (right mult)} \\
			& \mathbf{v}^{w}_{b_{k+1}} \leftarrow  \mathbf{v}^{w}_{b_{k+1}}   + \Delta \mathbf{v}^{w*}_{b_{k+1}} \\
			& \mathbf{b}_{ak+1} \leftarrow  \mathbf{b}_{ak+1}   + \Delta \mathbf{b}_{ak+1}^{*} \\
			& \mathbf{b}_{gk+1} \leftarrow  \mathbf{b}_{gk+1}   + \Delta \mathbf{b}_{gk+1}^{*} \\ 
		\end{aligned} 
	\end{equation}
	
	\subsection{Jacobian of IMU measurement error}
	To perform (\ref{eq:imu2}), the Jacobian $\mathbf{J}$ for the IMU measurement error must be calculated. It can be represented as follows.
	\begin{equation}
		\begin{aligned}
			\mathbf{J} & = \begin{bmatrix} \mathbf{J}[0] & \mathbf{J}[1] & \mathbf{J}[2] & \mathbf{J}[3] \end{bmatrix}   \\
			& = \begin{bmatrix}
				\frac{\partial \mathbf{e}_{\mathcal{B}}}{\partial [\mathbf{p}^{w}_{b_{k}}, \mathbf{q}^{w}_{b_{k}}]} & \frac{\partial \mathbf{e}_{\mathcal{B}}}{\partial [\mathbf{v}^{w}_{b_{k}}, \mathbf{b}_{ak}, \mathbf{b}_{gk}]} & \frac{\partial \mathbf{e}_{\mathcal{B}}}{\partial [\mathbf{p}^{w}_{b_{k+1}}, \mathbf{q}^{w}_{b_{k+1}}]} & \frac{\partial \mathbf{e}_{\mathcal{B}}}{\partial [\mathbf{v}^{w}_{b_{k+1}}, \mathbf{b}_{ak+1}, \mathbf{b}_{ak+1}]}
			\end{bmatrix} \\
			& = \frac{\partial }{\partial [\mathbf{p}^{w}_{b_{k}}, \mathbf{q}^{w}_{b_{k}}], [\mathbf{v}^{w}_{b_{k}}, \mathbf{b}_{ak}, \mathbf{b}_{gk}], [\mathbf{p}^{w}_{b_{k+1}}, \mathbf{q}^{w}_{b_{k+1}}], [\mathbf{v}^{w}_{b_{k+1}}, \mathbf{b}_{ak+1}, \mathbf{b}_{ak+1}]} 
			\begin{bmatrix} 
				\mathbf{R}^{b_{k}}_{w}(\mathbf{p}^{w}_{b_{k+1}} - \mathbf{p}^{w}_{b_{k}} - \mathbf{v}^{w}_{b_{k}}\Delta t_{k} + \frac{1}{2} \mathbf{g}^{w}\Delta t_{k}^{2}) - \hat{\alpha}^{b_{k}}_{b_{k+1}} \\ 
				\mathbf{R}^{b_{k}}_{w} ( \mathbf{v}^{w}_{b_{k+1}} - \mathbf{v}^{w}_{b_{k}} + \mathbf{g}^{w}\Delta t_{k}) - \hat{\beta}^{b_{k}}_{b_{k+1}} \\ 
				\Big( \hat{\gamma}^{b_{k}}_{b_{k+1}} \Big)^{-1} \otimes (\mathbf{q}^{w}_{b_{k}})^{-1} \otimes \mathbf{q}^{w}_{b_{k+1}} \\ 
				\mathbf{b}_{a_{k+1}} - \mathbf{b}_{a_{k}} \\ 
				\mathbf{b}_{g_{k+1}} - \mathbf{b}_{g_{k}} \\ 
			\end{bmatrix}  \\
			& = \begin{bmatrix} \mathbb{R}^{15 \times 7} &&  \mathbb{R}^{15 \times 9} && \mathbb{R}^{15 \times 7} && \mathbb{R}^{15 \times 9} \end{bmatrix} = \mathbb{R}^{15 \times 32}
		\end{aligned}
	\end{equation}
	
	\subsubsection{Lie theory-based SO(3) optimization} 
	\textbfazure{When calculating the above Jacobian, terms related to position $\mathbf{p}$, velocity $\mathbf{v}$, and biases $\mathbf{b}_{a}, \mathbf{b}_{g}$ are each 3-dimensional vectors, so they do not have any constraints when performing optimization updates. However, the quaternion $\mathbf{q}$ has 4 parameters and represents 3 degrees of freedom, which is more than the minimal degrees of freedom required to represent 3-dimensional rotation, thus having various constraints.} This is known as being over-parameterized. The disadvantages of over-parameterized representation include:
	\begin{itemize}
		\item Increased computation due to redundant parameters during optimization.
		\item Potential numerical instability issues due to additional degrees of freedom.
		\item The need to ensure that constraints are met every time parameters are updated.
	\end{itemize}
	
	Using lie theory, optimization can be performed free from constraints. Therefore, instead of using quaternion $\mathbf{q}$, lie algebra so(3) $[\boldsymbol{\theta}]_{\times}$ is used, freeing parameters from constraints. Here, $\boldsymbol{\theta} \in \mathbb{R}^{3}$ represents the angular velocity vector. Detailed content on SO(3)-based optimization is the same as in the \hyperref[sec:reproj-so3]{reprojection error section} and is omitted here.
	
	When using angular velocity vector $\boldsymbol{\theta}$, the original Jacobian of quaternion $\mathbf{q}$ is changed as follows.
	\begin{equation}
		\begin{aligned}
			& \frac{\partial \mathbf{e}_{\mathcal{B}}}{\partial \mathbf{q}^{w}_{b_{k}}} \quad \rightarrow \quad \frac{\partial \mathbf{e}_{\mathcal{B}}}{\partial \begin{bmatrix} 1 & \frac{1}{2}\boldsymbol{\theta}^{w}_{b_{k}} \end{bmatrix}} \\ 
			& \frac{\partial \mathbf{e}_{\mathcal{B}}}{\partial \mathbf{q}^{w}_{b_{k+1}}} \quad \rightarrow \quad \frac{\partial \mathbf{e}_{\mathcal{B}}}{\partial \begin{bmatrix} 1 & \frac{1}{2}\boldsymbol{\theta}^{w}_{b_{k+1}} \end{bmatrix}}
		\end{aligned}
	\end{equation}
	
	\mybox{Tip}{gray!40}{gray!10}{
		Given an arbitrary angle-axis vector $\boldsymbol{\theta} = \theta \mathbf{u}$, its corresponding exponential map can be expressed using an extended version of Euler's formula.
		\begin{equation} 
			\mathbf{q} \triangleq \text{Exp}(\boldsymbol{\theta}) = \text{Exp}(\theta \mathbf{u}) = e^{\theta \mathbf{u}/2} = \cos\frac{\theta}{2} + \mathbf{u}\sin\frac{\theta}{2} = \begin{bmatrix} 
				\cos(\theta/2) \\ \mathbf{u} \sin(\theta/2) 
			\end{bmatrix} 
		\end{equation}
		
		For sufficiently small $\boldsymbol{\theta}$ values, $\cos\frac{\theta}{2} \approx 1$ and $\sin\frac{\theta}{2} \approx \frac{\theta}{2}$ hold true, thus the following formula for sufficiently small quaternion values is valid.
		\begin{equation} 
			\mathbf{q} \approx \begin{bmatrix} 
				1 \\ \frac{1}{2}  \boldsymbol{\theta}
			\end{bmatrix} 
		\end{equation}
		
		More details can be found in the \href{https://alida.tistory.com/61\#4.4-perturbations,-uncertainties,-noise}{Quaternion kinematics for the error-state Kalman filter content summary} post, section 4.4.
	}
	
	\textbfazure{Typically, the errors used in optimization are small, so it is assumed that the error $\Big( \hat{\gamma}^{b_{k}}_{b_{k+1}} \Big)^{-1} \otimes (\mathbf{q}^{w}_{b_{k}})^{-1} \otimes \mathbf{q}^{w}_{b_{k+1}}$ is also small. Therefore, only the imaginary part $[x,y,z] = \frac{1}{2}\boldsymbol{\theta}$ of the actual quaternion $\mathbf{q} = [w,x,y,z]$ is used in optimization.} Through this, the $\gamma$ part is transformed as follows.
	\begin{equation}
		\begin{aligned}
			\gamma \quad & \rightarrow \quad 2[\gamma]_{xyz} = 2[x,y,z]  = \boldsymbol{\theta} \\ 
			\Big( \hat{\gamma}^{b_{k}}_{b_{k+1}} \Big)^{-1} \otimes (\mathbf{q}^{w}_{b_{k}})^{-1} \otimes \mathbf{q}^{w}_{b_{k+1}} \quad & \rightarrow \quad 2 \bigg[ \Big( \hat{\gamma}^{b_{k}}_{b_{k+1}} \Big)^{-1} \otimes (\mathbf{q}^{w}_{b_{k}})^{-1} \otimes \mathbf{q}^{w}_{b_{k+1}} \bigg]_{xyz}
		\end{aligned}
	\end{equation}
	
	The final SO(3) version IMU measurement error $\mathbf{e}_{\mathcal{B}}$ is as follows.
	\begin{equation}
		\boxed{\begin{aligned}
				\mathbf{e}_{\mathcal{B}}(\mathcal{X}_{k}) = \begin{bmatrix} 
					\mathbf{R}^{b_{k}}_{w}(\mathbf{p}^{w}_{b_{k+1}} - \mathbf{p}^{w}_{b_{k}} - \mathbf{v}^{w}_{b_{k}}\Delta t_{k} + \frac{1}{2} \mathbf{g}^{w}\Delta t_{k}^{2}) - \hat{\alpha}^{b_{k}}_{b_{k+1}} \\ 
					2 \bigg[ \Big( \hat{\gamma}^{b_{k}}_{b_{k+1}} \Big)^{-1} \otimes (\mathbf{q}^{w}_{b_{k}})^{-1} \otimes \mathbf{q}^{w}_{b_{k+1}} \bigg]_{xyz}\\ 
					\mathbf{R}^{b_{k}}_{w} ( \mathbf{v}^{w}_{b_{k+1}} - \mathbf{v}^{w}_{b_{k}} + \mathbf{g}^{w}\Delta t_{k}) - \hat{\beta}^{b_{k}}_{b_{k+1}} \\ 
					\mathbf{b}_{a_{k+1}} - \mathbf{b}_{a_{k}} \\ 
					\mathbf{b}_{g_{k+1}} - \mathbf{b}_{g_{k}} \\ 
				\end{bmatrix}
		\end{aligned} }
	\end{equation}
	
	For easier calculation of Jacobians for $[\mathbf{p}, \mathbf{q}], [\mathbf{v}, \mathbf{b}_{a}, \mathbf{b}_{g}]$, the order of the second line $\beta$ and the third line $\gamma$ in the original state variables was switched.
	
	The final SO(3) version IMU measurement error Jacobian can be calculated as follows. The detailed derivation process can be referred to in the \href{https://arxiv.org/pdf/1912.11986.pdf}{Formula Derivation and Analysis of VINS-Mono} paper's Appendix section.
	\begin{equation} 
		\boxed{ \begin{aligned} 
				\mathbf{J}[0]_{15\times 6} = \frac{\partial \mathbf{e}_{\mathcal{B}}}{\partial [\mathbf{p}^{w}_{b_{k}}, \mathbf{q}^{w}_{b_{k}}]} = \begin{bmatrix} 
					-\mathbf{R}^{b_{k}}_{w} & [\mathbf{R}^{b_{k}}_{w}(\mathbf{p}^{w}_{b_{k+1}} - \mathbf{p}^{w}_{b_{k}} - \mathbf{v}^{w}_{b_{k}}\Delta t_{k} + \frac{1}{2} \mathbf{g}^{w}\Delta t_{k}^{2})]_{\times} \\ 
					0 & [\gamma^{b_{k}}_{b_{k+1}}]_{R}[(\mathbf{q}^{w}_{b_{k+1}})^{-1} \otimes \mathbf{q}^{b_{k}}_{w}]_{L, 3\times3} \\ 
					0 & [\mathbf{R}^{b_{k}}_{w} (\mathbf{p}^{w}_{b_{k+1}} - \mathbf{p}^{w}_{b_{k}} + \mathbf{g}^{w} \Delta t_{k})]_{\times} \\ 
					0&0\\ 
					0&0 
				\end{bmatrix} 
		\end{aligned} }
	\end{equation}
	
	\begin{equation} 
		\boxed{ \begin{aligned} 
				\mathbf{J}[1]_{15\times9} = \frac{\partial \mathbf{e}_{\mathcal{B}} }{\partial [\mathbf{v}^{w}_{b_{k}}, \mathbf{b}_{a_{k}}, \mathbf{b}_{g_{k}}]} = \begin{bmatrix} 
					-\mathbf{R}^{b_{k}}_{w} \Delta t_{k} & -\mathbf{J}^{\alpha}_{b_{a}} & -\mathbf{J}^{\alpha}_{b_{g}} \\ 
					0&0& -[(\hat{\gamma}^{b_{k}}_{b_{k+1}})^{-1} \otimes (\mathbf{q}^{w}_{b_{k}})^{-1} \otimes \mathbf{q}^{w}_{b_{k+1}}]_{R,3\times3} \mathbf{J}^{\gamma}_{b_{g}} \\ 
					-\mathbf{R}^{b_{k}}_{w} & -\mathbf{J}^{\beta}_{b_{a}} & -\mathbf{J}^{\beta}_{b_{g}} \\ 
					0 & - \mathbf{I} & 0 \\ 
					0 & 0 &-\mathbf{I} 
				\end{bmatrix} 
		\end{aligned} }
	\end{equation}
	
	\begin{equation} 
		\boxed{ \begin{aligned} 
				\mathbf{J}[2]_{15\times 6} = \frac{\partial \mathbf{e}_{\mathcal{B}} }{\partial [\mathbf{p}^{w}_{b_{k+1}}, \mathbf{q}^{w}_{b_{k+1}}]} = \begin{bmatrix} 
					\mathbf{R}^{b_{k}}_{w} & 0 \\ 
					0 & [(\hat{\gamma}^{b_{k}}_{b_{k+1}})^{-1} \otimes (\mathbf{q}^{w}_{b_{k}})^{-1} \otimes \mathbf{q}^{w}_{b_{k+1}}]_{L} \\ 
					0&0\\ 
					0&0\\ 
					0&0 
				\end{bmatrix} 
		\end{aligned} }
	\end{equation}
	
	\begin{equation} 
		\boxed{ \begin{aligned} 
				\mathbf{J}[3]_{15\times9} = \frac{\partial \mathbf{e}_{\mathcal{B}} }{\partial [\mathbf{v}^{w}_{b_{k+1}}, \mathbf{b}_{a_{k+1}}, \mathbf{b}_{g_{k+1}}]} = \begin{bmatrix} 
					0&0&0\\ 
					0&0&0\\ 
					\mathbf{R}^{b_{k}}_{w} &0&0 \\ 
					0&\mathbf{I}&0 \\ 
					0&0&\mathbf{I} 
				\end{bmatrix} 
		\end{aligned} }
	\end{equation}
	
	~\\
	\textbfazure{NOTICE: } The original $\mathbf{J}[0], \mathbf{J}[2] \in \mathbb{R}^{15 \times 7}$, but since quaternion is updated based on SO(3) using only $[xyz]$ part, $w$ part is always 0. By omitting the $w$ part, $\mathbf{J}[0], \mathbf{J}[2] \in \mathbb{R}^{15 \times 6}$.
	~\\ 
	\textbfazure{NOTICE: } Looking at the above formula, it can be seen that another Jacobian $\mathbf{J}^{\alpha}_{b_{a}}, \mathbf{J}^{\alpha}_{b_{g}}, \mathbf{J}^{\beta}_{b_{a}}, \mathbf{J}^{\beta}_{b_{g}} \mathbf{J}^{\gamma}_{b_{g}}$ is used within the Jacobian. This refers to the partial Jacobians derived from the error-state equations of the IMU $\mathbf{J}^{b_{k}}_{b_{k+1}}$.
	
	\mybox{Tip}{gray!40}{gray!10}{
		The error-state equation for the discrete IMU signal is as follows. (Using Mid-point approximation)
		\begin{equation} 
			\begin{aligned} 
				\begin{bmatrix} 
					\delta \alpha_{k+1} \\ 
					\delta \boldsymbol{\theta}_{k+1} \\ 
					\delta \beta_{k+1} \\ 
					\delta \mathbf{b}_{a_{k+1}} \\ 
					\delta \mathbf{b}_{g_{k+1}} \\ 
				\end{bmatrix} 
				= \begin{bmatrix} 
					\mathbf{I} & \mathbf{F}_{01} & \delta t \mathbf{I} & \mathbf{F}_{03} & \mathbf{F}_{04} \\ 
					& \mathbf{F}_{11} &  &  & -\delta t \mathbf{I} \\ 
					& \mathbf{F}_{21} & \mathbf{I} & \mathbf{F}_{23} & \mathbf{F}_{24} \\ 
					&&&\mathbf{I}& \\ 
					&&&&\mathbf{I} 
				\end{bmatrix}  
				\begin{bmatrix} 
					\delta \alpha_{k} \\ 
					\delta \boldsymbol{\theta}_{k} \\ 
					\delta \beta_{k} \\ 
					\delta \mathbf{b}_{a_{k}} \\ 
					\delta \mathbf{b}_{g_{k}} \\ 
				\end{bmatrix} + 
				\begin{bmatrix} 
					\mathbf{G}_{00} & \mathbf{G}_{01} & \mathbf{G}_{02} & \mathbf{G}_{03} &  &  \\ 
					& -\delta t /2 \mathbf{I} &  & -\delta t /2 \mathbf{I} &  &  \\ 
					-\frac{\mathbf{R}_{k}\delta t}{2} & \mathbf{G}_{21} & -\frac{\mathbf{R}_{k+1}\delta t}{2} & \mathbf{G}_{23} &  &  \\ 
					&&&&\delta t \mathbf{I} &  \\ 
					&&&& & \delta t \mathbf{I} 
				\end{bmatrix} 
				\begin{bmatrix} 
					\mathbf{n}_{a_{k}} \\ 
					\mathbf{n}_{g_{k}} \\ 
					\mathbf{n}_{a_{k+1}} \\ 
					\mathbf{n}_{g_{k+1}} \\ 
					\mathbf{n}_{b_{a}} \\ 
					\mathbf{n}_{b_{g}} \\ 
				\end{bmatrix}  
			\end{aligned} 
		\end{equation}
		
		Here, the Jacobian for state variables $\mathbf{J}^{b_{k}}_{t}$ is updated as follows.
		\begin{equation} 
			\begin{aligned} 
				\mathbf{J}^{b_{k}}_{t+\delta t} = (\mathbf{I} + \mathbf{F}_{t}\delta t)\mathbf{J}^{b_{k}}_{t}, \ \ t \in [k,k+1] 
			\end{aligned} 
		\end{equation}
		
		For more details, refer to the \href{https://alida.tistory.com/64\#2.4.-error-state-kinematics-in-discrete-time}{[SLAM] Formula Derivation and Analysis of VINS-mono content summary} post, sections 2.3, 2.4.
	}
	
	\subsection{Code implementations}
	\begin{itemize}
		\item VINS-mono code: \href{https://github.com/HKUST-Aerial-Robotics/VINS-Mono/blob/master/vins_estimator/src/factor/integration_base.h#L180}{integration\_base.h\#L180}
		\begin{itemize}
			\item SO(3) version IMU measurement error $\mathbf{e}_{\mathcal{B}}$ is implemented here.
		\end{itemize}
		\item VINS-mono code: \href{https://github.com/HKUST-Aerial-Robotics/VINS-Mono/blob/master/vins_estimator/src/factor/imu_factor.h#L86}{imu\_factor.h\#L86}
		\begin{itemize}
			\item $\mathbf{J}[0], \mathbf{J}[1], \mathbf{J}[2], \mathbf{J}[3]$ are implemented here.
			\item Jacobian and error function are multiplied by the square root inverse of covariance $\sqrt{(\mathbf{P}^{b_{k}}_{b_{k+1}})^{-1}} = \sqrt{\boldsymbol{\Omega}_{\mathcal{B}}}$ in the form of information matrix.
			\begin{itemize}
				\item $\mathbf{e}_{_{\mathcal{B}},k} \quad \rightarrow \quad \sqrt{\boldsymbol{\Omega}_{\mathcal{B}}}^{\intercal}\mathbf{e}_{_{\mathcal{B}},k}$: in actual code implementation, the right error term is optimized.
				\item This is because the error function  $\mathbf{E}_{_{\mathcal{B}}}(\mathcal{X}) = \mathbf{e}_{\mathcal{B},k}^{\intercal}\boldsymbol{\Omega}_{\mathcal{B}}\mathbf{e}_{_{\mathcal{B}},k}$ is set in the code as the square root $\sqrt{\boldsymbol{\Omega}_{\mathcal{B}}}^{\intercal}\mathbf{e}_{_{\mathcal{B}},k}$.
			\end{itemize}
		\end{itemize}
		\item VINS-mono code: \href{https://github.com/HKUST-Aerial-Robotics/VINS-Mono/blob/master/vins_estimator/src/factor/integration_base.h#L90}{integration\_base.h\#L90}
		\begin{itemize}
			\item The state transition matrices $\mathbf{F}, \mathbf{G}$ for error state equations approximated by Mid-point method are implemented here.
			\item Jacobian update formula for IMU state variables $\mathbf{J}^{b_{k}}_{t+\delta t} = (\mathbf{I} + \mathbf{F}_{t}\delta t)\mathbf{J}^{b_{k}}_{t}$ is implemented here.
			\item Covariance update formula for IMU state variables $\mathbf{P}^{b_{k}}_{t+\delta t} = (\mathbf{I} + \mathbf{F}_{t}\delta t)\mathbf{P}^{b_{k}}_{t}(\mathbf{I}+\mathbf{F}_{t}\delta t)^{\intercal} + (\mathbf{G}_{t}\delta t) \mathbf{Q} (\mathbf{G}_{t}\delta t)^{\intercal}$ is implemented here.
		\end{itemize}
	\end{itemize}

\section{Other Jacobians}
\subsection{Jacobian of unit quaternion}
~\\
\textbf{NOMENCLATURE of Jacobian of unit quaternion}
\begin{itemize}
	\item $\mathbf{X} =  [X,Y,Z,1]^{\intercal} = [\tilde{\mathbf{X}}, 1]^{\intercal} \in \mathbb{P}^{3}$
	\item $\tilde{\mathbf{X}} = [X,Y,Z]^{\intercal} \in \mathbb{P}^{2}$
	\item $\mathbf{q} = [w,x,y,z]^{\intercal} = [w, \mathbf{v}]^{\intercal}$
	\begin{itemize}
		\item Quaternion represented using Hamilton notation. For detailed information, refer to \href{https://alida.tistory.com/60}{this post}.
	\end{itemize}
\end{itemize}

As explained in the previous section on reprojection error, the Jacobian is as follows:
\begin{equation}
	\begin{aligned}
		\mathbf{J}_{c}& = \frac{\partial \hat{\mathbf{p}}}{\partial \tilde{\mathbf{p}}} 
		\frac{\partial \tilde{\mathbf{p}}}{\partial \mathbf{X}'}
		\frac{\partial \mathbf{X}'}{\partial [\mathbf{R}, \mathbf{t}]} \\
	\end{aligned} 
\end{equation}

Among these, $\frac{\partial \mathbf{X}'}{\partial \mathbf{R}}$ is a Jacobian that can be used when rotation is represented by rotation matrix $\mathbf{R}$. In this section, the Jacobian $\frac{\partial \mathbf{X}'}{\partial \mathbf{q}}$ that can be used when the rotation is represented by the unit quaternion $\mathbf{q}$ is described.

When a point $\mathbf{X}$ in three-dimensional space is given, the point $\mathbf{X}'$ rotated by an arbitrary unit quaternion $\mathbf{q}$ can be represented as follows:
\begin{equation}
	\begin{aligned}
		\tilde{\mathbf{X}'} & = \mathbf{q} \otimes \tilde{\mathbf{X}} \otimes \mathbf{q}^{*} \end{aligned}
\end{equation}

Expanding this further:
\begin{equation}
	\begin{aligned}
		\tilde{\mathbf{X}'} & = \mathbf{q} \otimes \tilde{\mathbf{X}} \otimes \mathbf{q}^{*} \\
		& = (w+\mathbf{v}) \otimes \tilde{\mathbf{X}} \otimes (w-\mathbf{v}) \\
		& = w^{2}\tilde{\mathbf{X}} + w(\mathbf{v}\otimes \tilde{\mathbf{X}} - \tilde{\mathbf{X}}\otimes \mathbf{v}) - \mathbf{v}\otimes \tilde{\mathbf{X}} \otimes \mathbf{v} \\
		& = w^{2}\tilde{\mathbf{X}}+2w(\mathbf{v}\times \tilde{\mathbf{X}})- [(-\mathbf{v}^{\intercal}\tilde{\mathbf{X}} + \mathbf{v}\times \tilde{\mathbf{X}}) \otimes \mathbf{v}] \\
		& = w^{2}\tilde{\mathbf{X}} + 2w(\mathbf{v}\times \tilde{\mathbf{X}}) -[(-\mathbf{v}^{\intercal}\tilde{\mathbf{X}})\mathbf{v} + (\mathbf{v}\times \tilde{\mathbf{X}})\otimes \mathbf{v}] \\
		& = w^{2}\tilde{\mathbf{X}} + 2w(\mathbf{v} \times \tilde{\mathbf{X}}) - [(-\mathbf{v}^{\intercal}\tilde{\mathbf{X}})\mathbf{v} - (\mathbf{v}\times \tilde{\mathbf{X}})\times \mathbf{v}] \\
		& = w^{2}\tilde{\mathbf{X}} + 2w(\mathbf{v}\times \tilde{\mathbf{X}}) + 2(\mathbf{v}^{\intercal}\tilde{\mathbf{X}})\mathbf{v} - (\mathbf{v}^{\intercal}\mathbf{v})\tilde{\mathbf{X}}
	\end{aligned}
\end{equation}

Using this, the Jacobian with respect to the quaternion $\frac{\partial \tilde{\mathbf{X}'}}{\partial \mathbf{q}}$ can be determined. It is divided into the scalar part $\frac{\partial \tilde{\mathbf{X}'}}{\partial w}$ and the vector part $\frac{\partial \tilde{\mathbf{X}'}}{\partial \mathbf{v}}$ as follows:
\begin{equation}
	\begin{aligned}
		& \frac{\partial \tilde{\mathbf{X}'}}{\partial w} && =2(w \tilde{\mathbf{X}} + \mathbf{v} \times \tilde{\mathbf{X}}) \\
		& \frac{\partial \tilde{\mathbf{X}'}}{\partial \mathbf{v}} && = -2w[\tilde{\mathbf{X}}]_{\times} + 2(\mathbf{v}^{\intercal}\tilde{\mathbf{X}}\mathbf{I} + \mathbf{v}\tilde{\mathbf{X}}^{\intercal}) - 2 \tilde{\mathbf{X}}\mathbf{v}^{\intercal} \\
		& && = 2(\mathbf{v}^{\intercal}\tilde{\mathbf{X}}\mathbf{I} + \mathbf{v}\tilde{\mathbf{X}}^{\intercal} - \tilde{\mathbf{X}}\mathbf{v}^{\intercal} - w[\tilde{\mathbf{X}}]_{\times})
	\end{aligned}
\end{equation}

In this case, the $\tilde{\mathbf{X}}$ entering in the middle of quaternion multiplication is actually transformed into the form of a pure quaternion $[0, X, Y, Z]^{\intercal}$ with a scalar value of 0. \textbf{Therefore, in the above formula, the Jacobian with respect to the scalar $\frac{\partial \tilde{\mathbf{X}'}}{\partial w}$ is not calculated separately because it is not used in actual optimization, and only the Jacobian with respect to the vector $\frac{\partial \tilde{\mathbf{X}'}}{\partial \mathbf{v}}$ is calculated.}
\begin{equation}
	\begin{aligned}
		\tilde{\mathbf{X}'}  = \mathbf{q} \otimes \tilde{\mathbf{X}} \otimes \mathbf{q}^{*} \quad  \rightarrow \quad \begin{bmatrix} 0 \\ \tilde{\mathbf{X}'} \end{bmatrix} & = \mathbf{q} \otimes \begin{bmatrix} 0 \\  \tilde{\mathbf{X}} \end{bmatrix} \otimes \mathbf{q}^{*} \quad \cdots \text{ strict notation} \\
		\text{Then, } \frac{\partial \tilde{\mathbf{X}'}}{\partial w} &\text{ is going to be useless}
	\end{aligned}
\end{equation}

Additionally, assuming that the quaternion $\mathbf{q}$ is sufficiently small, it can be approximated as the identity ($\mathbf{q} \approx \mathbf{q}_{1} = [1, 0, 0, 0]^{\intercal}$), similar to the method previously used to approximate a sufficiently small rotation matrix $\mathbf{R} \approx \mathbf{I} + [\mathbf{w}]_{\times}$.
\begin{equation}
	\begin{aligned}
		& \left. \frac{\partial \tilde{\mathbf{X}'}}{\partial \mathbf{v}}\right\vert_{\mathbf{q} \approx \mathbf{q}_{1}}
		&& = 2(\mathbf{v}^{\intercal}\tilde{\mathbf{X}}\mathbf{I} + \mathbf{v}\tilde{\mathbf{X}}^{\intercal} - \tilde{\mathbf{X}}\mathbf{v}^{\intercal} - w[\tilde{\mathbf{X}}]_{\times}) \\ & && = -2[\tilde{\mathbf{X}}]_{\times}
	\end{aligned}
\end{equation}

Therefore, the final Jacobian with respect to the quaternion $\frac{\partial \tilde{\mathbf{X}'}}{\partial \mathbf{q}}$ is as follows.
\begin{equation}
	\boxed {\begin{aligned}
			\frac{\partial \tilde{\mathbf{X}'}}{\partial \mathbf{q}} =  -2[\tilde{\mathbf{X}}]_{\times} = -2 \begin{bmatrix} 0 & -Z & Y \\ Z & 0 & -X \\ -Y & X & 0 \end{bmatrix} \in \mathbb{R}^{3\times3}
	\end{aligned} }
\end{equation}

\subsubsection{Code Implementations}
\begin{itemize}
	\item ProSLAM code: \href{https://github.com/NamDinhRobotics/proSLAM/blob/ae0af871e67d9df8dc2a64fa527602e02c5e4072/executables/trajectory_analyzer.cpp#L253}{trajectory\_analyzer.cpp\#L253}
	\begin{itemize}
		\item Based on the blog post by jinyongjeong.
	\end{itemize}
\end{itemize}
	
\subsection{Jacobian of camera intrinsics}
~\\
\textbf{NOMENCLATURE of jacobian of camera intrinsics}
\begin{itemize}
	\item $\pi^{-1}(\cdot) = Z\mathbf{K}^{-1}(\cdot)$: Function that back-projects a point in the image plane to three-dimensional space
	\item $\pi(\cdot) = \pi_{k}(\pi_{h}(\cdot)) = \mathbf{K}(\frac{1}{Z} \cdot)$: Function that projects a point from three-dimensional space onto the image plane
	\item $\mathbf{K} = \begin{bmatrix} f_{x} & 0 & c_{x} \\ 0 & f_{y} & c_{y} \\ 0 & 0 & 1 \end{bmatrix}$: Camera intrinsic parameters
	\item $\mathbf{K}^{-1} = \begin{bmatrix} f_x^{-1} & 0 & -f_x^{-1}c_{x} \\ 0 & f_{y}^{-1} & -f_{y}^{-1}c_{y} \\ 0&0&1  \end{bmatrix}$
	\item $\tilde{\mathbf{K}} = \begin{bmatrix} f_{x} & 0 & c_{x} \\ 0 & f_{y} & c_{y}  \end{bmatrix}$: Omits the last row of the intrinsic parameters for projection from $\mathbb{P}^{2} \rightarrow \mathbb{R}^{2}$.
	\item $\mathbf{X} = [\tilde{\mathbf{X}}, 1]^{\intercal}$
\end{itemize}

Performing camera calibration for SLAM allows obtaining intrinsic parameters (intrinsic matrix) $\mathbf{c} = [f_x, f_y, c_x, c_y]$ and lens distortion parameters $\mathbf{d} = [k_1, k_2, p_1, p_2]$. However, since the calibration values do not exactly match the actual sensor parameters, they can be fine-tuned through optimization. This section describes the process of deriving the Jacobian $\mathbf{J}_{\mathbf{c}}$ for $\mathbf{c}$. It is assumed that the focal lengths $f_{x} \neq f_{y}$.

For example, consider deriving the Jacobian $\mathbf{J}_{\mathbf{c}}$ for the photometric error (\ref{eq:photo3}). It can be expressed as follows:
\begin{equation}
	\begin{aligned}
		\mathbf{J}_{\mathbf{c}} & = \frac{\partial \mathbf{e}}{\partial \mathbf{c}}  \\ & =  \frac{\partial \mathbf{I}}{\partial \mathbf{p}_{2}} \frac{\partial \mathbf{p}_{2}}{\partial \tilde{\mathbf{p}}_{2}} \frac{\partial \tilde{\mathbf{p}}_{2}}{\partial \mathbf{X}'} \frac{\partial \mathbf{X}'}{\partial \mathbf{c}}  \\
		& = \mathbb{R}^{1\times2} \cdot \mathbb{R}^{2\times3} \cdot \mathbb{R}^{3\times4} \cdot \mathbb{R}^{4\times4} = \mathbb{R}^{1\times4} 
	\end{aligned}
\end{equation}

The first term $\frac{\partial \mathbf{I}}{\partial \mathbf{p}_{2}}$ is the Jacobian required for calculating the photometric error, and the remaining three Jacobians are always required irrespective of the reprojection or photometric error terms. \textbf{Thus, deriving $\frac{\partial \mathbf{p}_{2}}{\partial\tilde{\mathbf{p}}_{2}}\frac{\partial \tilde{\mathbf{p}}_{2}}{\partial \mathbf{X}'}\frac{\partial \mathbf{X}'}{\partial \mathbf{c}}$ can be universally applied to the error terms used in SLAM.}

The relationship between points $\mathbf{p}_{1}, \mathbf{p}_{2}$ on the image planes of cameras $\{C_{1}\}, \{C_{2}\}$ can be expressed as follows:
\begin{equation}
	\begin{aligned}
		\mathbf{p}_{1}  & =\begin{bmatrix} u_{1} & v_{1} \end{bmatrix}^{\intercal} \\     
		\mathbf{p}_{2}  & =\begin{bmatrix} u_{2} & v_{2} \end{bmatrix}^{\intercal}
	\end{aligned}
\end{equation}

\begin{equation}
	\begin{aligned}
		\mathbf{p}_{2}  & = \pi(\mathbf{X}') \\
		& = \pi(\mathbf{RX} + \mathbf{t}) \\
		& = \pi(\mathbf{R}\pi^{-1}(\mathbf{p}_{1}) + \mathbf{t})  \quad \cdots \text{ apply back-projection}\\
		& = \pi(\mathbf{R}(Z\mathbf{K}^{-1}\mathbf{p}_{1}) + \mathbf{t}) \\
		& = \pi_{k}(\pi_{h}(\mathbf{R}(Z\mathbf{K}^{-1}\mathbf{p}_{1}) + \mathbf{t})) \\
		& =  \pi_{k}(\frac{Z}{Z'}\mathbf{R}\mathbf{K}^{-1}\mathbf{p}_{1} + \frac{1}{Z'}\mathbf{t}) \quad \cdots \text{ apply } \pi_{h}(\cdot) \\
		&= \frac{Z}{Z'} \tilde{\mathbf{K}}\mathbf{R}\mathbf{K}^{-1}\mathbf{p}_{1} + \frac{1}{Z'}\tilde{\mathbf{K}}\mathbf{t} \quad \cdots \text{ apply } \pi_{k}(\cdot)
	\end{aligned}
\end{equation}

Back projection of $\mathbf{p}_{1}$ followed by transformation matrix application leads to $\mathbf{p}_{2}$ due to a series of projections. As can be seen above, $\frac{\partial \mathbf{p}_{2}}{\partial\tilde{\mathbf{p}}_{2}}\frac{\partial \tilde{\mathbf{p}}_{2}}{\partial \mathbf{X}'}\frac{\partial \mathbf{X}'}{\partial \mathbf{c}}$ includes parameters from $\mathbf{p}_{2}$ to $\mathbf{c}$. Therefore, these three Jacobians must be combined to compute $\frac{\partial \mathbf{p}_{2}}{\partial \mathbf{c}}$ at once:
\begin{equation}
	\begin{aligned}
		\frac{\partial \mathbf{p}_{2}}{\partial \mathbf{c}}
		&= \frac{\partial} {\partial \mathbf{c}} \begin{bmatrix} u_{2} \\ v_{2} \end{bmatrix} \\ &= \frac{\partial} {\partial [f_x, f_y, c_x, c_y]} \begin{bmatrix}  f_{x}\tilde{u}_{2} + c_{x} \\ f_{y}\tilde{v}_{2} + c_{y} \end{bmatrix} \\
		& = \begin{bmatrix}
			\frac{\partial u_{2}}{\partial f_{x}} & \frac{\partial u_{2}}{\partial f_{y}}
			& \frac{\partial u_{2}}{\partial c_{x}}
			& \frac{\partial u_{2}}{\partial c_{y}}
			\\
			\frac{\partial v_{2}}{\partial f_{x}} & \frac{\partial v_{2}}{\partial f_{y}}
			& \frac{\partial v_{2}}{\partial c_{x}}
			& \frac{\partial v_{2}}{\partial c_{y}}
		\end{bmatrix} \\
		&=
		\begin{bmatrix}
			\tilde{u}_{2} + f_{x}\frac{\partial \tilde{u}_{2}}{\partial f_{x}} 
			& f_{x}\frac{\partial \tilde{u}_{2}}{\partial f_{y}}
			& f_{x}\frac{\partial \tilde{u}_{2}}{\partial c_{x}} + 1
			& f_{x}\frac{\partial \tilde{u}_{2}}{\partial c_{y}}
			\\
			f_{y}\frac{\partial \tilde{v}_{2}}{\partial f_{x}} 
			& \tilde{v}_{2} + f_{y}\frac{\partial \tilde{v}_{2}}{\partial f_{y}}
			& f_{y}\frac{\partial \tilde{v}_{2}}{\partial c_{x}}
			& f_{y}\frac{\partial \tilde{v}_{2}}{\partial c_{y}}+1
		\end{bmatrix} \in \mathbb{R}^{2\times4}
	\end{aligned} \label{eq:intrinsic2}
\end{equation}

The elements of the above equation should be calculated next:
\begin{equation} \begin{aligned}
		\begin{pmatrix} \frac{\partial \tilde{u}_{2}}{\partial f_{x}} & \frac{\partial \tilde{u}_{2}}{\partial f_{y}} & \frac{\partial \tilde{u}_{2}}{\partial c_{x}} & \frac{\partial \tilde{u}_{2}}{\partial c_{y}} \\ \frac{\partial \tilde{v}_{2}}{\partial f_{x}} & \frac{\partial \tilde{v}_{2}}{\partial f_{y}} & \frac{\partial \tilde{v}_{2}}{\partial c_{x}} & \frac{\partial \tilde{v}_{2}}{\partial c_{y}} \end{pmatrix} 
	\end{aligned} \label{eq:intrinsic1}  \end{equation}

To derive this, first compute $\tilde{\mathbf{p}}_{2} = [\tilde{u}_{2}, \tilde{v}_{2}, 1]^{\intercal}$ as follows:
\begin{equation}
	\begin{aligned}
		\tilde{\mathbf{p}}_{2} & = [\tilde{u}_{2}, \tilde{v}_{2}, 1]^{\intercal} \\ & = \frac{1}{Z'}\tilde{\mathbf{X}}'  \\
		&=
		\frac{1}{Z'}(\mathbf{R}\tilde{\mathbf{X}} + \mathbf{t}) \\
		&=
		\frac{Z}{Z'}\mathbf{R}\mathbf{K}^{-1}\mathbf{p}_{1} + \frac{1}{Z'}\mathbf{t} \\
		&=
		\frac{Z}{Z'} 
		\begin{bmatrix}
			& & \\
			& \mathbf{R} & \\
			& &
		\end{bmatrix}
		\begin{bmatrix}
			f_{x}^{-1}& & -f_{x}^{-1}c_{x}\\
			& f_{y}^{-1} & -f_{y}^{-1}c_{y} \\
			& & 1
		\end{bmatrix}
		\begin{bmatrix}
			u_{1} \\ v_{1} \\ 1
		\end{bmatrix} 
		+ \frac{1}{Z'}
		\begin{bmatrix}
			t_{x} \\ t_{y} \\ t_{z} 
		\end{bmatrix}  \\
		&=
		\frac{Z}{Z'}
		\begin{bmatrix}
			& & \\
			& \mathbf{R} & \\
			& &
		\end{bmatrix}
		\begin{bmatrix}
			f_{x}^{-1}(u_{1}-c_{x}) \\ 
			f_{y}^{-1}(v_{1}-c_{y}) \\ 
			1
		\end{bmatrix} 
		+ \frac{1}{Z'}
		\begin{bmatrix}
			t_{x} \\ t_{y} \\ t_{z}
		\end{bmatrix} \\
		&=
		\frac{Z}{Z'}
		\begin{bmatrix}
			r_{11}f_{x}^{-1}(u_{1}-c_{x}) + r_{12}f_{y}^{-1}(v_{1}-c_{y}) + r_{13} \\ 
			r_{21}f_{x}^{-1}(u_{1}-c_{x}) + r_{22}f_{y}^{-1}(v_{1}-c_{y}) + r_{23} \\ 
			r_{31}f_{x}^{-1}(u_{1}-c_{x}) + r_{32}f_{y}^{-1}(v_{1}-c_{y}) + r_{33}
		\end{bmatrix} 
		+ \frac{1}{Z'}
		\begin{bmatrix}
			t_{x} \\ t_{y} \\ t_{z}
		\end{bmatrix} \\
	\end{aligned}
\end{equation}

This equation can be organized as follows:
\begin{equation} \begin{aligned} \begin{bmatrix} \tilde{u}_{2} \\ \tilde{v} _{2}\\1 \end{bmatrix}  &= \begin{bmatrix} \frac {r_{11}f_{x}^{-1}(u_{1}-c_{x}) + r_{12}f_{y}^{-1}(v_{1}-c_{y}) + r_{13} + \frac{1}{Z}t_{x}} {r_{31}f_{x}^{-1}(u_{1}-c_{x}) + r_{32}f_{y}^{-1}(v_{1}-c_{y}) + r_{33} + \frac{1}{Z}t_{z}}  \\   \frac {r_{21}f_{x}^{-1}(u_{1}-c_{x}) + r_{22}f_{y}^{-1}(v_{1}-c_{y}) + r_{23} + \frac{1}{Z}t_{y}} {r_{31}f_{x}^{-1}(u_{1}-c_{x}) + r_{32}f_{y}^{-1}(v_{1}-c_{y}) + r_{33} + \frac{1}{Z}t_{z}}  \\ 1 \end{bmatrix} \end{aligned} \end{equation}

Based on this, (\ref{eq:intrinsic1}) can be derived as follows:
\begin{equation} \begin{aligned}
		\frac{\partial \tilde{u}_{2}}{\partial f_{x}} & = \frac{Z}{Z'}(r_{31}\tilde{u}_{2} - r_{11})f_{x}^{-2}(u_{1}-c_{x}) \\
		\frac{\partial \tilde{u}_{2}}{\partial f_{y}} & = \frac{Z}{Z'}(r_{32}\tilde{u}_{2} - r_{12})f_{y}^{-2}(v_{1}-c_{y}) \\
		\frac{\partial \tilde{u}_{2}}{\partial c_{x}} & = \frac{Z}{Z'}(r_{31}\tilde{u}_{2} - r_{11})f_{x}^{-1} \\
		\frac{\partial \tilde{u}_{2}}{\partial c_{y}} & = \frac{Z}{Z'}(r_{32}\tilde{u}_{2} - r_{12})f_{y}^{-1} \\
		\frac{\partial \tilde{v}_{2}}{\partial f_{x}} & = \frac{Z}{Z'}(r_{31}\tilde{v}_{2} - r_{21})f_{x}^{-2}(u_{1}-c_{x}) \\
		\frac{\partial \tilde{v}_{2}}{\partial f_{y}} & = \frac{Z}{Z'}(r_{32}\tilde{v}_{2} - r_{22})f_{y}^{-2}(v_{1}-c_{y}) \\
		\frac{\partial \tilde{v}_{2}}{\partial c_{x}} & = \frac{Z}{Z'}(r_{31}\tilde{v}_{2} - r_{21})f_{x}^{-1} \\
		\frac{\partial \tilde{u}_{2}}{\partial c_{y}} & = \frac{Z}{Z'}(r_{32}\tilde{v}_{2} - r_{22})f_{y}^{-1}
\end{aligned}  \end{equation}

Finally, (\ref{eq:intrinsic2}) appears as follows:
\begin{equation} \boxed{
		\begin{aligned}
			\frac{\partial \mathbf{p}_{2}}{\partial \mathbf{c}}
			&=
			\begin{bmatrix}
				\frac{\partial u_{2}}{\partial f_{x}} & \frac{\partial u_{2}}{\partial f_{y}}
				& \frac{\partial u_{2}}{\partial c_{x}}
				& \frac{\partial u_{2}}{\partial c_{y}}
				\\
				\frac{\partial v_{2}}{\partial f_{x}} & \frac{\partial v_{2}}{\partial f_{y}}
				& \frac{\partial v_{2}}{\partial c_{x}}
				& \frac{\partial v_{2}}{\partial c_{y}}
			\end{bmatrix} \\
			&=
			\begin{bmatrix}
				\tilde{u}_{2} + f_{x}\frac{\partial \tilde{u}_{2}}{\partial f_{x}} 
				& f_{x}\frac{\partial \tilde{u}_{2}}{\partial f_{y}}
				& f_{x}\frac{\partial \tilde{u}_{2}}{\partial c_{x}} + 1
				& f_{x}\frac{\partial \tilde{u}_{2}}{\partial c_{y}}
				\\
				f_{y}\frac{\partial \tilde{v}_{2}}{\partial f_{x}} 
				& \tilde{v}_{2} + f_{y}\frac{\partial \tilde{v}_{2}}{\partial f_{y}}
				& f_{y}\frac{\partial \tilde{v}_{2}}{\partial c_{x}}
				& f_{y}\frac{\partial \tilde{v}_{2}}{\partial c_{y}}+1
			\end{bmatrix} \\
			& = \begin{bmatrix}
				\tilde{u}_{2} + \frac{Z}{Z'}f_{x}^{-1}(r_{31}\tilde{u}_{2} - r_{11})(u_{1}-c_{x})&
				\frac{Z}{Z'}f_{x}f_{y}^{-2}(r_{32}\tilde{u}_{2} - r_{12})(v_{1}-c_{y})&
				\frac{Z}{Z'}(r_{31}\tilde{u}_{2} - r_{11}) + 1&
				\frac{Z}{Z'}f_{x}f_{y}^{-1}(r_{32}\tilde{u}_{2} - r_{12})
				\\
				\frac{Z}{Z'}f_{x}^{-2}f_{y}(r_{31}\tilde{v}_{2} - r_{21})(u_{1}-c_{x})&
				\tilde{v}_{2} + \frac{Z}{Z'}f_{y}^{-1}(r_{32}\tilde{v}_{2} - r_{22})(v_{1}-c_{y})&
				\frac{Z}{Z'}f_{x}^{-1}f_{y}(r_{31}\tilde{v}_{2} - r_{21})&
				\frac{Z}{Z'}(r_{32}\tilde{u}_{2} - r_{12}) + 1
			\end{bmatrix} \in \mathbb{R}^{2\times4}
	\end{aligned} }
\end{equation}

\subsubsection{Code Implementations}
\begin{itemize}
	\item DSO code: \href{https://github.com/JakobEngel/dso/blob/master/src/FullSystem/Residuals.cpp#L123}{Residuals.cpp\#L123}
	\begin{itemize}
		\item For detailed explanation of the code, refer to \href{https://alida.tistory.com/46#6.-code-review}{[SLAM] Direct Sparse Odometry (DSO) Paper and Code Review (2)}.
	\end{itemize}
\end{itemize}

	\subsection{Jacobian of inverse depth}
	~\\
	\textbf{NOMENCLATURE of Jacobian of inverse depth}
	\begin{itemize}
		\item $\mathbf{X} =  [X,Y,Z,1]^{\intercal} = [\tilde{\mathbf{X}}, 1]^{\intercal} \in \mathbb{P}^{3}$
		\item $\tilde{\mathbf{X}} = [X,Y,Z]^{\intercal} \in \mathbb{P}^{2}$
		\item $\rho = \frac{1}{Z}, \rho^{-1} = Z$
	\end{itemize}
	
	\subsubsection{Inverse depth parameterization}
	In SLAM, inverse depth parameterization refers to representing a 3D point $\mathbf{X}$ not with three parameters $[X,Y,Z,1]$ but with a single parameter (the reciprocal of $Z$, $\rho$). This allows a 3D point $\mathbf{X}$ to be fully represented using only the inverse depth $\rho$, given the pixel location $\mathbf{p} = [u,v]$ on the image plane. This offers computational advantages as only one parameter needs to be estimated during optimization.
	
	\subsubsection{Jacobian of inverse depth}
	Let's assume calculating the Jacobian $\mathbf{J}_{\rho}$ for photometric error. It can be expressed as follows:
	\begin{equation}
		\begin{aligned}
			\mathbf{J}_{\rho} & = \frac{\partial \mathbf{e}}{\partial \rho}  \\ & =  \frac{\partial \mathbf{I}}{\partial \mathbf{p}_{2}} \frac{\partial \mathbf{p}_{2}}{\partial \tilde{\mathbf{p}}_{2}} \frac{\partial \tilde{\mathbf{p}}_{2}}{\partial \mathbf{X}'} \frac{\partial \mathbf{X}'}{\partial \rho}  \\
			& = \mathbb{R}^{1\times2} \cdot \mathbb{R}^{2\times3} \cdot \mathbb{R}^{3\times4} \cdot \mathbb{R}^{4\times1} = \mathbb{R}^{1\times1} 
		\end{aligned}
	\end{equation}
	
	Here, the term $\frac{\partial \mathbf{I}}{\partial \mathbf{p}_{2}}$ is the Jacobian needed for computing the photometric error, and the remaining three Jacobians are always required regardless of the reprojection or photometric error terms. \textbf{Therefore, computing $\frac{\partial \mathbf{p}_{2}}{\partial\tilde{\mathbf{p}}_{2}}\frac{\partial \tilde{\mathbf{p}}_{2}}{\partial \mathbf{X}'}\frac{\partial \mathbf{X}'}{\partial  \rho}$ can be universally applied to the error terms used in SLAM.}
	
	First, let's express $\frac{\partial \tilde{\mathbf{p}}_{2}}{\partial \mathbf{X}'}$ in terms of inverse depth, equivalent to substituting $\rho' = \frac{1}{Z'}$:
	\begin{equation}
		\boxed{  \begin{aligned}
				\frac{\partial \tilde{\mathbf{p}}_{2}}{\partial \mathbf{X}'} & = 
				\frac{\partial [\tilde{u}_{2}, \tilde{v}_{2}, 1]}{\partial \mathbf{X}'} \\
				& = \begin{bmatrix} \rho' & 0 & -\rho'^{2}X' & 0\\
					0 & \rho' & -\rho'^{2}Y' &0 \\ 0 & 0 & 0 & 0 \end{bmatrix} \in \mathbb{R}^{3\times 4}
		\end{aligned} }
	\end{equation}
	
	Next, calculate $\frac{\partial \mathbf{X}'}{\partial \rho}$. The expression for $\mathbf{X}'$ can be decomposed as follows:
	\begin{equation}
		\begin{aligned}
			\mathbf{X}' = \begin{bmatrix} \tilde{\mathbf{X}'} \\ 1 \end{bmatrix} & = \begin{bmatrix}\mathbf{R}\tilde{\mathbf{X}} + \mathbf{t} \\ 1 \end{bmatrix} \\ & = \begin{bmatrix} \mathbf{R}\big(Z\mathbf{K}^{-1}\tilde{\mathbf{X}} \big) + \mathbf{t} \\ 1 \end{bmatrix} \\ &= \begin{bmatrix} \mathbf{R}\big( \frac{\mathbf{K}^{-1}\tilde{\mathbf{X}}}{\rho} \big) + \mathbf{t} \\ 1 \end{bmatrix}
		\end{aligned} \label{eq:invd1}
	\end{equation}
	
	Using the above, derive $\frac{\partial \mathbf{X}'}{\partial \rho}$ as follows:
	\begin{equation}
		\boxed{ \begin{aligned}
				\frac{\partial \mathbf{X}'}{\partial \rho} &= \begin{bmatrix}-\mathbf{R} \big( \frac{\mathbf{K}^{-1}\tilde{\mathbf{X}}}{\rho^{2}} \big) \\ 0 \end{bmatrix} \\
				&= \begin{bmatrix} -\frac{\tilde{\mathbf{X}}' - \mathbf{t}}{\rho} \\ 0 \end{bmatrix} \\
				& = -\rho^{-1}\begin{bmatrix} X' - t_{x} \\ Y' - t_{y} \\ Z' - t_{z} \\ 0 \end{bmatrix} \in \mathbb{R}^{4\times 1}
		\end{aligned} } 
	\end{equation}
	
	Using these two Jacobians, finally compute $\frac{\partial \mathbf{p}_{2}}{\partial \rho}$ as follows:
	\begin{equation}
		\boxed{ \begin{aligned}
				\frac{\partial \mathbf{p}_{2}}{\partial \rho} &  = \frac{\partial \mathbf{p}_{2}}{\partial\tilde{\mathbf{p}}_{2}}\frac{\partial \tilde{\mathbf{p}}_{2}}{\partial \mathbf{X}'}\frac{\partial \mathbf{X}'}{\partial  \rho}  \\
				&=  \begin{bmatrix} f_x & 0 & c_{x} \\ 0 & f_y & c_{y} \end{bmatrix} \begin{bmatrix} \rho' & 0 & -\rho'^{2}X' & 0\\
					0 & \rho' & -\rho'^{2}Y' &0 \\ 0 & 0 & 0 & 0 \end{bmatrix} \cdot  -\rho^{-1}\begin{bmatrix} X' - t_{x} \\ Y' - t_{y} \\ Z' - t_{z} \\ 0 \end{bmatrix}  \\
				&= -\rho^{-1}\rho' \begin{bmatrix} f_x(\tilde{u}_{2}t_{z} - t_{x}) \\ f_y(\tilde{v}_{2}t_{z} - t_{y})  \end{bmatrix} \in \mathbb{R}^{2 \times 1}
		\end{aligned} } 
	\end{equation}
	- $\tilde{u}_{2} = \frac{X'}{Z'} = \rho'  X'$ \\
	- $\tilde{v}_{2} = \frac{Y'}{Z'} = \rho'  Y'$ \\
	
	\subsubsection{Code Implementations}
	\begin{itemize}
		\item DSO code: \href{https://github.com/JakobEngel/dso/blob/master/src/FullSystem/CoarseInitializer.cpp#L424}{CoarseInitializer.cpp\#L424}
		\begin{itemize}
			\item For a detailed explanation of the code, refer to \href{https://alida.tistory.com/46#6.-code-review}{[SLAM] Direct Sparse Odometry (DSO) Paper and Code Review (2)}.
		\end{itemize}
	\end{itemize}

\section{References}
\begin{enumerate}[label={[\arabic*]}]
	\item \href{https://alida.tistory.com/51}{[Blog] [SLAM] Bundle Adjustment Concept Review: Reprojection error}
	\item \href{https://alida.tistory.com/52}{[Blog] [SLAM] Optical Flow and Direct Method Concept and Code Review: Photometric error}
	\item \href{https://alida.tistory.com/16}{[Blog] [SLAM] Pose Graph Optimization Concept Explanation and Example Code Analysis: Relative pose error}
	\item \href{https://alida.tistory.com/12}{[Blog] Plücker Coordinate Concept Summary: Line projection error}
	\item \href{https://alida.tistory.com/64}{[Blog] [SLAM] Formula Derivation and Analysis of the VINS-mono Content Summary: IMU measurement error}
\end{enumerate}

	\section{Revision log} 
	\begin{itemize}
		\item 1st: 2023-01-21
		\item 2nd: 2023-01-22
		\item 3rd: 2023-01-25
		\item 4th: 2023-01-28
		\item 5th: 2023-09-26
		\item 6th: 2023-11-14
		\item 7th: 2024-02-06
		\item 8th: 2024-04-02
		\item 9th: 2024-05-01
	\end{itemize}

\end{document}